\documentclass{article}
\usepackage{arxiv}
\renewcommand{\today}{November 25, 2026}
\usepackage[utf8]{inputenc}
\usepackage[T1]{fontenc}
\usepackage[svgnames,dvipsnames]{xcolor}
\usepackage{makecell,colortbl,dcolumn,stfloats,xspace,footmisc}

\usepackage[authoryear]{natbib}
\usepackage{amsmath,amssymb,amsfonts,bm}
\usepackage{amsthm}
\usepackage{graphicx}
\usepackage{booktabs}
\usepackage{array}
\usepackage{multirow}
\usepackage{subcaption}
\usepackage{tikz}
\usepackage{float}
\usepackage{placeins}
\usepackage{siunitx}

\usepackage{amsmath,amsfonts,bm}

\def\eqref#1{equation~\ref{#1}}

\def\1{\bm{1}}

\DeclareMathAlphabet{\mathsfit}{\encodingdefault}{\sfdefault}{m}{sl}
\SetMathAlphabet{\mathsfit}{bold}{\encodingdefault}{\sfdefault}{bx}{n}

\def\sQ{{\mathbb{Q}}}
\def\sR{{\mathbb{R}}}

\def\sZ{{\mathbb{Z}}}

\usepackage{xcolor}%
\AtBeginDocument{\expandafter\def\expandafter\UrlBreaks\expandafter{\UrlBreaks\do\-}}

\newcommand{\Dset}{\mathbb{D}}                 %
\newcommand{\Dn}{\mathbb{D}_{N}}               %
\newcommand{\Aadd}{\mathcal{A}^{+}_{N}}        %
\newcommand{\Amul}{\mathcal{A}^{\times}_{N}}   %
\newcommand{\epsadd}{\varepsilon_{+}}
\newcommand{\epsmul}{\varepsilon_{\times}}
\newcommand{\simplex}{\mathcal{S}}               %

\newcommand{\str}[1]{\mathrm{str}(#1)}        %
\newcommand{\rel}{\mathrm{rel}}                %
\newcommand{\distP}{\mathcal{P}}              %
\newcommand{\norm}[1]{\left\lVert #1\right\rVert}

\newtheorem{theorem}{Theorem}[section]
\newtheorem{lemma}[theorem]{Lemma}
\newtheorem{proposition}[theorem]{Proposition}
\newtheorem{corollary}[theorem]{Corollary}
\theoremstyle{definition}
\newtheorem{definition}[theorem]{Definition}

\theoremstyle{remark}
\newtheorem{remark}[theorem]{Remark}
\theoremstyle{plain}

\usepackage{url}
\usepackage[colorlinks]{hyperref}
\hypersetup{linkcolor=DarkSlateGrey, urlcolor=DarkSlateGrey, citecolor=DarkSlateGrey,
  pdftitle={Neural Isomorphic Fields: A Transformer-based Algebraic Numerical Embedding},
  pdfauthor={Hamidreza Sadeghi, Saeedeh Momtazi, Reza Safabakhsh}}

\title{Neural Isomorphic Fields: A Transformer-based Algebraic Numerical Embedding}

\author{
  Hamidreza Sadeghi\\
  Department of Computer Engineering\\
  Amirkabir University of Technology\\
  Tehran, Iran\\
  \And
  Saeedeh Momtazi\thanks{Corresponding author.}\\
  Department of Computer Engineering\\
  Amirkabir University of Technology\\
  Tehran, Iran\\
  \texttt{momtazi@aut.ac.ir}\\
  \And
  Reza Safabakhsh\\
  Department of Computer Engineering\\
  Amirkabir University of Technology\\
  Tehran, Iran\\
}
\begin{document}
\renewcommand{\topfraction}{.9}
\renewcommand{\bottomfraction}{.8}
\renewcommand{\textfraction}{.07}
\renewcommand{\floatpagefraction}{.85}
\setcounter{topnumber}{9}
\setcounter{bottomnumber}{9}
\setcounter{totalnumber}{20}
\setlength{\floatsep}{10pt plus 2pt minus 2pt}
\setlength{\textfloatsep}{12pt plus 2pt minus 2pt}
\setlength{\intextsep}{10pt plus 2pt minus 2pt}
\makeatletter
\setlength{\@fptop}{0pt}
\setlength{\@fpsep}{10pt plus 1fil}
\setlength{\@fpbot}{0pt plus 1fil}
\makeatother

\maketitle
\setcounter{footnote}{0}

\begin{abstract}
Neural network models often face challenges when processing very small or very large numbers due to issues such as overflow, underflow, and unstable output variations. To mitigate these problems, we propose using embedding vectors for numbers instead of directly using their raw values. These embeddings aim to retain essential algebraic properties while preventing numerical instabilities. We introduce Neural Isomorphic Fields (NIF), a digit-sequence Transformer autoencoder trained with two learned commutative operators and an antisymmetric neural comparator. At zero loss, the embeddings provably form an ordered integral domain isomorphic to the decimal fractions, whose field of fractions is isomorphic to the rationals. For the trained model, we bound errors of operations, chains and algebraic laws by decoding margins, defects and Lipschitz constants. Trained on up to 25 digits, NIF reaches at least 99.9\% exact match on closure, identity, associativity and distributivity and 99.8\% on ten-step chains and the order relation. With four channels per decimal place, reconstruction and both operations exceed 99.7\% exact match on numbers of up to 200 digits whose parts stay within training lengths, so the dimension grows linearly. In a language model, frozen operators raise exact match on 14--20-digit word problems from 12.0\% (operators off) to 39.7\%.\footnote{The code and the results of all experiments are available at \url{https://github.com/HRSadeghi/Neural-Isomorphic-Fields}.}
\end{abstract}

\section{Introduction}

Representing and processing numerical data is a core challenge in various artificial intelligence domains, including Natural Language Processing (NLP) and Computer Vision (CV). One of the methods to represent and process the numbers is to embed them. A key distinction arises between \textit{numeral embeddings}, which treat numbers as discrete linguistic tokens (e.g., 25 as a word) \citep{jiang-etal-2020-learning}, and \textit{number embeddings}, which aim to encode the continuous, quantitative properties of numerical values (e.g., the real number 25) \citep{zhou2025foneprecisesingletokennumber, golkar2023xval}. Most existing works have primarily focused on \textit{numeral embeddings}, addressing challenges such as lexical variability and contextual usage of numeric tokens \citep{jiang-etal-2020-learning, sundararaman-etal-2020-methods}. However, they often fail to reflect deeper algebraic structures inherent to numbers. To address this, we propose a new framework called \textit{Neural Isomorphic Fields} (\textit{NIF}). This framework leverages algebraic principles to construct number embeddings that preserve the intrinsic mathematical properties of numbers, enabling richer and more semantically meaningful representations.

Advances in deep learning have shown that embedding spaces can capture complex relationships in data. For instance, static word embeddings like Word2Vec \citep{mikolov2013wor2vec}, GloVe \citep{pennington-etal-2014-glove} and FastText \citep{bojanowski2017fasttext} as well as contextualized word embeddings like BERT \citep{devlin2019bert}, XLNet \citep{yang2019xlnet}, BART \citep{lewis-etal-2020-bart}, ELECTRA \citep{clarck2020electra}, ALBERT \citep{lan2020ALBERT}, ERNIE \citep{zhang-etal-2019-ernie}, T5 \citep{raffel-2020-t5}, BigBird \citep{zaheer2020bigbird} and DeBERTa \citep{he2021deberta} have demonstrated the power of vector spaces in preserving syntactic and semantic relationships between words. Similarly, graph embeddings \citep{hamilton2017IRLLG, kipf2017semisupervised, veličković2018graph, xu2018how, you2020GCLA} have been successful in preserving the structural properties of graphs. These methods inspire us to explore the potential of embedding spaces in preserving the algebraic properties of numbers.

In the context of numerical data, works such as Neural Arithmetic Logic Units (NALU) \citep{trask2018NALU} and Neural Arithmetic Units (NAU) \citep{mnu-madsen-johansen-2020} have proposed neural network architectures that are specifically designed to perform arithmetic operations. These architectures can explicitly learn addition and multiplication operations, marking significant progress towards integrating arithmetic capabilities within neural networks. Additionally, approaches like Neural GPUs \citep{kaiser2016neural} and Program-Executing Networks \citep{zaremba2014learning} demonstrate the potential of neural networks to execute algorithms, including those for arithmetic operations. These models work on raw values or digit strings and give no embedding vector for a number, so other models cannot use them as number representations. \cite{jiang-etal-2020-learning} proposed two novel numeral embedding methods to address the Out-Of-Vocabulary (OOV) problem for numerals. However, these embeddings do not fully possess the ability to understand and work with numerals or numbers, which is known as \textit{numeracy} \citep{georgios-etal-2018-numeracy}. Numeracy includes understanding the order of numbers or numerals, their magnitude, their units, and their arithmetic operations. The NIF embedding introduced in this paper generates a distinct embedding vector for each number and, by being trained to preserve the structure and algebraic properties, addresses several requirements of numeracy.

To maintain algebraic properties, we construct number embeddings such that the resulting set of embedding vectors closely approximates an isomorphic structure to the numbers they represent. Rational and real numbers, under the operations of addition and multiplication, form ordered fields \citep{herstein1996abstract, rudin1976principles}. A finite digit sequence, however, can only represent a decimal fraction $m/10^{\nu}$ with $m\in\sZ$ and $\nu\ge 0$. Decimal fractions form an ordered integral domain, not a field; $1/3$, for example, has no finite decimal form. We therefore build the embedding space as an (ideally) isomorphic copy of the ordered ring of decimal fractions. The rational numbers are then obtained as its field of fractions, where $p/q$ is represented by the pair of embeddings of $p$ and $q$ (Section~\ref{sec:papdt}). Given that the decimal fractions are countable \citep{lin1985set}, we can define a probability distribution over them where the probability assigned to any individual number is nonzero. This set serves as the basis for the isomorphic mapping in our framework.

It is important to emphasize that while the theoretical construction assumes ideal conditions for forming an isomorphic structure, neural networks inherently introduce errors, preventing the realization of these ideal conditions in practice. Consequently, the structure induced on the embedding vectors only approximates an isomorphic mapping rather than achieving it perfectly. For this reason we treat two cases separately. In the ideal zero-loss case, we prove that the embedding space is isomorphic to the decimal fractions and its field of fractions to $\sQ$. For the trained model, whose losses are not zero, we bound and measure the errors (Section~\ref{sec:approximate}).

To effectively embed numbers, it is essential to first establish their representation method. In prior models such as NALU and NAU, numbers have been represented using floating-point or integer formats. Additionally, some earlier approaches have employed scientific notation \citep{zhang-etal-2020-language-embeddings} or subword tokenization methods. Tokenization techniques like Byte Pair Encoding (BPE) \citep{gage1994bpe}, WordPiece \citep{wu2016wordPiece}, and SentencePiece \citep{kudo-richardson-2018-sentencepiece} are widely used in Language Models (LMs) \citep{radford2018improving, radford2019language} and Large Language Models (LLMs) \citep{brown2020gpt3, touvron2023llama, touvron2023llama2} to represent both words and numbers. However, our model requires a comprehensive and precise representation of numbers, particularly because very large or very small numbers cannot be processed by neural networks due to underflow or overflow issues. Therefore, rather than relying on conventional methods like scientific notation or tokenization, we represent numbers as sequences of their individual digits. This approach ensures that all numbers with a finite decimal representation and a bounded number of digits can be effectively processed by the neural network without encountering numerical limitations.

A Transformer \citep{vaswani2017transformer} encodes the digit sequence into place-value slots, one for each decimal exponent. The multiplication operator follows a split of the product into steps whose values are no larger than those of an addition, and these steps are supervised in training (Section~\ref{sec:proposedmethod}).

\paragraph{Novelty} Existing number embeddings encode the magnitude, the digits or the floating-point format of a number \citep{golkar2023xval, han2023lunalanguageunderstandingnumber, zhou2025foneprecisesingletokennumber, jin2024embedding, kreitner2025efficientnumeracylanguagemodels}. They are not trained so that arithmetic, order and the algebraic laws hold in the embedding space. Neural arithmetic units \citep{trask2018NALU, mnu-madsen-johansen-2020} compute on raw values and give no embedding of a number. Representation learning with an algebraic objective, such as homomorphism autoencoders \citep{keurti2023homomorphism}, learns group actions from observed transitions, not the ring structure of numbers (Section~\ref{sec:related_works}, Table~\ref{tab:related_compare}). NIF differs from this work in three ways. First, one digit-based encoder is trained \emph{jointly} with two learned commutative operators and a learned comparator, so that addition, multiplication and order carry over to the embedding space. Second, the target structure is stated exactly (the ordered ring of decimal fractions and its field of fractions $\sQ$), with sufficient conditions for an isomorphism and error bounds for the trained, non-ideal model. Third, the model is tested axiom by axiom and not only by end-task accuracy.
\par

The key contributions of this paper are:
\begin{itemize}
	\item Introducing the NIF framework for embedding numbers while training the embedding space to reproduce their algebraic properties (addition, multiplication and order), together with learned commutative operators and an antisymmetric neural comparator. The embedding is divided into place-value slots, and the multiplication operator computes the product in steps with small values, each supervised in training. The operators read the digit of each place from its slot. An encoder trained for reconstruction only does not give such a code, and the operators then fail on products; the jointly trained encoder learns it from the digit tokens.

	\item Establishing ideal conditions and proving that under these conditions the embedding space is an ordered integral domain isomorphic to the decimal fractions, and its field of fractions an ordered field isomorphic to $\sQ$.

	\item Deriving bounds for the trained model with non-zero losses, which relate decoding margins, algebraic defects and the Lipschitz constants of the operators to errors in single operations, in chained computations and in the associative and distributive laws.

	\item Presenting comprehensive evaluation criteria for the addition and multiplication operators, the order and the induced field of fractions. We add ablations of the operators and the training objective, a study of number length and embedding dimension, an analysis of error propagation, experimental comparisons with LUNA, FoNE and xVal, each with the decoder and operators that its representation allows, and, under identical conditions, with FoNE and one-hot digit codes, and a downstream test in which the trained operators work inside the layers of a language model.
\end{itemize}

The paper is organized as follows: Section \ref{sec:related_works} reviews prior approaches to number embeddings and their limitations in preserving algebraic properties like addition and multiplication. Section \ref{sec:proposedmethod} presents our novel method for constructing number embeddings that maintain these properties, the ideal conditions, and the analysis of the trained model. Section \ref{sec:evaluation_criteria} defines the evaluation metrics for assessing algebraic properties such as identity, closure, and invertibility. Section \ref{sec:expriments} details the experimental setup, data generation, model architecture, and results, with a focus on arithmetic operations, including the downstream evaluation. Finally, Section \ref{sec:conclusion} summarizes the findings and Section \ref{sec:future_works} discusses future directions.

\FloatBarrier
\section{Related work}\label{sec:related_works}

Handling numerical inputs in neural networks has received significant attention from researchers in recent years. Our work builds upon and extends these efforts by introducing a novel approach to number embedding that preserves algebraic properties. The related work in this area can be categorized into four groups: number embeddings (\ref{subsec:number_embeddings}), numeral embeddings (\ref{subsec:numeral_embeddings}), arithmetic operations in neural networks (\ref{subsec:arithmetic_operations_in_neural_networks}), and algebraic approaches to neural networks (\ref{subsec:algebraic_approaches_to_neural}). In this section we review related work in these four fields; Table~\ref{tab:related_compare} summarizes the comparison.

\subsection{Number embeddings}\label{subsec:number_embeddings}

Recent advancements in number embeddings have proposed various approaches to improve the representation of numerical values within language models. \cite{zhou2025foneprecisesingletokennumber} introduce \textit{FoNE}, a single-token embedding method that employs Fourier features to encode numbers. By utilizing Fourier transformations based on the digits of numbers, FoNE constructs compact and precise embeddings. Given an embedding vector of length $d$, this method can embed numbers of at most $d/2$ digits. It is important to note that FoNE does not aim to enable direct arithmetic operations such as addition or multiplication in the embedding space; rather, it evaluates the effectiveness of number embeddings for such operations by applying them within a large language model.

In a complementary direction, \cite{golkar2023xval} propose \textit{xVal}, a continuous embedding approach tailored for scientific language models. xVal represents numbers as continuous vectors that preserve essential numerical properties, such as magnitude and relative differences, thereby enhancing model performance in domains like physics and engineering where discretization can impose significant limitations. xVal scales one fixed vector by the (normalized) value of the number. Its map is therefore additive in the value, but the inputs must be normalized, and numbers of very different magnitudes lose resolution.

\cite{han2023lunalanguageunderstandingnumber} contribute to this line of research with \textit{LUNA}, a framework aimed at improving number understanding in transformers. LUNA introduces a modular ``Number Plugin'' architecture, treating numbers differently from textual tokens through specialized augmentations. This mechanism enriches number representations with a dedicated embedding layer that captures both the value and the structural format of numbers more effectively.

Building upon digit-based encoding schemes, \cite{sivakumar-moosavi-2025-leverage} propose an explicit aggregation method for digit embeddings. Their approach involves decomposing numbers into individual digits, embedding each digit, and then computing a weighted sum of these embeddings based on positional weights reflecting the base-10 numeral system. This aggregated embedding is then integrated into transformer models either by introducing a special token in the input sequence or by incorporating an auxiliary loss function during training. Unlike methods that split numbers into digit tokens and rely on the model to implicitly learn the aggregation of digit embeddings, this explicit approach provides direct supervision, enhancing the model's numerical reasoning capabilities without necessitating architectural modifications.

Two recent methods start from the IEEE~754 floating-point format. Floating-Point Embedding (FPE) \citep{jin2024embedding} computes the embedding of a number from its floating-point representation, and BitTokens \citep{kreitner2025efficientnumeracylanguagemodels} use one token per number whose embedding holds its binary floating-point code. Both keep a fixed significand (24 bits in single and 53 bits in double precision, about 7 and 16 decimal digits), so the low-order digits of long numbers are lost. On the output side, \cite{zausinger2025regress} replace the cross-entropy loss on number tokens with a loss closer to regression. These methods make language models better with numbers, but, like the methods above, they train no operations and no order in the embedding space.

Overall, all these methods share the objective of mapping numbers to fixed-length vectors. Moreover, all methods, except for xVal, rely on digit-based or bit-based encoding schemes, enabling them to robustly handle very large or very small numbers (for the bit-based ones, within the precision of the floating-point format). In this regard, they align closely with the approach adopted in our work. NIF, in turn, fixes a window of place-value slots in advance, so its dimension grows linearly with the number of digits (Section~\ref{subsubsec:ablation}).

\subsection{Numeral embeddings}\label{subsec:numeral_embeddings}

 Jiang et al. proposed two methods for numeral embedding to handle OOV numbers \citep{jiang-etal-2020-learning}. Their approach uses either a self-organizing map (SOM) or Gaussian mixture model (GMM) to induce a finite set of prototype numerals, representing the embedding of a numeral as a weighted average of these prototypes. While effective for certain tasks, this method does not explicitly preserve algebraic operations.

 \cite{sundararaman-etal-2020-methods}  introduced DICE (Deterministic, Independent-of-Corpus Embeddings)  for numerals, where cosine similarity reflects actual distance on the real numbers. They also proposed a regularization approach to learn model-based embeddings of numbers in context. \cite{thawani2021survey} survey number representations in NLP.

\subsection{Arithmetic operations in neural networks}\label{subsec:arithmetic_operations_in_neural_networks}

\cite{Franco1998solving} designed feed-forward multi-layer neural networks capable of performing basic arithmetic operations. However, their work focused on solving specific tasks rather than embedding arithmetic functionality inherently within the number representations. Subsequently, \cite{trask2018NALU} introduced the Neural Arithmetic Logic Unit (NALU), enhancing neural networks with explicit mechanisms for executing arithmetic operations. Although NALU can compute operations such as addition and multiplication between two inputs, it does not provide a dedicated embedding for individual numbers. Neural Arithmetic Units \citep{mnu-madsen-johansen-2020} make such units more stable and better at extrapolation for addition and multiplication, and Neural Power Units \citep{heim2020npu} learn multiplication, division and arbitrary powers in one layer. Both work on raw scalar inputs.

More recently, \cite{mcleish2024transformers} demonstrated that standard transformer architectures, when equipped with suitably structured embeddings, are capable of performing arithmetic tasks. They proposed \emph{Abacus Embeddings}, a method that encodes each digit's position relative to the number's start, enabling transformers to achieve state-of-the-art performance on arithmetic benchmarks without requiring any architectural modifications. Other work changes the format of the digit string, for example by reversing the digits or by adding the length of the number \citep{lee2024teaching, schwartz-etal-2024-numerologic}. Interpretability studies find that pre-trained language models add numbers with Fourier features \citep{zhou2024fourier} or with representations on helices \citep{kantamneni2025trigonometry}, and that they combine many simple heuristics for arithmetic \citep{nikankin2025arithmetic}. These studies look at how a model handles digit tokens. NIF instead gives one vector per number, and the geometry of these vectors is trained to follow the operations.

\subsection{Algebraic approaches to neural networks}\label{subsec:algebraic_approaches_to_neural}

\cite{parada2021alg} studied algebraic neural networks (AlgNNs) with commutative algebras, unifying diverse architectures under algebraic signal processing. Their work on stability to deformations provides insights into the algebraic properties of neural networks, which our research builds upon. \cite{Zimmer1997} took steps towards an algebraic theory of neural networks by defining a new notion of sequential composition. This work focused on network composition and does not apply algebraic concepts directly to number representations.

Group-structured representation learning looks for representations on which a group acts linearly \citep{higgins2018definition}. The homomorphism autoencoder of \cite{keurti2023homomorphism} learns such a representation from observed transitions, training an equivariant encoder together with a homomorphism from the acting group to a group of invertible matrices. Its objective is the closest to ours, but it covers one group action. NIF targets two operations, the distributive law that links them and an order, that is, the structure of an ordered ring. Studies of grokking show that small networks trained on modular addition form internal representations that reflect the cyclic group \citep{power2022grokking, nanda2023progress}. In that setting the structure appears inside a network trained for one task on a finite group; NIF imposes it on an explicit embedding of an infinite ring.

\begin{table}[htbp]
	\centering
	
	\caption{Comparison of NIF with related number representations and arithmetic models. \textquotedblleft Ops.\ in emb.\ space\textquotedblright: operations that are trained or defined on the embedding vectors; \textquotedblleft Alg.\ eval.\textquotedblright: evaluation of individual algebraic laws; $<_{\bar\omega}$: order induced by the antisymmetrized comparator (Section~\ref{sec:neural_order}).}
	\label{tab:related_compare}
	\renewcommand{\arraystretch}{1.15}
	\footnotesize
	\begin{tabular}{@{}>{\raggedright\arraybackslash}p{0.3\textwidth} >{\raggedright\arraybackslash}p{0.11\textwidth} >{\raggedright\arraybackslash}p{0.19\textwidth} >{\raggedright\arraybackslash}p{0.11\textwidth} l l@{}}
		\toprule
		\textbf{Method} & \textbf{Encoder input} & \textbf{Structure targeted} & \textbf{Ops.\ in emb.\ space} & \textbf{Order} & \textbf{Alg.\ eval.} \\
		\midrule
		DICE \citep{sundararaman-etal-2020-methods} & value & distance, magnitude & -- & -- & -- \\
		xVal \citep{golkar2023xval} & normalized value & linear in the value & $+$ (implicit) & implicit & -- \\
		LUNA \citep{han2023lunalanguageunderstandingnumber} & digit string & value and format & -- & -- & -- \\
		FoNE \citep{zhou2025foneprecisesingletokennumber} & value & periodic (Fourier) features per digit place & -- & -- & -- \\
		FPE, BitTokens \citep{jin2024embedding, kreitner2025efficientnumeracylanguagemodels} & IEEE~754 bits & floating-point format & -- & -- & -- \\
		NALU, NAU, NPU \citep{trask2018NALU, mnu-madsen-johansen-2020, heim2020npu} & raw scalars & arithmetic units & $+,\times,\div$, powers on scalars & -- & -- \\
		Homomorphism AE \citep{keurti2023homomorphism} & observations & group action & group action & -- & -- \\
		\textbf{NIF (ours)} & canonical digit string & ordered ring $\Dset$, field of fractions $\sQ$ & learned $\dot{+},\dot{\times}$ & learned $<_{\bar\omega}$ & yes \\
		\bottomrule
	\end{tabular}
\end{table}

\FloatBarrier
\section{Proposed method} \label{sec:proposedmethod}

We construct embedding vectors for numbers, with two learned binary operations and a learned comparison, so that the embedding space reproduces the algebraic structure of the numbers. Since finite digit sequences represent exactly the decimal fractions (Section~\ref{subsec:number_representation}), the target is the ordered ring of decimal fractions; the ordered field of rational numbers follows as its field of fractions (Corollary~\ref{cor:fraction_field}). We assume that the readers are familiar with the definitions of a group, a field, isomorphism, and order\footnote{To maintain brevity, these definitions are not provided in this paper; instead, the reader is referred to books on abstract algebra \citep{herstein1996abstract}, mathematical analysis \citep{rudin1976principles}, and set theory \citep{lin1985set} for detailed explanations. For an order on a set and an ordered field, we follow \citet[Definitions~1.5 and~1.17]{rudin1976principles}; definitions specific to this paper are given below.}.

\subsection{Number representation}\label{subsec:number_representation}
To embed numbers, it is essential to determine their representation format. A practical approach to represent both large and small numbers is to use a sequence of digits and symbols (positive, negative, decimal point). Therefore, the number \(\mathbf{x}\) is represented as the sequence \((x_i)_{i=1}^n\) where \(x_i \in \{0, \ldots, 9, \text{.}, -, +, s\}\) (also, refer to Table \ref{tab:notations} for other notations). To reconstruct the number in the decoder, a start token ($s$) is required, resulting in an input vocabulary of 14 tokens (plus a padding token). No end token is needed, since the decoder predicts the length of its output (Section~\ref{subsec:architecture}).

\paragraph{Represented numbers} A finite digit sequence can only represent a number with a terminating decimal expansion, so we state the domain of the encoder explicitly. Let
\begin{equation}\label{eq:decimal_fractions}
	\Dset := \sZ\!\left[\tfrac{1}{10}\right] = \left\{ m\,10^{-\nu} \;:\; m\in\sZ,\ \nu\in\{0,1,2,\dots\}\, \right\}
\end{equation}
be the set of \emph{decimal fractions}. The \emph{canonical string} $\str{a}$ of $a\in\Dset$ consists of the sign token ($-$ if $a<0$ and $+$ otherwise), the decimal digits of $\lfloor |a| \rfloor$ without leading zeros (the single digit $0$ if $\lfloor |a|\rfloor=0$), and, only if $|a|\notin\sZ$, the decimal point followed by the shortest digit string that represents $|a|-\lfloor |a|\rfloor$ (i.e., without trailing zeros). The map $a\mapsto\str{a}$ is injective. Every finite digit sequence with at most one decimal point represents an element of $\Dset$, and every element of $\Dset$ has exactly one canonical string. The encoder is therefore a map $\varphi:\Dset\rightarrow\sR^d$, $a\mapsto\varphi(\str{a})$; numbers such as $1/3$, whose decimal expansion does not terminate, lie outside its domain. In the implementation, each number is converted to its canonical string before it is encoded or used as a target.

\begin{lemma}[Algebraic structure of the domain]\label{lem:domain}
	The set $\Dset$ has the following properties.
	\begin{enumerate}
		\item[(a)] $\Dset$ is a subring of $\sQ$ that contains $\sZ$; with the usual order, $(\Dset,+,\times,<)$ is an ordered integral domain (Definition~\ref{def:ordered_domain}).
		\item[(b)] A rational number $p/q$ with $p\in\sZ$, $q\in\sZ_{>0}$ and $\gcd(p,q)=1$ belongs to $\Dset$ if and only if $q=2^i5^j$ for some integers $i,j\ge 0$.
		\item[(c)] The group $\Dset^{\times}$ of units (invertible elements) of $\Dset$ consists exactly of the numbers $\pm 2^i5^j$ with $i,j\in\sZ$. In particular, $\Dset$ is not a field: $3\in\Dset$, but $1/3\notin\Dset$.
		\item[(d)] The field of fractions $\mathrm{Frac}(\Dset)$ of $\Dset$ is (canonically isomorphic to) $\sQ$.
	\end{enumerate}
\end{lemma}

\begin{remark}[Consequence for the algebraic claims]\label{rem:domain}
	By Lemma~\ref{lem:domain}, the encoder can reproduce on its domain the ordered integral domain $(\Dset,+,\times,<)$, but not the field $\sQ$, since most elements of $\Dset$ have no multiplicative inverse in $\Dset$. All exact statements in Section~\ref{sec:papdt} are therefore made for $\Dset$. With the same representation, $\sQ$ is obtained as the field of fractions of the embedding space, where a rational number $p/q$ is represented by the pair of embedding vectors $(\varphi(p),\varphi(q))$ (Corollary~\ref{cor:fraction_field}).
\end{remark}

\paragraph{Finite precision} In practice, the number of digits is limited to a maximum $N$ (Section~\ref{sec:expriments}); $n$ always denotes the length of a single token sequence $\mathbf{x}=(x_i)_{i=1}^n$. Let
\begin{equation}\label{eq:Dn}
	\Dn := \{ a\in\Dset \;:\; \str{a} \text{ contains at most } N \text{ digits}\}.
\end{equation}
Only the digits $0$--$9$ of $\str{a}$ are counted, including the leading $0$ of a number with $|a|<1$; an element of $\Dn$ thus has at most $N$ integer digits or at most $N-1$ fractional digits. The set $\Dn$ is finite and closed under neither addition nor multiplication; the product of two $N$-digit integers, for example, can have $2N$ digits. Statements about the trained model therefore use the \emph{admissible pairs}
\begin{equation}\label{eq:admissible}
	\Aadd := \{(a,b)\in\Dn^2 \;:\; a+b\in\Dn\},\qquad \Amul := \{(a,b)\in\Dn^2 \;:\; ab\in\Dn\}.
\end{equation}
\par

\begin{table}[t]
	\centering
	\caption{Notation.}
	\label{tab:notations}
	\renewcommand{\arraystretch}{1.12}
	\footnotesize
	\begin{tabular}{@{}l l >{\leavevmode}p{0.66\textwidth}@{}}
		\toprule
		\textbf{Notation} & \textbf{Short form} & \textbf{Description} \\
		\midrule
		$\sQ$ & -- & The set of rational numbers \\
		$\sR$ & -- & The set of real numbers \\
		$\Dset$ & -- & The ring of decimal fractions $\sZ[1/10]$, Eq.~(\ref{eq:decimal_fractions}) \\
		$\Dn$ & -- & Decimal fractions with at most $N$ digits, Eq.~(\ref{eq:Dn}) \\
		$N$ & -- & Maximum number of digits of a represented number \\
		$\Dset^{\times}$, $\mathrm{Frac}(\Dset)$ & -- & Units of $\Dset$ and field of fractions of $\Dset$, Lemma~\ref{lem:domain} \\
		$\str{a}$ & -- & Canonical string of $a\in\Dset$, Section~\ref{subsec:number_representation} \\
		$\bot$ & -- & Decoder output that is not the canonical string of an element of $\Dset$ \\
		$K$, $c$, $\mathbf{e}_k$ & -- & Number of place-value slots of an embedding, channels per slot ($d=Kc$), embedding of the exponent of slot $k$ \\
		$\Aadd$, $\Amul$ & -- & Admissible pairs for addition and multiplication, Eq.~(\ref{eq:admissible}) \\
		$\varphi_{\Theta_E}$ & $\varphi$ & The function $\varphi_{\Theta_E}:\Dset \rightarrow \sR^d$ equivalent to the encoder neural network with trainable parameter $\Theta_E$ \\
		$\varphi_{\Theta_D}^\dagger$ & $\varphi^\dagger$ & The function $\varphi_{\Theta_D}^\dagger:\sR^d \rightarrow \Dset\cup\{\bot\}$ equivalent to the decoder neural network (with greedy decoding) with trainable parameter $\Theta_D$ \\
		$*_{\Theta}$ & $\dot{*}$ & A custom binary operator ($\sR^d \times \sR^d \rightarrow \sR^d$) with trainable parameter $\Theta$ \\
		$+$ & -- & The usual addition operator defined on rational numbers and real numbers \\
		$+_{\Theta}$ & $\dot{+}$ & A custom binary addition operator with trainable parameter $\Theta$ \\
		$\times$ & -- & The usual multiplication operator defined on rational numbers and real numbers \\
		$\times_{\Theta}$ & $\dot{\times}$ & A custom binary multiplication operator with trainable parameter $\Theta$ \\
		$\omega_\Theta$ & $\omega$ & The function $\omega_{\Theta}:\sR^d\times \sR^d \rightarrow \simplex$ equivalent to the neural comparator with trainable parameter $\Theta$ \\
		$\bar{\omega}$, $\mathbf{P}$, $\simplex$ & -- & Antisymmetrized comparator, Eq.~(\ref{eq:symmetrization}); $\mathbf{P}$ swaps the first two coordinates; $\simplex$ is the probability simplex in $\sR^3$ \\
		$<_{\omega_\Theta}$ & $<_{\omega}$ & The relation constructed using $\omega_\Theta$, Definition~\ref{def:induced_relation} \\
		$<$ & -- & The usual  order defined on  rational numbers and real numbers \\
		$\distP$ & -- & A probability distribution on $\Dset$ with $\distP(a)>0$ for all $a\in\Dset$ \\
		$H$ & -- & $H=\varphi(\Dset)$ \\
		$h$ or $h_i$ & -- & Each element of $H$ \\
		$\Delta_E$, $\Delta_{\Dset}$ & -- & Discrepancies on $\sR^d$ and on $\Dset\cup\{\bot\}$: $\Delta(x,y)\ge 0$, with equality iff $x=y$ (e.g., a metric) \\
		$\epsadd$, $\epsmul$ & -- & Additive and multiplicative defects, Definition~\ref{def:defects} \\
		$\rho(a)$, $\gamma(a)$ & -- & Decoding radius and decoding margin of $a$, Definition~\ref{def:radius} and Lemma~\ref{lem:margin} \\
		$F_H$, $\oplus$, $\otimes$, $\prec$ & -- & Field of fractions of $H$ and its operations and order, Corollary~\ref{cor:fraction_field} \\
		$L_{+}$, $L_{\times}$ & -- & Lipschitz constants of $\dot{+}$ and $\dot{\times}$ in each argument \\
		$(H, \dot{+}, \dot{\times}, <_{\omega})$ & $(H, \dot{+}, \dot{\times})$ & The neural ordered integral domain of Definition~\ref{def:neural_ordered_domain}\\
		$\mathbf{x}$ & -- & The representation of the number as a sequence of digits ($\mathbf{x}=(x_i)_{i=1}^n$); $n$ is its length\\
		$a, b, c$ & -- & Elements of $\Dset$ (numbers)\\
		$1_{\{true\}}(.)$ & $1(.)$ & Indicator function\\
		\bottomrule
	\end{tabular}
\end{table}

\subsection{Model architecture}\label{subsec:architecture}

The NIF model, similar to the BART model \citep{lewis-etal-2020-bart}, combines bidirectional and auto-regressive transformers to form an auto-encoder neural network. However, our model differs in two key ways: (1) it is not denoising, and (2) instead of generating a matrix, it outputs a single vector from the transformer encoder, which is then passed to the decoder. This vector serves as the embedding for the input sequence. Therefore, if $\mathbf{x}=(x_i)_{i=1}^n$ is the encoder's input number, the embedding vector $h$ for $\mathbf{x}$ is generated. Each number is encoded separately; embeddings of different numbers interact only inside the operators $\dot{+}$ and $\dot{\times}$, whose results are again vectors in $\sR^d$, and inside the comparator. All Transformer blocks use pre-normalization with RMSNorm \citep{zhang2019rmsnorm} and SwiGLU feed-forward layers \citep{shazeer2020glu}.

\paragraph{Place-value slots} Besides its token embedding, each digit token gets a learned embedding of its exponent, that is, of its place value (the digit $d$ in $\ldots+d\cdot10^{e}+\ldots$ has exponent $e$), as in Abacus embeddings \citep{mcleish2024transformers} and position coupling \citep{cho2024position}. The exponent of each token follows from its position relative to the decimal point and is computed when the string is tokenized. The embedding vector consists of $K$ \emph{place-value slots} with $c$ channels each, $z=(z_1,\dots,z_K)$ with $z_k\in\sR^{c}$ and $d=Kc$; slot $k$ stands for the exponent $e_k=e_{\min}+k-1$. The encoder fills the slots by cross-attention from $K$ learned slot queries to its output states, with an additive bias toward the digit token of exponent $e_k$, followed by two Transformer blocks over the slots; each slot is scaled to unit root mean square, which fixes the scale of the embeddings. The sign token and the decimal point share one extra exponent embedding (no place value); the slot queries attend to them without a bias, so the sign reaches the slots through this attention and the slot blocks. We use $e_{\min}=-26$ and $K=52$ (exponents $-26,\dots,25$) and $c=10$, so $d=520$. An element of $\mathbb{D}_{25}$ uses at most the integer places $0,\dots,24$ and the fractional places $-1,\dots,-24$; the other three slots are spare. Longer numbers need more slots ($K=2N+2$, Section~\ref{subsubsec:ablation}).

\paragraph{Decoder input and length} The decoder reads $z$ as $K$ slot vectors, each with the embedding $\mathbf{e}_k$ of its exponent (Figure~\ref{fig:nif_model}). A length head (attention pooling over the slots and two classifiers) predicts the number of integer and fractional digits (a number with $|a|<1$ has one integer digit, the $0$), and these fix the exponent of each output position. The decoder cross-attends to the slots with the same place-value bias and decodes greedily; at each position, only the token type (sign, digit or decimal point) allowed by the predicted length can be chosen. An output that is not a canonical string, for example one with a leading or trailing zero, counts as an error. The decoder never receives the length of the target, so $\varphi^\dagger(z)$ depends on $z$ alone.

For a vector $z\in\sR^d$, $\varphi^\dagger(z)\in\Dset\cup\{\bot\}$ denotes the decoded output; a canonical string is identified with the number it represents, and $\bot$ stands for an output that is not the canonical string of an element of $\Dset$.

We denote the encoder function as $\varphi$ and the decoder function as $\varphi^\dagger$. The objective is to construct the set $H = \varphi(\Dset)$ and define operations $+_{\Theta_1}$ and $\times_{\Theta_2}$ (abbreviated $\dot{+}$ and $\dot{\times}$, Table~\ref{tab:notations}) and a comparator $\omega$ such that $(H, +_{\Theta_1}, \times_{\Theta_2}, <_\omega)$ forms an ordered integral domain isomorphic to $(\Dset,+,\times,<)$, whose field of fractions is then an ordered field isomorphic to $\sQ$. For this construction to hold, the function $\varphi$ must act as an isomorphism. Section \ref{sec:ano} explores the construction of these operations, while Section \ref{sec:papdt} addresses the training of the NIF necessary for building the ordered structure. Furthermore, since the elements of $H$ are represented as vectors generated by a neural network, Section \ref{sec:neural_order} defines the concept of a neural order, which is essential for establishing the properties of the ordered structure. Section~\ref{sec:approximate} then examines how far a trained model with non-zero losses is from these ideal conditions.

\begin{figure}[htbp]
	\centering
	\scalebox{0.95}{

\begin{tikzpicture}[x=0.75pt,y=0.75pt,yscale=-1,xscale=1]
	
	\draw  [line width=1.5] [fill={rgb, 255:red, 219; green, 234; blue, 243 }  ,fill opacity=1 ] (177,75) -- (282,75) -- (282,228) -- (177,228) -- cycle ;
	\draw  [fill={rgb, 255:red, 10; green, 0; blue, 0 }  ,fill opacity=1 ] (187.8,241.7) -- (192.3,229.7) -- (196.8,241.7) -- (192.3,235.7) -- cycle ;
	\draw [line width=1.5]    (192.26,234.5) -- (192.26,260.36) ;
	\draw  [fill={rgb, 255:red, 10; green, 0; blue, 0 }  ,fill opacity=1 ] (207.8,241.74) -- (212.3,229.74) -- (216.8,241.74) -- (212.3,235.74) -- cycle ;
	\draw  [fill={rgb, 255:red, 10; green, 0; blue, 0 }  ,fill opacity=1 ] (262.37,241.67) -- (266.87,229.67) -- (271.37,241.67) -- (266.87,235.67) -- cycle ;
	\draw [line width=1.5]    (212.26,234.5) -- (212.26,260.36) ;
	\draw [line width=1.5]    (266.99,234.5) -- (266.99,260.36) ;
	\draw  [fill={rgb, 255:red, 10; green, 0; blue, 0 }  ,fill opacity=1 ] (242.9,241.7) -- (247.4,229.7) -- (251.9,241.7) -- (247.4,235.7) -- cycle ;
	\draw [line width=1.5]    (247.4,234.5) -- (247.4,260.36) ;
	\draw [color={rgb, 255:red, 0; green, 0; blue, 0 }  ,draw opacity=1 ][line width=1.5]    (282.43,124.1) -- (375.53,124.13) ;
	\draw [color={rgb, 255:red, 0; green, 0; blue, 0 }  ,draw opacity=1 ][line width=1.5]    (332.56,124.29) -- (332.6,180.38) ;
	\draw [color={rgb, 255:red, 0; green, 0; blue, 0 }  ,draw opacity=1 ][line width=1.5]    (331.6,179.89) -- (375.8,179.9) ;
	\draw [color={rgb, 255:red, 0; green, 0; blue, 0 }  ,draw opacity=1 ][line width=1.5]    (331.73,139.24) -- (376.76,139.2) ;
	\draw [color={rgb, 255:red, 0; green, 0; blue, 0 }  ,draw opacity=1 ][line width=1.5]    (332.21,165.31) -- (375.42,165.3) ;
	\draw  [color={rgb, 255:red, 0; green, 0; blue, 0 }  ,draw opacity=1 ][fill={rgb, 255:red, 0; green, 0; blue, 0 }  ,fill opacity=1 ] (368.05,119.57) -- (380.05,124.07) -- (368.05,128.57) -- (374.05,124.07) -- cycle ;
	\draw  [color={rgb, 255:red, 0; green, 0; blue, 0 }  ,draw opacity=1 ][fill={rgb, 255:red, 0; green, 0; blue, 0 }  ,fill opacity=1 ] (368,134.63) -- (380,139.13) -- (368,143.63) -- (374,139.13) -- cycle ;
	\draw  [color={rgb, 255:red, 0; green, 0; blue, 0 }  ,draw opacity=1 ][fill={rgb, 255:red, 0; green, 0; blue, 0 }  ,fill opacity=1 ] (368,160.75) -- (380,165.25) -- (368,169.75) -- (374,165.25) -- cycle ;
	\draw  [color={rgb, 255:red, 0; green, 0; blue, 0 }  ,draw opacity=1 ][fill={rgb, 255:red, 0; green, 0; blue, 0 }  ,fill opacity=1 ] (368,175.42) -- (380,179.92) -- (368,184.42) -- (374,179.92) -- cycle ;
	\draw [color={rgb, 255:red, 0; green, 0; blue, 0 }  ,draw opacity=1 ][line width=1.5]    (282.47,139.09) -- (292.4,139.09) -- (302.65,139.1) ;
	\draw  [color={rgb, 255:red, 0; green, 0; blue, 0 }  ,draw opacity=1 ][fill={rgb, 255:red, 0; green, 0; blue, 0 }  ,fill opacity=1 ] (296.9,134.63) -- (308.9,139.13) -- (296.9,143.63) -- (302.9,139.13) -- cycle ;
	\draw [color={rgb, 255:red, 0; green, 0; blue, 0 }  ,draw opacity=1 ][line width=1.5]    (282.18,164.8) -- (292.11,164.81) -- (302.37,164.81) ;
	\draw  [color={rgb, 255:red, 0; green, 0; blue, 0 }  ,draw opacity=1 ][fill={rgb, 255:red, 0; green, 0; blue, 0 }  ,fill opacity=1 ] (296.9,160.35) -- (308.9,164.85) -- (296.9,169.35) -- (302.9,164.85) -- cycle ;
	\draw [color={rgb, 255:red, 0; green, 0; blue, 0 }  ,draw opacity=1 ][line width=1.5]    (282.42,179.87) -- (292.35,179.88) -- (302.6,179.89) ;
	\draw  [color={rgb, 255:red, 0; green, 0; blue, 0 }  ,draw opacity=1 ][fill={rgb, 255:red, 0; green, 0; blue, 0 }  ,fill opacity=1 ] (296.9,175.29) -- (308.9,179.79) -- (296.9,184.3) -- (302.9,179.79) -- cycle ;
	\draw  [line width=1.5] [color={rgb, 255:red, 0; green, 0; blue, 0 }  ,draw opacity=1 ][fill={rgb, 255:red, 244; green, 211; blue, 201 }  ,fill opacity=1 ] (382,75) -- (508.4,75) -- (508.4,228) -- (382,228) -- cycle ;
	\draw  [fill={rgb, 255:red, 10; green, 0; blue, 0 }  ,fill opacity=1 ] (403.8,241.7) -- (408.3,229.7) -- (412.8,241.7) -- (408.3,235.7) -- cycle ;
	\draw [line width=1.5]    (408.26,234.5) -- (408.26,260.36) ;
	\draw  [fill={rgb, 255:red, 10; green, 0; blue, 0 }  ,fill opacity=1 ] (423.8,241.74) -- (428.3,229.74) -- (432.8,241.74) -- (428.3,235.74) -- cycle ;
	\draw  [fill={rgb, 255:red, 10; green, 0; blue, 0 }  ,fill opacity=1 ] (478.37,241.67) -- (482.87,229.67) -- (487.37,241.67) -- (482.87,235.67) -- cycle ;
	\draw [line width=1.5]    (428.26,234.5) -- (428.26,260.36) ;
	\draw [line width=1.5]    (482.99,234.5) -- (482.99,260.36) ;
	\draw  [fill={rgb, 255:red, 10; green, 0; blue, 0 }  ,fill opacity=1 ] (458.9,241.7) -- (463.4,229.7) -- (467.9,241.7) -- (463.4,235.7) -- cycle ;
	\draw [line width=1.5]    (463.4,234.5) -- (463.4,260.36) ;
	
	\draw  [fill={rgb, 255:red, 10; green, 0; blue, 0 }  ,fill opacity=1 ] (403.8,56.2) -- (408.3,44.2) -- (412.8,56.2) -- (408.3,50.2) -- cycle ;
	\draw [line width=1.5]    (408.26,49) -- (408.26,74.86) ;
	\draw  [fill={rgb, 255:red, 10; green, 0; blue, 0 }  ,fill opacity=1 ] (423.8,56.24) -- (428.3,44.24) -- (432.8,56.24) -- (428.3,50.24) -- cycle ;
	\draw  [fill={rgb, 255:red, 10; green, 0; blue, 0 }  ,fill opacity=1 ] (478.37,56.17) -- (482.87,44.17) -- (487.37,56.17) -- (482.87,50.17) -- cycle ;
	\draw [line width=1.5]    (428.26,49) -- (428.26,74.86) ;
	\draw [line width=1.5]    (482.99,49) -- (482.99,74.86) ;
	\draw  [fill={rgb, 255:red, 10; green, 0; blue, 0 }  ,fill opacity=1 ] (458.9,56.2) -- (463.4,44.2) -- (467.9,56.2) -- (463.4,50.2) -- cycle ;
	\draw [line width=1.5]    (463.4,49) -- (463.4,74.86) ;

	\draw (478,31.5) node [anchor=north west][inner sep=0.75pt]   [align=left] {{\fontfamily{ptm}\selectfont {\scriptsize $\displaystyle \hat{x}_n$}}};
	\draw (450.67,31.5) node [anchor=north west][inner sep=0.75pt]   [align=left] {{\fontfamily{ptm}\selectfont {\scriptsize $\displaystyle \hat{x}_{n-1}$}}};
	\draw (423,31.5) node [anchor=north west][inner sep=0.75pt]   [align=left] {{\scriptsize {\fontfamily{ptm}\selectfont $\displaystyle \hat{x}_2$}}};
	\draw (403,31.5) node [anchor=north west][inner sep=0.75pt]   [align=left] {{\scriptsize {\fontfamily{ptm}\selectfont $\displaystyle \hat{x}_1$}}};
	\draw (395,137) node [anchor=north west][inner sep=0.75pt]   [align=left] {\begin{minipage}[lt]{70.75pt}\setlength\topsep{0pt}
			\begin{center}
				Autoregressive\\Decoder
			\end{center}
			
	\end{minipage}};
	\draw (438.31,237.89) node [anchor=north west][inner sep=0.75pt]   [align=left] {...};
	\draw (438.31,52.39) node [anchor=north west][inner sep=0.75pt]   [align=left] {...};
	\draw (190,137) node [anchor=north west][inner sep=0.75pt]   [align=left] {\begin{minipage}[lt]{55.97pt}\setlength\topsep{0pt}
			\begin{center}
				{\fontfamily{ptm}\selectfont Bidirectional}\\{\fontfamily{ptm}\selectfont Encoder}
			\end{center}
			
	\end{minipage}};
	\draw (222.31,237.89) node [anchor=north west][inner sep=0.75pt]   [align=left] {...};
	\draw (187,262) node [anchor=north west][inner sep=0.75pt]   [align=left] {{\footnotesize {\fontfamily{ptm}\selectfont $\displaystyle s$}}};
	\draw (207,262) node [anchor=north west][inner sep=0.75pt]   [align=left] {{\scriptsize {\fontfamily{ptm}\selectfont $\displaystyle x_1$}}};
	\draw (235.67,262) node [anchor=north west][inner sep=0.75pt]   [align=left] {{\fontfamily{ptm}\selectfont {\scriptsize $\displaystyle x_{n-1}$}}};
	\draw (262,262) node [anchor=north west][inner sep=0.75pt]   [align=left] {{\fontfamily{ptm}\selectfont {\scriptsize $\displaystyle x_n$}}};
	\draw (303.7,129) node [anchor=north west][inner sep=0.75pt]   [align=left] {{\tiny $\displaystyle z_2$}};
	\draw (300.7,153.87) node [anchor=north west][inner sep=0.75pt]   [align=left] {{\tiny $\displaystyle z_{K-1}$}};
	
	\draw (289.7,145.87) node [anchor=north west][inner sep=0.75pt] [rotate=-90]  [align=left] {...};
	
	\draw (345.7,142.87) node [anchor=north west][inner sep=0.75pt]  [rotate=-90]  [align=left] {...};
	
	\draw (303.7,169.64) node [anchor=north west][inner sep=0.75pt]   [align=left] {{\tiny $\displaystyle z_K$}};
	\draw (303.7,112.36) node [anchor=north west][inner sep=0.75pt]   [align=left] {{\tiny $\displaystyle z_1$}};
	\draw (402,263.5) node [anchor=north west][inner sep=0.75pt]   [align=left] {{\footnotesize $\displaystyle s$}};
	\draw (423,262) node [anchor=north west][inner sep=0.75pt]   [align=left] {{\scriptsize {\fontfamily{ptm}\selectfont $\displaystyle \hat{x}_1$}}};
	\draw (450.67,262) node [anchor=north west][inner sep=0.75pt]   [align=left] {{\fontfamily{ptm}\selectfont {\scriptsize $\displaystyle \hat{x}_{n-2}$}}};
	\draw (478,262) node [anchor=north west][inner sep=0.75pt]   [align=left] {{\fontfamily{ptm}\selectfont {\scriptsize $\displaystyle \hat{x}_{n-1}$}}};
	\draw (333,112.4) node [anchor=north west][inner sep=0.75pt]   [align=left] {{\tiny $\displaystyle +\mathbf{e}_1$}};
	\draw (333,129.43) node [anchor=north west][inner sep=0.75pt]   [align=left] {{\tiny $\displaystyle +\mathbf{e}_2$}};
	\draw (333,155.03) node [anchor=north west][inner sep=0.75pt]   [align=left] {{\tiny $\displaystyle +\mathbf{e}_{K-1}$}};
	\draw (333,169.6) node [anchor=north west][inner sep=0.75pt]   [align=left] {{\tiny $\displaystyle +\mathbf{e}_K$}};

\end{tikzpicture}

}
	\caption{Architecture of the NIF model. The embedding is read as $K$ place-value slots $z_1,\dots,z_K$; each slot gets the embedding $\mathbf{e}_k$ of its exponent before it enters the decoder.}
	\label{fig:nif_model}
\end{figure}

\subsection{Abelian neural operator} \label{sec:ano}
An Abelian Neural Operator (ANO) is a binary operator that, first, is implemented as a neural network, and second, exhibits commutative property. This operator can be defined as $*_\Theta: \sR^d \times \sR^d \rightarrow \sR^d$. Ideally, it maps $H\times H$ into $H=\varphi(\Dset)$, so that $*_\Theta$ is closed on $H$.

\subsubsection{Abelian concatenation layer}
The Abelian Concatenation (AC) layer is an intermediate layer designed to enable our model to learn complex addition and multiplication operators. In this layer, input vectors must be combined in a manner that preserves the commutative property. In the DeepSets paper \citep{zaheer2017deepset}, it is demonstrated that under certain conditions, a neural network remains invariant under various permutations of a set input, inspiring the development of a layer designed to preserve the commutative property. To achieve such a layer, consider two input vectors \( h_1, h_2 \in \mathbb{R}^d \). These vectors are concatenated to form two new vectors: \( [h_1, h_2] \) and \( [h_2, h_1] \). A nonlinear function (equivalent to a nonlinear neural network) is then applied to these concatenated vectors to produce two output vectors \( t_1 \) and \( t_2 \), as described in Equations (\ref{eq:ab_concat1}) and (\ref{eq:ab_concat2}).

\begin{align}
	t_1 = f([h_1,h_2];\theta)\label{eq:ab_concat1} \\
	t_2 = f([h_2,h_1];\theta)\label{eq:ab_concat2}
\end{align}

where $\displaystyle [.,.]$ is the concatenation operator and $\displaystyle f:\sR^{2d} \rightarrow \sR^d$ is a nonlinear function equivalent to a neural network with trainable parameters $\displaystyle \theta$, e.g., a dense layer with ReLU activation function. In our implementation, $f$ is a small Transformer over slots. The two halves of $[h_1,h_2]$ become $2K$ slot tokens (with embeddings of the slot and of the half) and pass through one or two pre-norm blocks with rotary position embeddings over the slot index \citep{su2024roformer} and a SwiGLU MLP that has extra periodic units $\sin(Wx)$, $\cos(Wx)$, as in Fourier Analysis Networks \citep{dong2025fan}. The first $K$ tokens are mapped back to $\sR^d$ with a residual connection, $f([h_1,h_2];\theta)=h_1+g([h_1,h_2];\theta)$. The dense layer $\sigma(W[h_1,h_2]+b)$ is tested as an ablation (Section~\ref{subsubsec:ablation}).

\begin{figure}[htbp]
	\centering
	\scalebox{.7}{
\begin{tikzpicture}[x=0.75pt,y=0.75pt,yscale=-1,xscale=1]
	
	\draw  [dash pattern={on 5.63pt off 4.5pt}][line width=1.5]  (104,86) -- (507.87,86) -- (507.87,240) -- (104,240) -- cycle ;
	
	\draw  [color={rgb, 255:red, 29; green, 89; blue, 183 }  ,draw opacity=1 ][line width=1.5]  (169.4,230.5) .. controls (152.56,230.5) and (138.9,216.84) .. (138.9,200) -- (138.9,123) .. controls (138.9,106.16) and (152.56,92.5) .. (169.4,92.5) -- (169.4,92.5) .. controls (186.24,92.5) and (199.9,106.16) .. (199.9,123) -- (199.9,200) .. controls (199.9,216.84) and (186.24,230.5) .. (169.4,230.5) -- cycle ;
	\draw [line width=1.5]    (200.8,137.5) -- (256,137.5) ;
	\draw [shift={(257.2,137.5)}, rotate = 180] [color={rgb, 255:red, 0; green, 0; blue, 0 }  ][line width=1.5]    (14.21,-4.28) .. controls (9.04,-1.82) and (4.3,-0.39) .. (0,0) .. controls (4.3,0.39) and (9.04,1.82) .. (14.21,4.28)   ;
	\draw [line width=1.5]    (200.8,191.4) -- (256,191.4) ;
	\draw [shift={(257.2,191.4)}, rotate = 180] [color={rgb, 255:red, 0; green, 0; blue, 0 }  ][line width=1.5]    (14.21,-4.28) .. controls (9.04,-1.82) and (4.3,-0.39) .. (0,0) .. controls (4.3,0.39) and (9.04,1.82) .. (14.21,4.28)   ;
	\draw  [dash pattern={on 0.84pt off 2.51pt}]  (272.1,137.5) -- (284.6,137.5) ;
	\draw  [dash pattern={on 0.84pt off 2.51pt}]  (272.1,191.4) -- (284.6,191.4) ;
	\draw [line width=1.5]    (290,137.5) -- (345,137.5) ;
	\draw [shift={(346.2,137.5)}, rotate = 180] [color={rgb, 255:red, 0; green, 0; blue, 0 }  ][line width=1.5]    (14.21,-4.28) .. controls (9.04,-1.82) and (4.3,-0.39) .. (0,0) .. controls (4.3,0.39) and (9.04,1.82) .. (14.21,4.28)   ;
	\draw [line width=1.5]    (290,191.4) -- (345,191.4) ;
	\draw [shift={(346.2,191.4)}, rotate = 180] [color={rgb, 255:red, 0; green, 0; blue, 0 }  ][line width=1.5]    (14.21,-4.28) .. controls (9.04,-1.82) and (4.3,-0.39) .. (0,0) .. controls (4.3,0.39) and (9.04,1.82) .. (14.21,4.28)   ;
	\draw  [color={rgb, 255:red, 20; green, 19; blue, 19 }  ,draw opacity=1 ] (479.5,165.22) .. controls (479.5,160.38) and (483.43,156.45) .. (488.27,156.45) .. controls (493.12,156.45) and (497.05,160.38) .. (497.05,165.22) .. controls (497.05,170.07) and (493.12,174) .. (488.27,174) .. controls (483.43,174) and (479.5,170.07) .. (479.5,165.22) -- cycle ;
	\draw [line width=1.5]    (497.05,165.22) -- (529.5,165.57) ;
	\draw [shift={(532.5,165.6)}, rotate = 180.61] [color={rgb, 255:red, 0; green, 0; blue, 0 }  ][line width=1.5]    (14.21,-4.28) .. controls (9.04,-1.82) and (4.3,-0.39) .. (0,0) .. controls (4.3,0.39) and (9.04,1.82) .. (14.21,4.28)   ;
	\draw [line width=1.5]    (418,137.5) -- (489.3,137.5) ;
	\draw [line width=1.5]    (418,191.4) -- (489.3,191.4) ;
	\draw [line width=1.5]    (488.3,137.4) -- (488.27,156.45) ;
	\draw [line width=1.5]    (488.27,191.5) -- (488.27,174) ;
	\draw [line width=1.5]    (72,137.5) -- (127,137.5) ;
	\draw [shift={(128.2,137.5)}, rotate = 180] [color={rgb, 255:red, 0; green, 0; blue, 0 }  ][line width=1.5]    (14.21,-4.28) .. controls (9.04,-1.82) and (4.3,-0.39) .. (0,0) .. controls (4.3,0.39) and (9.04,1.82) .. (14.21,4.28)   ;
	\draw [line width=1.5]    (71.99,191.4) -- (126.99,191.4) ;
	\draw [shift={(128.2,191.4)}, rotate = 180] [color={rgb, 255:red, 0; green, 0; blue, 0 }  ][line width=1.5]    (14.21,-4.28) .. controls (9.04,-1.82) and (4.3,-0.39) .. (0,0) .. controls (4.3,0.39) and (9.04,1.82) .. (14.21,4.28)   ;
	\draw  [color={rgb, 255:red, 29; green, 89; blue, 183 }  ,draw opacity=1 ][line width=1.5]  (387.67,230.5) .. controls (370.82,230.5) and (357.17,216.84) .. (357.17,200) -- (357.17,123) .. controls (357.17,106.16) and (370.82,92.5) .. (387.67,92.5) -- (387.67,92.5) .. controls (404.51,92.5) and (418.17,106.16) .. (418.17,123) -- (418.17,200) .. controls (418.17,216.84) and (404.51,230.5) .. (387.67,230.5) -- cycle ;
	
	\draw (489.0,165.4) node    {$+$};
	\draw (305.83,118) node [anchor=north west][inner sep=0.75pt]    {$t_{1}^{k-1}$};
	\draw (305.8,171) node [anchor=north west][inner sep=0.75pt]    {$t_{2}^{k-1}$};
	\draw (530,145.87) node [anchor=north west][inner sep=0.75pt]    {$h_1*_{\Theta} \ h_2$};
	\draw (445.83,119) node [anchor=north west][inner sep=0.75pt]    {$t_{1}^{k}$};
	\draw (445.8,172) node [anchor=north west][inner sep=0.75pt]    {$t_{2}^{k}$};
	\draw (215.8,119) node [anchor=north west][inner sep=0.75pt]    {$t_{1}^{1}$};
	\draw (215.83,172) node [anchor=north west][inner sep=0.75pt]    {$t_{2}^{1}$};
	\draw (75.53,123) node [anchor=north west][inner sep=0.75pt]    {$h_{1}$};
	\draw (75.53,176) node [anchor=north west][inner sep=0.75pt]    {$h_{2}$};
	\draw (170.97,161.5) node  [rotate=-270] [align=left] {AC \ Layer};
	\draw (387.64,162.5) node  [rotate=-270] [align=left] {AC \ Layer};

\end{tikzpicture}}
	\caption{Stacked Abelian Neural Operator.}
	\label{fig:SAC_layer}
\end{figure}

\subsubsection{Stacked AC layers}
Multiple AC layers can be stacked to capture more complex operators, as illustrated in Figure \ref{fig:SAC_layer}. This stacking is feasible because the input vector dimensions to a layer align with the output vector dimensions from that layer. After stacking several layers, the final output consists of two vectors. The sum of these two vectors yields the final vector, and this addition naturally maintains the commutative property (Lemma~\ref{lem:commutativity}).

Now consider a sequence of \( k \) AC layers, where for each layer \( i \) (\(1 \leq i \leq k\)), the inputs are \( t_1^{i-1} \) and \( t_2^{i-1} \), the outputs are \( t_1^{i} \) and \( t_2^{i} \), and the corresponding trainable parameter is denoted by \( \theta_i \). Let \( \Theta = \{\theta_1, \dots, \theta_k\} \) represent the collection of all trainable parameters across all \( k \) layers.  For \( h_1, h_2 \in \mathbb{R}^d \), the operation \( h_1 *_{\Theta} h_2 \) is defined as follows:

\begin{align}
	h_1*_{\Theta} \ h_2 = t_1^{k}+t_2^{k} =f([t_1^{k-1},t_2^{k-1}];\theta_k) + f([t_2^{k-1},t_1^{k-1}];\theta_k)\label{eq:neural_op}
\end{align}

where $\displaystyle t_1^0 = h_1, t_2^0 = h_2$ and for $\displaystyle 1\leq i\leq k$,

\begin{align}
	t_1^i = f([t_1^{i-1}, t_2^{i-1}];\theta_i) \label{eq:ac_rec1}\\
	t_2^i = f([t_2^{i-1}, t_1^{i-1}];\theta_i) \label{eq:ac_rec2}
\end{align}

\begin{lemma}[Commutativity by construction]\label{lem:commutativity}
	For every $k\ge 1$, all parameters $\Theta$ and all $h_1,h_2\in\sR^d$, $h_1 *_\Theta h_2 = h_2 *_\Theta h_1$.
\end{lemma}
\noindent Commutativity thus holds exactly and need not be learned. The other field axioms (associativity, identities, inverses and distributivity) are only encouraged by the training objective.

\subsubsection{Staged multiplication operator}\label{sec:staged}
In a product, each output digit depends on all digit pairs $(a_i,b_j)$ with $i+j$ up to its exponent, and the column sums $\sum_{i+j=k}a_ib_j$ reach several hundred. In our experiments, a stack of generic AC layers did not learn the resulting carries (Section~\ref{subsubsec:ablation}). We therefore match the AC layers of $\dot{\times}$ to a split of the product into steps whose values are no larger than those of an addition. For $a,b\in\Dset$ with digits $a_i$ and $b_j$ (the digits of exponents $i$ and $j$) and an output exponent $k$ (in this section, $k$ indexes exponents, not AC layers), let
\begin{equation}\label{eq:LH}
	L_k:=\sum_{i+j=k}(a_ib_j\bmod 10),\qquad H_k:=\sum_{i+j=k}\Big\lfloor\frac{a_ib_j}{10}\Big\rfloor,
\end{equation}
let $u^{(m)}$ denote the $m$-th decimal digit of a non-negative integer $u$, and define
\begin{equation}\label{eq:ST}
	S_k:=L_k^{(0)}+L_{k-1}^{(1)}+L_{k-2}^{(2)}+H_{k-1}^{(0)}+H_{k-2}^{(1)}+H_{k-3}^{(2)},\qquad
	T_k:=(S_k\bmod 10)+\lfloor S_{k-1}/10\rfloor .
\end{equation}

\begin{lemma}[Small-sum decomposition of the product]\label{lem:small_sum}
	Let $a,b\in\Dset$ be such that the shorter of the two canonical strings has $m\le 25$ digits. Then $L_k\le 9m$ and $H_k\le 8m$ have at most three digits, $0\le S_k\le 40$, $0\le T_k\le 13$, and $|ab|=\sum_k T_k10^{k}$. Moreover, if the digits of $|ab|$ are obtained from $(T_k)$ by a carry pass from the least significant exponent upwards, $d_k=(T_k+\kappa_k)\bmod 10$ and $\kappa_{k+1}=\lfloor(T_k+\kappa_k)/10\rfloor$ with $\kappa_{k_0}=0$ at the lowest exponent $k_0$, then every carry satisfies $\kappa_k\in\{0,1\}$.
\end{lemma}

The lemma covers every admissible product, since both operands have at most $N=25$ digits; for $N\le 200$, the same argument holds with four-digit $L_k$ and $H_k$, and $S_k$ then has two further terms (Appendix~\ref{app:proof_small_sum}). The operator $\dot{\times}$ has four AC layers, one per step: the first produces the digits of $L_k$ and $H_k$, the second $S_k\bmod 10$ and $\lfloor S_k/10\rfloor$, and the last two form $T_k$ and perform the carry pass, which has the form of an addition. The intermediate states have $c_{\mathrm{int}}=32$ channels per slot (channels $10\to32\to32\to32\to10$, with residual connections in the middle two layers) and are computed in single precision (bfloat16 represents integers exactly only up to $256$, and $L_k\le 9N$ exceeds this for $N>28$). The operator has the structure of Eq.~(\ref{eq:neural_op}), with $h_1\dot{\times}h_2=t_1^4+t_2^4$, so Lemma~\ref{lem:commutativity} applies (its proof does not use the number of channels). Like the other operators, it receives only $\varphi(a)$ and $\varphi(b)$, never the digit strings, and returns a vector in $\sR^d$. Column sums are formed only at the $K$ slot places, so digit products below the lowest slot ($10^{-(N+1)}$) are cut off; among admissible pairs, this affects only products with trailing zeros whose operands have more than $N+1$ fractional digits together, such as some pairs $x\cdot x^{-1}$ (Section~\ref{sec:expriments}). In Eq.~(\ref{eq:ST}), terms whose exponent lies outside the slot window are zero. The signs of the operands are read by a sign classifier on each slot, whose logits are averaged over the slots and enter the first layer through $\tanh$; the signs are hint targets. The first layer starts with three components:
\begin{enumerate}
	\item a learned classifier that reads the digit of each slot of both inputs from the $c$ channels of the slot. Its prediction becomes a one-hot vector (arg max), so the counts below are exact; no gradient flows from the counts to the classifier, which is trained by the hint terms of Remark~\ref{rem:biases}. The loss of the product reaches the encoder through the slot states, which also enter the first layer directly, and through the soft digit probabilities and the sign features;
	\item the place-coupled product of the two digit indicators, which gives for each output exponent $k$ the number $n_k(p,q)$ of digit pairs $(p,q)$ with $i+j=k$. This is the multiplicative counterpart of position coupling (places $i$ and $j$ meet at $i+j$) and is computed with a fast Fourier transform along the slots;
	\item a linear map from the counts $n_k(\cdot,\cdot)$ to $(L_k,H_k)$, fitted in closed form by least squares rather than by gradient steps (Appendix~\ref{app:implementation}), whose exact solution is the table $(p,q)\mapsto(pq\bmod 10,\lfloor pq/10\rfloor)$, together with periodic features $\cos(\omega_m x)$, $\sin(\omega_m x)$ of $x\in\{L_k,H_k\}$. Their learnable frequencies start at $\omega_m=2\pi/10^m$, $m=1,2,3$, so the decimal digits of $L_k$ and $H_k$ can be read linearly (the decimal prior of FoNE \citep{zhou2025foneprecisesingletokennumber}).
\end{enumerate}
In $f([h_1,h_2];\theta_1)$, the output of these components for the ordered pair $(h_1,h_2)$ is added to the slot tokens of $h_1$, so the first layer keeps the form of Eqs.~(\ref{eq:ab_concat1})--(\ref{eq:ab_concat2}).

\begin{remark}[Inductive biases of $\dot{\times}$]\label{rem:biases}
	The fixed connectivity $(i,j)\mapsto i+j$, the decimal initial frequencies and the exact one-hot digits encode the place-value structure of decimal numbers but compute no result. The strongest bias is that the first layer reads the operand digits from the embeddings, which is possible only because the encoder arranges the embedding in place-value slots (Section~\ref{subsec:architecture}). The digit classifier, the digit readers and the carry logic are learned by gradient descent, and the map from the counts to $(L_k,H_k)$ is fitted by least squares. During training, auxiliary heads read the intermediate slot states and are supervised with the quantities of Lemma~\ref{lem:small_sum} (hint supervision, as in neural algorithmic reasoning \citep{velickovic2022clrs}); they are not used at test time. Because a structured operator could take over part of the algebraic structure from the embedding geometry (Remark~\ref{rem:transport}), the pairs $(x,x^{-1})$ of the multiplicative invertibility test are kept out of training, all other test tuples are drawn independently of the training tuples (long operands practically never repeat), and Sections~\ref{sec:expriments} and~\ref{subsubsec:ablation} report results by number length and the contribution of each component. The operator needs a clean code of the digit in every slot. Fixed one-hot digit slots give such a code, and the operator then reaches 99.95\% (Table~\ref{tab:abl_operator}). An encoder trained for reconstruction only does not, and the same operator and supervision reach only 29.00--43.66\% exact match on products (Section~\ref{sec:comparison}). The one-hot counts, the least-squares fit and the decimal frequencies only speed up training (Table~\ref{tab:abl_operator}).
\end{remark}

\subsection{Neural order
} \label{sec:neural_order}

In this section, we define neural ordering, which extends the definition of order on sets. The key difference is that, in this definition, the ordering is represented by a neural network. Specifically, the neural network ($\omega$) receives two embedding vectors as inputs and produces one of three possible states representing the relationship between these two vectors.

Let $\simplex := \{w\in[0,1]^3 : w_1+w_2+w_3=1\}$ and let $\omega:\sR^d\times\sR^d\rightarrow\simplex$ be a neural network (a \emph{comparator}) whose three outputs are the scores of the statements ``the first argument is smaller'', ``the first argument is larger'' and ``the arguments are equal''. Let $\mathbf{P}$ be the $3\times 3$ permutation matrix that swaps the first two coordinates.

\begin{definition}[Antisymmetric comparator]\label{def:antisymmetric}
	A comparator $\omega$ is \emph{antisymmetric} if $\omega(h_2,h_1)=\mathbf{P}\,\omega(h_1,h_2)$ for all $h_1,h_2\in\sR^d$.
\end{definition}

Every comparator can be made antisymmetric at the cost of a second forward pass:
\begin{equation}\label{eq:symmetrization}
	\bar{\omega}(h_1,h_2) := \tfrac{1}{2}\big(\omega(h_1,h_2) + \mathbf{P}\,\omega(h_2,h_1)\big).
\end{equation}
Since $\mathbf{P}^2=\mathbf{I}$, the map $\bar\omega$ takes values in $\simplex$, is antisymmetric, and equals $\omega$ if $\omega$ is already antisymmetric. We use $\bar\omega$ in training and in evaluation, and report the consistency of the two readings of the raw comparator $\omega$ as a separate test (Section~\ref{sec:evaluation_criteria}).

\begin{definition}[Relation induced by a comparator]\label{def:induced_relation}
	For a comparator $\omega$ and $h_1,h_2\in\sR^d$, write $\omega=(\omega_1,\omega_2,\omega_3)$ and define
	\begin{equation}\label{eq:induced_relation}
		h_1 <_{\omega} h_2 \;:\Longleftrightarrow\; \omega_1(h_1,h_2) > \max\{\omega_2(h_1,h_2),\, \omega_3(h_1,h_2)\}.
	\end{equation}
	We write $h_1 \approx_\omega h_2$ if neither $h_1<_\omega h_2$ nor $h_2<_\omega h_1$.
\end{definition}

In general, $h_1\approx_\omega h_2$ does not imply $h_1=h_2$.

\begin{definition}[Neural order]\label{def:neural_order}
	Let $S\subseteq\sR^d$. The relation $<_\omega$ is a \emph{neural order on $S$} if its restriction to $S$ is an order in the sense of \citet[Definition~1.5]{rudin1976principles}, i.e., for all $h_1,h_2,h_3\in S$:
	\begin{enumerate}
		\item[(N1)] (trichotomy) exactly one of $h_1<_\omega h_2$, $h_1=h_2$, $h_2<_\omega h_1$ holds;
		\item[(N2)] (transitivity) $h_1<_\omega h_2$ and $h_2<_\omega h_3$ imply $h_1<_\omega h_3$.
	\end{enumerate}
\end{definition}

The word \emph{neural} only indicates that the relation is computed by the comparator; the order itself is the usual one.

\begin{proposition}[Properties guaranteed by antisymmetry]\label{prop:antisymmetric}
	Let $\omega$ be an antisymmetric comparator. Then, for all $h_1,h_2\in\sR^d$:
	\begin{enumerate}
		\item[(a)] $h_1<_\omega h_1$ does not hold (irreflexivity);
		\item[(b)] $h_1<_\omega h_2$ implies that $h_2<_\omega h_1$ does not hold (asymmetry);
		\item[(c)] $h_2<_\omega h_1$ holds if and only if $\omega_2(h_1,h_2)>\max\{\omega_1(h_1,h_2),\omega_3(h_1,h_2)\}$, so the two possible readings of a pair coincide.
	\end{enumerate}
	Consequently, $<_\omega$ is a neural order on $S\subseteq\sR^d$ if and only if it is transitive on $S$ and connected on $S$, i.e., $h_1<_\omega h_2$ or $h_2<_\omega h_1$ for all distinct $h_1,h_2\in S$.
\end{proposition}

By Proposition~\ref{prop:antisymmetric}, irreflexivity and asymmetry hold by construction for an antisymmetric comparator, and only transitivity and connectedness must be learned. An order on a set is not enough for an ordered ring or field, however; the order must also agree with the operations.

\begin{definition}[Ordered integral domain, ordered field]\label{def:ordered_domain}
	Let $R$ be a commutative ring with identity $1\neq 0$ and without zero divisors (an integral domain), equipped with an order $<$ on the set $R$. Then $(R,+,\times,<)$ is an \emph{ordered integral domain} if, for all $x,y,z\in R$,
	\begin{enumerate}
		\item[(O1)] $y<z$ implies $x+y<x+z$;
		\item[(O2)] $0<x$ and $0<y$ imply $0<xy$.
	\end{enumerate}
	If, in addition, $R$ is a field, then $(R,+,\times,<)$ is an \emph{ordered field} \citep[Definition~1.17]{rudin1976principles}.
\end{definition}

\begin{definition}[Neural ordered integral domain and field]\label{def:neural_ordered_domain}
	Let $H\subseteq\sR^d$, let $\dot{+},\dot{\times}:\sR^d\times\sR^d\rightarrow\sR^d$ be operators that are closed on $H$, and let $\omega$ be a comparator such that $<_\omega$ is a neural order on $H$. The tuple $(H,\dot{+},\dot{\times},<_\omega)$ is a \emph{neural ordered integral domain} (respectively, a \emph{neural ordered field}) if, with the operations and the order restricted to $H$, it is an ordered integral domain (respectively, an ordered field) in the sense of Definition~\ref{def:ordered_domain}.
\end{definition}

\subsection{Preserving algebraic properties during training} \label{sec:papdt}
In this section, we examine how $H$ can be transformed into an ordered integral domain isomorphic to $\Dset$, and its field of fractions into an ordered field isomorphic to $\sQ$. We examine the conditions necessary for $H$ to become an ordered integral domain. Proposition~\ref{prop:transport} demonstrates the ideal conditions for transferring an ordered algebraic structure. The proof of this proposition and of all other results can be found in Appendix~\ref{appendix:theorems}.

\begin{figure}[htbp]

	\centering
	\resizebox{0.92\textwidth}{!}{

\tikzset{every picture/.style={line width=0.75pt}} %

\begin{tikzpicture}[x=0.75pt,y=0.75pt,yscale=-1,xscale=1]
	
	\draw  [fill={rgb, 255:red, 219; green, 234; blue, 243 }  ,fill opacity=1 ] (149.95,35.04) -- (149.9,50.29) -- (149.9,63.77) -- (100.2,79.79) -- (100.2,19.04) -- cycle ;
	\draw  [fill={rgb, 255:red, 244; green, 211; blue, 201 }  ,fill opacity=1 ] (346.95,19) -- (346.9,50.29) -- (346.9,79.8) -- (297.2,63.8) -- (297.2,35) -- cycle ;
	\draw  [color={rgb, 255:red, 0; green, 0; blue, 0 }  ,draw opacity=1 ][fill={rgb, 255:red, 0; green, 0; blue, 0 }  ,fill opacity=1 ] (383.02,46.82) -- (388.9,49.24) -- (383.02,51.66) -- (384.2,49.24) -- cycle ;
	\draw [line width=1.5]    (384.27,49.36) -- (346.87,49.33) ;

	\draw  [fill={rgb, 255:red, 219; green, 234; blue, 243 }  ,fill opacity=1 ] (149.95,318.04) -- (149.9,333.29) -- (149.9,346.77) -- (100.2,362.79) -- (100.2,302.04) -- cycle ;
	\draw  [fill={rgb, 255:red, 244; green, 211; blue, 201 }  ,fill opacity=1 ] (346.95,302) -- (346.9,333.29) -- (346.9,362.8) -- (297.2,346.8) -- (297.2,318) -- cycle ;
	\draw  [color={rgb, 255:red, 0; green, 0; blue, 0 }  ,draw opacity=1 ][fill={rgb, 255:red, 0; green, 0; blue, 0 }  ,fill opacity=1 ] (383.02,329.82) -- (388.9,332.24) -- (383.02,334.66) -- (384.2,332.24) -- cycle ;
	\draw [line width=1.5]    (384.27,332.36) -- (346.87,332.33) ;

	\draw  [fill={rgb, 255:red, 219; green, 234; blue, 243 }  ,fill opacity=1 ] (149.95,390.04) -- (149.9,405.29) -- (149.9,418.77) -- (100.2,434.79) -- (100.2,374.04) -- cycle ;
	\draw  [fill={rgb, 255:red, 244; green, 211; blue, 201 }  ,fill opacity=1 ] (346.95,374) -- (346.9,405.29) -- (346.9,434.8) -- (297.2,418.8) -- (297.2,390) -- cycle ;
	\draw  [color={rgb, 255:red, 0; green, 0; blue, 0 }  ,draw opacity=1 ][fill={rgb, 255:red, 0; green, 0; blue, 0 }  ,fill opacity=1 ] (383.02,401.82) -- (388.9,404.24) -- (383.02,406.66) -- (384.2,404.24) -- cycle ;
	\draw [line width=1.5]    (384.27,404.36) -- (346.87,404.33) ;

	\draw  [fill={rgb, 255:red, 244; green, 211; blue, 201 }  ,fill opacity=1 ] (346.95,89.4) -- (346.9,120.69) -- (346.9,150.2) -- (297.2,134.2) -- (297.2,105.4) -- cycle ;
	\draw  [color={rgb, 255:red, 0; green, 0; blue, 0 }  ,draw opacity=1 ][fill={rgb, 255:red, 0; green, 0; blue, 0 }  ,fill opacity=1 ] (383.02,117.22) -- (388.9,119.64) -- (383.02,122.06) -- (384.2,119.64) -- cycle ;
	\draw [line width=1.5]    (384.27,119.76) -- (346.87,119.73) ;
	
	\draw  [color={rgb, 255:red, 0; green, 0; blue, 0 }  ,draw opacity=1 ][fill={rgb, 255:red, 0; green, 0; blue, 0 }  ,fill opacity=1 ] (92.39,46.85) -- (98.26,49.27) -- (92.39,51.69) -- (93.56,49.27) -- cycle ;
	\draw [line width=1.5]    (93.63,49.39) -- (56.23,49.36) ;

	\draw  [color={rgb, 255:red, 0; green, 0; blue, 0 }  ,draw opacity=1 ][fill={rgb, 255:red, 0; green, 0; blue, 0 }  ,fill opacity=1 ] (92.39,329.85) -- (98.26,332.27) -- (92.39,334.69) -- (93.56,332.27) -- cycle ;
	\draw [line width=1.5]    (93.63,332.39) -- (56.23,332.36) ;

	\draw  [color={rgb, 255:red, 0; green, 0; blue, 0 }  ,draw opacity=1 ][fill={rgb, 255:red, 0; green, 0; blue, 0 }  ,fill opacity=1 ] (92.39,401.85) -- (98.26,404.27) -- (92.39,406.69) -- (93.56,404.27) -- cycle ;
	\draw [line width=1.5]    (93.63,404.39) -- (56.23,404.36) ;

	\draw  [fill={rgb, 255:red, 219; green, 234; blue, 243 }  ,fill opacity=1 ] (149.95,105.44) -- (149.9,120.69) -- (149.9,134.17) -- (100.2,150.19) -- (100.2,89.44) -- cycle ;
	\draw  [color={rgb, 255:red, 0; green, 0; blue, 0 }  ,draw opacity=1 ][fill={rgb, 255:red, 0; green, 0; blue, 0 }  ,fill opacity=1 ] (92.39,117.25) -- (98.26,119.67) -- (92.39,122.09) -- (93.56,119.67) -- cycle ;
	\draw [line width=1.5]    (93.63,119.79) -- (56.23,119.76) ;
	
	\draw  [fill={rgb, 255:red, 244; green, 211; blue, 201 }  ,fill opacity=1 ] (347.35,159.36) -- (347.3,190.65) -- (347.3,220.16) -- (297.6,204.16) -- (297.6,175.36) -- cycle ;
	\draw  [color={rgb, 255:red, 0; green, 0; blue, 0 }  ,draw opacity=1 ][fill={rgb, 255:red, 0; green, 0; blue, 0 }  ,fill opacity=1 ] (383.42,187.18) -- (389.3,189.6) -- (383.42,192.02) -- (384.6,189.6) -- cycle ;
	\draw [line width=1.5]    (384.67,189.72) -- (347.27,189.69) ;
	
	\draw  [fill={rgb, 255:red, 244; green, 211; blue, 201 }  ,fill opacity=1 ] (347.35,229.76) -- (347.3,261.05) -- (347.3,290.56) -- (297.6,274.56) -- (297.6,245.76) -- cycle ;
	\draw  [color={rgb, 255:red, 0; green, 0; blue, 0 }  ,draw opacity=1 ][fill={rgb, 255:red, 0; green, 0; blue, 0 }  ,fill opacity=1 ] (383.42,257.58) -- (389.3,260) -- (383.42,262.42) -- (384.6,260) -- cycle ;
	\draw [line width=1.5]    (384.67,260.12) -- (347.27,260.09) ;
	
	\draw  [color={rgb, 255:red, 248; green, 13; blue, 42 }  ,draw opacity=1 ][fill={rgb, 255:red, 248; green, 13; blue, 42 }  ,fill opacity=1 ] (289.39,47.01) -- (295.26,49.44) -- (289.39,51.86) -- (290.56,49.44) -- cycle ;

	\draw  [color={rgb, 255:red, 184; green, 4; blue, 242 }  ,draw opacity=1 ][fill={rgb, 255:red, 184; green, 4; blue, 242 }  ,fill opacity=1 ] (289.39,330.01) -- (295.26,332.44) -- (289.39,334.86) -- (290.56,332.44) -- cycle ;
	
	\draw  [color={rgb, 255:red, 73; green, 4; blue, 127 }  ,draw opacity=1 ][fill={rgb, 255:red, 73; green, 4; blue, 127 }  ,fill opacity=1 ] (289.39,402.01) -- (295.26,404.44) -- (289.39,406.86) -- (290.56,404.44) -- cycle ;
	
	\draw [color={rgb, 255:red, 248; green, 13; blue, 42 }  ,draw opacity=1 ][fill={rgb, 255:red, 248; green, 13; blue, 42 }  ,fill opacity=1 ][line width=1.5]    (290.63,49.56) -- (150.4,49.41) ;
	
	\draw [color={rgb, 255:red, 184; green, 4; blue, 242 }  ,draw opacity=1 ][fill={rgb, 255:red, 248; green, 13; blue, 42 }  ,fill opacity=1 ][line width=1.5]    (290.63,332.56) -- (150.4,332.41) ;
	
	\draw [color={rgb, 255:red, 73; green, 4; blue, 127 }  ,draw opacity=1 ][fill={rgb, 255:red, 248; green, 13; blue, 42 }  ,fill opacity=1 ][line width=1.5]    (290.63,404.56) -- (150.4,404.41) ;

	\draw  [color={rgb, 255:red, 0; green, 30; blue, 255 }  ,draw opacity=1 ][fill={rgb, 255:red, 0; green, 30; blue, 255 }  ,fill opacity=1 ] (289.47,117.31) -- (295.35,119.74) -- (289.47,122.16) -- (290.64,119.74) -- cycle ;
	\draw [color={rgb, 255:red, 0; green, 30; blue, 255 }  ,draw opacity=1 ][fill={rgb, 255:red, 0; green, 30; blue, 255 }  ,fill opacity=1 ][line width=1.5]    (290.71,119.86) -- (150.4,119.71) ;
	\draw   (190.12,172.57) -- (224.26,172.57) -- (224.26,206.7) -- (190.12,206.7) -- cycle ;
	\draw [color={rgb, 255:red, 0; green, 30; blue, 255 }  ,draw opacity=1 ][fill={rgb, 255:red, 0; green, 30; blue, 255 }  ,fill opacity=1 ][line width=1.5]    (169.4,255) -- (169.4,120.21) ;
	\draw  [color={rgb, 255:red, 0; green, 30; blue, 255 }  ,draw opacity=1 ][fill={rgb, 255:red, 0; green, 30; blue, 255 }  ,fill opacity=1 ] (181.72,183.12) -- (187.59,185.55) -- (181.72,187.97) -- (182.89,185.55) -- cycle ;
	\draw [color={rgb, 255:red, 0; green, 30; blue, 255 }  ,draw opacity=1 ][fill={rgb, 255:red, 0; green, 30; blue, 255 }  ,fill opacity=1 ][line width=1.5]    (183,185.67) -- (168.4,185.4) ;
	
	\draw [color={rgb, 255:red, 248; green, 13; blue, 42 }  ,draw opacity=1 ][fill={rgb, 255:red, 0; green, 30; blue, 255 }  ,fill opacity=1 ][line width=1.5]    (159.5,265) -- (159.5,50.42) ;
	\draw  [color={rgb, 255:red, 248; green, 13; blue, 42 }  ,draw opacity=1 ][fill={rgb, 255:red, 248; green, 13; blue, 42 }  ,fill opacity=1 ] (181.48,191.59) -- (187.36,194.01) -- (181.48,196.43) -- (182.66,194.01) -- cycle ;
	\draw [color={rgb, 255:red, 248; green, 13; blue, 42 }  ,draw opacity=1 ][fill={rgb, 255:red, 248; green, 13; blue, 42 }  ,fill opacity=1 ][line width=1.5]    (182.73,194.13) -- (158.9,193.86) ;
	
	\draw  [color={rgb, 255:red, 0; green, 30; blue, 255 }  ,draw opacity=1 ][fill={rgb, 255:red, 0; green, 30; blue, 255 }  ,fill opacity=1 ] (181.72,253.37) -- (187.59,255.8) -- (181.72,258.22) -- (182.89,255.8) -- cycle ;
	\draw [color={rgb, 255:red, 0; green, 30; blue, 255 }  ,draw opacity=1 ][fill={rgb, 255:red, 0; green, 30; blue, 255 }  ,fill opacity=1 ][line width=1.5]    (183.4,255.92) -- (168.4,255.65) ;
	
	\draw  [color={rgb, 255:red, 248; green, 13; blue, 42 }  ,draw opacity=1 ][fill={rgb, 255:red, 248; green, 13; blue, 42 }  ,fill opacity=1 ] (181.48,261.84) -- (187.36,264.26) -- (181.48,266.68) -- (182.66,264.26) -- cycle ;
	\draw [color={rgb, 255:red, 248; green, 13; blue, 42 }  ,draw opacity=1 ][fill={rgb, 255:red, 248; green, 13; blue, 42 }  ,fill opacity=1 ][line width=1.5]    (182.73,264.38) -- (158.9,264.11) ;
	
	\draw  [color={rgb, 255:red, 184; green, 4; blue, 242 }  ,draw opacity=1 ][fill={rgb, 255:red, 184; green, 4; blue, 242 }  ,fill opacity=1 ] (289.97,187.62) -- (295.84,190.05) -- (289.97,192.47) -- (291.14,190.05) -- cycle ;
	\draw [color={rgb, 255:red, 184; green, 4; blue, 242 }  ,draw opacity=1 ][fill={rgb, 255:red, 184; green, 4; blue, 242 }  ,fill opacity=1 ][line width=1.5]    (291.21,190.17) -- (224.7,189.9) ;
	\draw  [color={rgb, 255:red, 73; green, 4; blue, 127 }  ,draw opacity=1 ][fill={rgb, 255:red, 73; green, 4; blue, 127 }  ,fill opacity=1 ] (289.97,257.87) -- (295.84,260.3) -- (289.97,262.72) -- (291.14,260.3) -- cycle ;
	\draw [color={rgb, 255:red, 73; green, 4; blue, 127 }  ,draw opacity=1 ][fill={rgb, 255:red, 73; green, 4; blue, 127 }  ,fill opacity=1 ][line width=1.5]    (291.21,260.42) -- (224.7,260.15) ;
	\draw   (190.12,243) -- (224.26,243) -- (224.26,277.13) -- (190.12,277.13) -- cycle ;
	
	\draw (42.38,44.54) node [anchor=north west][inner sep=0.75pt]  [font=\small] [align=left] {{\tiny $a$}};
	\draw (18.38,327.54) node [anchor=north west][inner sep=0.75pt]  [font=\small] [align=left] {{\tiny $a + b$}};
	\draw (18.38,399.54) node [anchor=north west][inner sep=0.75pt]  [font=\small] [align=left] {{\tiny $a \times b$}};
	\draw (42,114.94) node [anchor=north west][inner sep=0.75pt]  [font=\small] [align=left] {{\tiny $b$}};
	\draw (390.38,45.54) node [anchor=north west][inner sep=0.75pt]  [font=\small] [align=left] {{\tiny $a$}};
	\draw (390.38,327) node [anchor=north west][inner sep=0.75pt]  [font=\small] [align=left] {{\tiny $a + b$}};
	\draw (390.38,399) node [anchor=north west][inner sep=0.75pt]  [font=\small] [align=left] {{\tiny $a \times b$}};
	\draw (390.38,115.94) node [anchor=north west][inner sep=0.75pt]  [font=\small] [align=left] {{\tiny $b$}};
	\draw (390.38,184.9) node [anchor=north west][inner sep=0.75pt]  [font=\small] [align=left] {{\tiny $a + b$}};
	\draw (390.38,255.3) node [anchor=north west][inner sep=0.75pt]  [font=\small] [align=left] {{\tiny $a \times b$}};
	\draw (207.61,34.6) node [anchor=north west][inner sep=0.75pt]  [font=\small] [align=left] {{\tiny $\displaystyle h_1 = \varphi(a)$}};
	\draw (190.61,317.6) node [anchor=north west][inner sep=0.75pt]  [font=\small] [align=left] {{\tiny $\displaystyle h_3 = \varphi(a+b)$}};
	\draw (190.61,389.6) node [anchor=north west][inner sep=0.75pt]  [font=\small] [align=left] {{\tiny $\displaystyle h_4=\varphi(a\times b)$}};
	\draw (207.61,105) node [anchor=north west][inner sep=0.75pt]  [font=\small] [align=left] {{\tiny $\displaystyle h_2 = \varphi(b)$}};
	\draw (234.05,174.98) node [anchor=north west][inner sep=0.75pt]  [font=\small] [align=left] {{\tiny $\displaystyle h_1 +_{\Theta_1} h_2$}};
	\draw (234.05,245.21) node [anchor=north west][inner sep=0.75pt]  [font=\small] [align=left] {{\tiny $\displaystyle h_1 \times_{\Theta_2} h_2$}};
	\draw (196,180.14) node [anchor=north west][inner sep=0.75pt]  [font=\small] [align=left] {{\fontfamily{ptm}\selectfont {\tiny ANO}}};
	\draw (195.75,190.14) node [anchor=north west][inner sep=0.75pt]  [font=\small] [align=center] {{\fontfamily{ptm}\selectfont {\tiny $\displaystyle +_{\Theta_1}$}}};
	\draw (196.5,250.5) node [anchor=north west][inner sep=0.75pt]  [font=\small] [align=left] {{\fontfamily{ptm}\selectfont {\tiny ANO}}};
	\draw (195.9,260.14) node [anchor=north west][inner sep=0.75pt]  [font=\small] [align=center] {{\fontfamily{ptm}\selectfont {\tiny $\displaystyle \times_{\Theta_2}$}}};
	\draw (107,115.6) node [anchor=north west][inner sep=0.75pt]   [align=left] {{\scriptsize {\fontfamily{ptm}\selectfont Encoder}}};
	\draw (107,44.9) node [anchor=north west][inner sep=0.75pt]   [align=left] {{\scriptsize {\fontfamily{ptm}\selectfont Encoder}}};
	\draw (107,327.9) node [anchor=north west][inner sep=0.75pt]   [align=left] {{\scriptsize {\fontfamily{ptm}\selectfont Encoder}}};
	\draw (107,399.9) node [anchor=north west][inner sep=0.75pt]   [align=left] {{\scriptsize {\fontfamily{ptm}\selectfont Encoder}}};
	\draw (304.8,114.3) node [anchor=north west][inner sep=0.75pt]   [align=left] {{\scriptsize {\fontfamily{ptm}\selectfont Decoder}}};
	\draw (304.8,184.76) node [anchor=north west][inner sep=0.75pt]   [align=left] {{\scriptsize {\fontfamily{ptm}\selectfont Decoder}}};
	\draw (304.8,256.16) node [anchor=north west][inner sep=0.75pt]   [align=left] {{\scriptsize {\fontfamily{ptm}\selectfont Decoder}}};
	\draw (304.8,44.4) node [anchor=north west][inner sep=0.75pt]   [align=left] {{\scriptsize {\fontfamily{ptm}\selectfont Decoder}}};
	\draw (304.8,327.4) node [anchor=north west][inner sep=0.75pt]   [align=left] {{\scriptsize {\fontfamily{ptm}\selectfont Decoder}}};
	\draw (304.8,399.4) node [anchor=north west][inner sep=0.75pt]   [align=left] {{\scriptsize {\fontfamily{ptm}\selectfont Decoder}}};

\draw  [dash pattern={on 1.69pt off 2.76pt}][line width=1.5]  (600.53,420.11) -- (456.62,420.26) -- (456.39,200.24) -- (600.3,200.09) -- cycle ;
\draw   (593.05,340.69) .. controls (593.06,345.27) and (589.35,348.99) .. (584.76,349) -- (547.47,349.04) .. controls (542.89,349.04) and (539.17,345.33) .. (539.16,340.74) -- (539.14,315.84) .. controls (539.13,311.26) and (542.85,307.54) .. (547.43,307.54) -- (584.72,307.5) .. controls (589.3,307.49) and (593.02,311.2) .. (593.03,315.79) -- cycle ;
\draw  [color={rgb, 255:red, 184; green, 4; blue, 242 }  ,draw opacity=1 ][fill={rgb, 255:red, 184; green, 4; blue, 242 }  ,fill opacity=1 ] (551.83,356.9) -- (554.25,351.02) -- (556.68,356.9) -- (554.25,355.72) -- cycle ;
\draw [color={rgb, 255:red, 184; green, 4; blue, 242 }  ,draw opacity=1 ][fill={rgb, 255:red, 184; green, 4; blue, 242 }  ,fill opacity=1 ][line width=1.5]    (554.26,356.21) -- (554.3,410.3) ;
\draw  [color={rgb, 255:red, 184; green, 4; blue, 242 }  ,draw opacity=1 ][fill={rgb, 255:red, 184; green, 4; blue, 242 }  ,fill opacity=1 ] (575.08,357.13) -- (577.5,351.25) -- (579.93,357.12) -- (577.5,355.95) -- cycle ;
\draw [color={rgb, 255:red, 184; green, 4; blue, 242 }  ,draw opacity=1 ][fill={rgb, 255:red, 184; green, 4; blue, 242 }  ,fill opacity=1 ][line width=1.5]    (577.51,355.98) -- (577.55,410.3) ;
\draw   (519.05,340.77) .. controls (519.06,345.35) and (515.35,349.07) .. (510.76,349.07) -- (473.47,349.11) .. controls (468.89,349.12) and (465.17,345.41) .. (465.16,340.82) -- (465.14,315.92) .. controls (465.13,311.34) and (468.85,307.62) .. (473.43,307.61) -- (510.72,307.57) .. controls (515.3,307.57) and (519.02,311.28) .. (519.03,315.87) -- cycle ;
\draw  [color={rgb, 255:red, 73; green, 4; blue, 127 }  ,draw opacity=1 ][fill={rgb, 255:red, 73; green, 4; blue, 127 }  ,fill opacity=1 ] (477.83,356.98) -- (480.25,351.1) -- (482.68,356.97) -- (480.25,355.8) -- cycle ;
\draw [color={rgb, 255:red, 73; green, 4; blue, 127 }  ,draw opacity=1 ][fill={rgb, 255:red, 73; green, 4; blue, 127 }  ,fill opacity=1 ][line width=1.5]    (480.26,356.29) -- (480.3,410.4) ;
\draw  [color={rgb, 255:red, 73; green, 4; blue, 127 }  ,draw opacity=1 ][fill={rgb, 255:red, 73; green, 4; blue, 127 }  ,fill opacity=1 ] (501.08,357.2) -- (503.5,351.32) -- (505.93,357.2) -- (503.5,356.03) -- cycle ;
\draw [color={rgb, 255:red, 73; green, 4; blue, 127 }  ,draw opacity=1 ][fill={rgb, 255:red, 73; green, 4; blue, 127 }  ,fill opacity=1 ][line width=1.5]    (503.51,356.06) -- (503.55,410.4) ;
\draw   (537.89,266.64) .. controls (537.9,271.77) and (533.74,275.94) .. (528.61,275.94) .. controls (523.48,275.95) and (519.31,271.79) .. (519.31,266.66) .. controls (519.3,261.53) and (523.46,257.37) .. (528.59,257.36) .. controls (533.72,257.36) and (537.89,261.51) .. (537.89,266.64) -- cycle ;
\draw [line width=1.5]    (566.98,307.64) .. controls (568.62,281.64) and (564.94,270.31) .. (537.89,266.64) ;
\draw [line width=1.5]    (490.98,307.23) .. controls (488.62,282.95) and (492.28,269.61) .. (519.31,266.66) ;
\draw  [color={rgb, 255:red, 0; green, 0; blue, 0 }  ,draw opacity=1 ][fill={rgb, 255:red, 0; green, 0; blue, 0 }  ,fill opacity=1 ] (525.6,214.64) -- (528.02,208.76) -- (530.45,214.64) -- (528.02,213.46) -- cycle ;
\draw [color={rgb, 255:red, 0; green, 0; blue, 0 }  ,draw opacity=1 ][fill={rgb, 255:red, 184; green, 4; blue, 242 }  ,fill opacity=1 ][line width=1.5]    (528.03,213.5) -- (528.07,257.8) ;

\draw (543.15,390.9) node [anchor=north west][inner sep=0.75pt]  [rotate=-269.94] [align=left] {{\tiny $h_3$}};
\draw (563.81,410.54) node [anchor=north west][inner sep=0.75pt]  [rotate=-269.94] [align=left] {{\tiny $h_1 +_{\Theta_1} h_2$}};
\draw (555.58,322.79) node [anchor=north west][inner sep=0.75pt]  [font=\normalsize] [align=left] {$\Delta_E$};
\draw (469.15,390.9) node [anchor=north west][inner sep=0.75pt]  [rotate=-269.94] [align=left] {{\tiny $h_4$}};
\draw (490,410.54) node [anchor=north west][inner sep=0.75pt]  [rotate=-269.94] [align=left] {{\tiny $\displaystyle h_1 \times_{\Theta_2} h_2$}};
\draw (481.58,322.87) node [anchor=north west][inner sep=0.75pt]  [font=\normalsize] [align=left] {$\Delta_E$};
\draw (511.34,248.98) node [anchor=north west][inner sep=0.75pt]  [rotate=-269.94] [align=left] {{\normalfont $\ell_{\mathrm{iso}}$}};
\draw (535.32,272.33) node [anchor=north west][inner sep=0.75pt]  [rotate=-179.94] [align=left] {{\fontfamily{ptm}\selectfont $+$}};

\end{tikzpicture}
}
	\caption{Overall training procedure of the Neural Isomorphic Field. The figure shows reconstruction, the decoder loss on the operator outputs and $\ell_{\mathrm{iso}}$; the order loss, the compositions and the hint terms of $\dot{\times}$ are given in Section~\ref{sec:papdt} and Table~\ref{tab:hparams}.}
	\label{fig:overall_training}
\end{figure}

\begin{proposition}[Transport of an ordered algebraic structure]\label{prop:transport}
	Let $(R,+_1,\times_1,<_1)$ be an ordered integral domain (respectively, an ordered field), let $X$ be a set, let $\varphi:R\rightarrow X$ be a map, let $+_2,\times_2:X\times X\rightarrow X$ be binary operations, and let $<_2\,\subseteq X\times X$ be a relation such that, for all $a,b\in R$,
	\begin{align}
		\varphi(a+_1 b) &= \varphi(a)+_2 \varphi(b), \label{eq:addition_iso}\\
		\varphi(a\times_1 b) &= \varphi(a)\times_2 \varphi(b), \label{eq:multiplication_iso}\\
		a <_1 b &\Longleftrightarrow \varphi(a) <_2 \varphi(b). \label{eq:order_field}
	\end{align}
	Then $\varphi$ is injective, $\varphi(R)$ is closed under $+_2$ and $\times_2$, the structure $(\varphi(R),+_2,\times_2,<_2)$ (with operations and relation restricted to $\varphi(R)$) is an ordered integral domain (respectively, an ordered field) with zero $\varphi(0)$ and identity $\varphi(1)$, and $\varphi:R\rightarrow\varphi(R)$ is an isomorphism of ordered rings (respectively, of ordered fields).
\end{proposition}

Conditions (\ref{eq:addition_iso}), (\ref{eq:multiplication_iso}), and (\ref{eq:order_field}) in Proposition~\ref{prop:transport} ensure the establishment of an ordered integral domain. To satisfy these conditions in the NIF model, two loss functions can be defined: loss function $\ell_{\mathrm{iso}}$ addresses conditions (\ref{eq:addition_iso}) and (\ref{eq:multiplication_iso}), while loss function $\ell_{\mathrm{ord}}$ addresses condition (\ref{eq:order_field}). To determine the ideal conditions for $H$ to be isomorphic to $\Dset$, Lemmas~\ref{lem:rec}, \ref{lem:iso}, and \ref{lem:ord} and Theorem~\ref{theorem:total_conds} explore the necessary criteria. These results also provide the definitions of $\ell_{\mathrm{iso}}$ and $\ell_{\mathrm{ord}}$. Additionally, achieving isomorphism requires the loss function $\ell_{\mathrm{rec}}$, which represents the reconstruction loss in autoencoder neural networks, and its definition is also provided. Figure \ref{fig:overall_training} shows the overall training procedure for creating a Neural Isomorphic Field.

In this section, $\distP$ is a probability distribution on $\Dset$ with $\distP(a)>0$ for all $a\in\Dset$; $a\sim\distP$ denotes a draw from $\distP$ and $a,b\sim\distP$ two independent draws, so each pair $(a,b)$ has probability $\distP(a)\distP(b)>0$. The functions $\Delta_{\Dset}$ and $\Delta_E$ are discrepancies on $\Dset\cup\{\bot\}$ and on $\sR^d$, that is, non-negative functions that are zero exactly on the diagonal (any metric is one, and so is the squared Euclidean distance). Such a distribution exists because $\Dset$ is countable; the sampler of Section~\ref{subsec:distribution_and_numerical_dataset} is an example.

\begin{lemma}[{Reconstruction loss and injectivity}]\label{lem:rec}
If $\ell_{\mathrm{rec}} = \mathbb{E}_{a\sim\mathcal{P}}[\Delta_{\Dset}(\varphi^\dagger(\varphi(a)), a)] = 0$, then $\varphi^\dagger(\varphi(a))=a$ for every $a\in\Dset$; in particular, $\varphi$ is injective.
\end{lemma}

\begin{lemma}[Isomorphism loss and homomorphism]\label{lem:iso}
	Let $*\in\{+,\times\}$ and let $\dot{*}$ be the corresponding learned operator. If $\mathbb{E}_{a,b\sim\mathcal{P}}[\Delta_E(\varphi(a * b),\ \varphi(a)\,\dot{*}\,\varphi(b))]=0$, then $\varphi(a*b)=\varphi(a)\,\dot{*}\,\varphi(b)$ for all $a,b\in\Dset$.
\end{lemma}

\begin{lemma}[{Order loss and neural order}]\label{lem:ord}
If
	\begin{align}
		\ell_{\mathrm{ord}} = \mathbb{E}_{a,b\sim\mathcal{P}}[ (1(a<b) - 1(\varphi(a)<_{\omega} \varphi(b)))^2] = 0,
	\end{align}
then $a<b \Leftrightarrow \varphi(a)<_\omega\varphi(b)$ for all $a,b\in\Dset$, $\varphi$ is injective, and $<_\omega$ is a neural order on $\varphi(\Dset)$.
\end{lemma}

\begin{theorem}[{Ideal conditions}]\label{theorem:total_conds}
	 If
	\begin{align}\label{total_loss_eq}
		\ell_{\mathrm{total}} = \ell_{\mathrm{rec}} + \ell_{\mathrm{iso}} + \ell_{\mathrm{ord}} = 0,
	\end{align}

	where

		\begin{align}
			\ell_{\mathrm{rec}} &= \mathbb{E}_{a\sim\mathcal{P}}[\Delta_{\Dset}(\varphi^\dagger(\varphi(a)), a)] \\
			\ell_{\mathrm{ord}} &= \mathbb{E}_{a,b\sim\mathcal{P}}[ (1(a<b) - 1(\varphi(a)<_{\omega} \varphi(b)))^2] \\
			\ell_{\mathrm{iso}} &= \mathbb{E}_{a,b\sim\mathcal{P}}[\Delta_E(\varphi(a + b), \ \varphi(a) \ \dot{+} \  \varphi(b))] \\
			&\quad + \mathbb{E}_{a,b\sim\mathcal{P}}[\Delta_E(\varphi(a \times b), \ \varphi(a) \ \dot{\times} \ \varphi(b))]
		\end{align}
	then $H=\varphi(\Dset)$ is closed under $\dot{+}$ and $\dot{\times}$, the tuple $(H,\dot{+},\dot{\times},<_\omega)$ is a neural ordered integral domain (Definition~\ref{def:neural_ordered_domain}), and $\varphi:\Dset\rightarrow H$ is an isomorphism of ordered rings onto $(H,\dot{+},\dot{\times},<_\omega)$ whose inverse is the restriction of $\varphi^\dagger$ to $H$. The same conclusion holds for every weighted objective $\lambda_{\mathrm{rec}}\ell_{\mathrm{rec}}+\lambda_{\mathrm{iso}}\ell_{\mathrm{iso}}+\lambda_{\mathrm{ord}}\ell_{\mathrm{ord}}=0$ with $\lambda_{\mathrm{rec}},\lambda_{\mathrm{iso}},\lambda_{\mathrm{ord}}>0$.

\end{theorem}

The rational numbers are obtained from $H$ through the field-of-fractions construction, which needs only the two learned operators and the comparator. For $h\in H$, let $-h:=\varphi(-\varphi^{-1}(h))$ be its additive inverse in $H$, and let $H^{*}:=H\setminus\{\varphi(0)\}$. Under the conditions of Theorem~\ref{theorem:total_conds}, $-h=\varphi(-1)\dot{\times}h$, so $-h$ is also computed in the embedding space.

\begin{corollary}[Neural isomorphic field of fractions]\label{cor:fraction_field}
	Assume the conditions of Theorem~\ref{theorem:total_conds}. On $H\times H^{*}$ define $(h_1,h_1')\sim(h_2,h_2')$ if and only if $h_1\dot{\times}h_2'=h_2\dot{\times}h_1'$, write $[h,h']$ for the equivalence class of $(h,h')$, and let $F_H:=(H\times H^{*})/\!\sim$. For classes represented with positive denominators, i.e., $\varphi(0)<_\omega h_1'$ and $\varphi(0)<_\omega h_2'$, define
	\begin{align}
		[h_1,h_1']\oplus[h_2,h_2'] &:= [\,h_1\dot{\times}h_2' \,\dot{+}\, h_2\dot{\times}h_1',\ h_1'\dot{\times}h_2'\,], \label{eq:frac_add}\\
		[h_1,h_1']\otimes[h_2,h_2'] &:= [\,h_1\dot{\times}h_2,\ h_1'\dot{\times}h_2'\,], \label{eq:frac_mul}\\
		[h_1,h_1']\prec[h_2,h_2'] &:\Longleftrightarrow h_1\dot{\times}h_2' <_\omega h_2\dot{\times}h_1'. \label{eq:frac_ord}
	\end{align}
	Then:
	\begin{enumerate}
		\item[(a)] $\sim$ is an equivalence relation, every class has a representative with a positive denominator, and $\oplus$, $\otimes$ and $\prec$ are well defined;
		\item[(b)] $(F_H,\oplus,\otimes,\prec)$ is an ordered field;
		\item[(c)] the map $\Phi:\sQ\rightarrow F_H$, $\Phi(a/b):=[\varphi(a),\varphi(b)]$ for $a,b\in\Dset$ with $b>0$, is well defined and is an isomorphism of ordered fields;
		\item[(d)] the map that sends $r\in\sQ$, written in lowest terms as $r=p/q$ with $p\in\sZ$, $q\in\sZ_{>0}$ and $\gcd(p,q)=1$, to the vector $(\varphi(p),\varphi(q))\in\sR^{2d}$ is injective, and its value is a representative of $\Phi(r)$.
	\end{enumerate}
\end{corollary}

Corollary~\ref{cor:fraction_field} explains the name \emph{Neural Isomorphic Field}: under the ideal conditions, the learned operators and the comparator turn the embedding space into an ordered field isomorphic to $\sQ$, and the trained model approximates this structure (Section~\ref{sec:approximate}). For example, $1/3$ is represented by $[\varphi(1),\varphi(3)]$ and its inverse by $[\varphi(3),\varphi(1)]$, although $1/3\notin\Dset$; division by $[h_2,h_2']$, that is, $[h_1,h_1']\otimes[h_2',h_2]$, thus needs only $\dot{\times}$ (if $h_2<_\omega\varphi(0)$, the representative $[\varphi(-1)\dot{\times}h_2',\varphi(-1)\dot{\times}h_2]$ of the inverse keeps the denominator positive). Equality, addition, multiplication and comparison of rational numbers need a bounded number of applications of $\dot{+}$, $\dot{\times}$ and $\omega$ to embeddings of elements of $\Dset$, so the algebraic tests on $\Dset$ (Section~\ref{sec:expriments}) also measure the induced structure on $\sQ$, and the fraction-field tests (Section~\ref{sec:evaluation_criteria}) evaluate it explicitly. One vector of $\sR^d$ represents an element of $\Dset$, whereas a rational number is represented by a pair, a fixed-length vector of $\sR^{2d}$.

\begin{remark}[Role of the operator class]\label{rem:transport}
	The conditions of Theorem~\ref{theorem:total_conds} constrain $\varphi$, $\dot{+}$ and $\dot{\times}$ together. If the operators could be any functions, then for every injective $\varphi$ the definitions $h_1\dot{+}h_2:=\varphi(\varphi^{-1}(h_1)+\varphi^{-1}(h_2))$ and $h_1\dot{\times}h_2:=\varphi(\varphi^{-1}(h_1)\varphi^{-1}(h_2))$ on $H\times H$ would satisfy Eqs.~(\ref{eq:addition_iso})--(\ref{eq:multiplication_iso}) exactly (transport of structure), whatever the geometry of $H$. Theorem~\ref{theorem:total_conds} is thus a sufficient condition whose practical meaning depends on the class of operators. In NIF, the operators are commutative by construction (Lemma~\ref{lem:commutativity}), consist of a few ($k$) shared AC layers, and are tested on pairs that do not occur in the training data. Their accuracy on such held-out pairs (Section~\ref{sec:expriments}) is the evidence that the structure lies in the embedding geometry and generalizes, instead of being memorized by the operators. For the same reason, deeper operators are not always better (Section~\ref{sec:approximate} and the ablation study).
\end{remark}

\begin{remark}[Ideal conditions versus the trained model]\label{rem:ideal_vs_practice}
	Theorem~\ref{theorem:total_conds} and Corollary~\ref{cor:fraction_field} describe an ideal case that a trained network does not reach, for three reasons. The losses of a trained model are positive; the training distribution lives on the finite set $\Dn$ (Section~\ref{subsec:distribution_and_numerical_dataset}), so $\distP(a)=0$ for $a\notin\Dn$, and $\Dn$ is closed under neither operation; and the losses are optimized through differentiable surrogates (Lemma~\ref{lem:surrogates}). The trained NIF is therefore not an exact isomorphism but an approximation of the ideal structure. For the field of fractions, the operations of Eqs.~(\ref{eq:frac_add})--(\ref{eq:frac_ord}) are evaluated on their cross-products, and the fraction-field tests decode the resulting vectors and compare them in $\sQ$ (Section~\ref{sec:evaluation_criteria}). Section~\ref{sec:approximate} shows how the remaining errors affect single operations, chains of operations, and the associative and distributive laws.
\end{remark}

In the implementation, $\ell_{\mathrm{rec}}$ is the token-level cross-entropy of the decoder under teacher forcing plus the cross-entropy of the two length classifiers, and $\ell_{\mathrm{ord}}$ is replaced by the cross-entropy
\[
\ell^{\mathrm{CE}}_{\mathrm{ord}}=\mathbb{E}_{a,b\sim\distP}\bigl[-\ln\bar\omega_{\rel(a,b)}(\varphi(a),\varphi(b))\bigr],
\]
where $\rel(a,b)\in\{1,2,3\}$ indicates whether $a<b$, $a>b$ or $a=b$. In $\ell_{\mathrm{iso}}$, $\Delta_E$ is the mean squared error and the target $\varphi(a*b)$ is held constant (stop-gradient), so $\ell_{\mathrm{iso}}$ pulls the operator outputs toward the embeddings rather than the embeddings toward each other (their scale is fixed by the per-slot normalization of Section~\ref{subsec:architecture}). Three more groups of non-negative terms enter the objective: the decoder cross-entropy on the operator outputs $\varphi(a)\,\dot{*}\,\varphi(b)$ with target $a*b$; closure terms, namely $\ell_{\mathrm{iso}}$ and the decoder cross-entropy for compositions (namely $(a*b)*c$, $((a+b)+c)+d$, and both sides of $a(b+c)$) and $\ell^{\mathrm{CE}}_{\mathrm{ord}}$ for pairs of operator outputs $(\varphi(a)\dot{+}\varphi(b),\varphi(a)\dot{+}\varphi(c))$; and, for $\dot{\times}$, the hint terms of Remark~\ref{rem:biases}. Theorem~\ref{theorem:total_conds} applies whenever $\ell_{\mathrm{rec}}$, $\ell_{\mathrm{iso}}$ and $\ell_{\mathrm{ord}}$ are zero; the other terms serve to reach this state in training. Apart from the sampler pairs of Section~\ref{subsec:distribution_and_numerical_dataset}, training uses cancellations $a+(-a)$ and $a+(-a+\delta)$ and products $x\cdot y=10^{k}$ with $x=\pm2^i5^j$ and $k\neq 0$, but never the pairs $(x,x^{-1})$ of the multiplicative invertibility test. Appendix~\ref{app:implementation} gives the weights and all other details. The next lemma shows that these surrogates control the losses of Theorem~\ref{theorem:total_conds}.

\begin{lemma}[Cross-entropy surrogates]\label{lem:surrogates}
	For $a\in\Dset$, let $y^*(a)=(y^*_1,\dots,y^*_n)$, $n=n(a)$, be the token sequence of $\str{a}$ and $l_{\mathrm{int}}(a)$, $l_{\mathrm{frac}}(a)$ its numbers of integer and fractional digits. Let $\pi^{\mathrm{int}}(\cdot\mid z)$ and $\pi^{\mathrm{frac}}(\cdot\mid z)$ be the distributions of the two length classifiers, let $\pi_i(\cdot\mid y^*_{<i},z)$ be the decoder's distribution for the $i$-th token given the correct lengths, the correct prefix and the decoder input $z$, and let $\ell^{\mathrm{CE}}_{\mathrm{rec}}=\mathbb{E}_{a\sim\distP}\big[-\ln\pi^{\mathrm{int}}(l_{\mathrm{int}}(a)\mid\varphi(a))-\ln\pi^{\mathrm{frac}}(l_{\mathrm{frac}}(a)\mid\varphi(a))+\sum_{i=1}^n -\ln \pi_i(y^*_i\mid y^*_{<i},\varphi(a))\big]$. Then
	\begin{equation}\label{eq:surrogates}
		\Pr_{a\sim\distP}\big[\varphi^\dagger(\varphi(a))\neq a\big] \le \frac{\ell^{\mathrm{CE}}_{\mathrm{rec}}}{\ln 2},
		\qquad
		\ell_{\mathrm{ord}} \le \frac{\ell^{\mathrm{CE}}_{\mathrm{ord}}}{\ln 2},
	\end{equation}
	where $\ell_{\mathrm{ord}}$ is computed with the relation $<_{\bar\omega}$. In particular, the cross-entropy losses vanish only if $\ell_{\mathrm{rec}}=0$ (with the discrete metric $\Delta_{\Dset}(a,b)=1(a\neq b)$) and $\ell_{\mathrm{ord}}=0$.
\end{lemma}

\subsection{Approximate algebraic structure under non-zero losses}\label{sec:approximate}

This section links quantities that are small but not zero after training to failures of the algebraic laws on the finite domain $\Dn$. Throughout, $\norm{\cdot}$ is the Euclidean norm on $\sR^d$ and $\norm{\cdot}_\infty$ the maximum norm.

\begin{definition}[Algebraic defects]\label{def:defects}
	The additive and multiplicative defects of a trained NIF are
	\begin{align}
		\epsadd(a,b) &:= \norm{\varphi(a)\dot{+}\varphi(b)-\varphi(a+b)} \quad \text{for } (a,b)\in\Aadd,\\
		\epsmul(a,b) &:= \norm{\varphi(a)\dot{\times}\varphi(b)-\varphi(ab)} \quad \text{for } (a,b)\in\Amul .
	\end{align}
\end{definition}

\begin{definition}[Decoding radius]\label{def:radius}
	For $a\in\Dn$, let $\mathcal{R}(a)$ be the set of $\delta>0$ such that $\varphi^\dagger(z)=a$ for every $z\in\sR^d$ with $\norm{z-\varphi(a)}<\delta$. The \emph{decoding radius} of $a$ is $\rho(a):=\sup\mathcal{R}(a)$ if $\mathcal{R}(a)\neq\emptyset$, and $\rho(a):=0$ otherwise.
\end{definition}

By definition, $\norm{z-\varphi(a)}<\rho(a)$ implies $\varphi^\dagger(z)=a$. The radius is positive only if $a$ is reconstructed correctly, and it is large when the decoder is confident and smooth around $\varphi(a)$:

\begin{lemma}[Margin bound on the decoding radius]\label{lem:margin}
	Let $a\in\Dn$ with token sequence $y^*=(y^*_1,\dots,y^*_n)$, let $\psi_i(z)\in\sR^{V}$ be the decoder logits for the $i$-th token given the decoder input $z$, the correct lengths and the correct prefix $y^*_{<i}$, where $V$ is the number of tokens, and let $\psi^{\mathrm{int}}(z)$ and $\psi^{\mathrm{frac}}(z)$ be the logits of the two length classifiers. With the margin $m(v,y):=v_y-\max_{y'\neq y}v_{y'}$ of a logit vector $v$ at the class $y$, define the \emph{decoding margin}
	\begin{equation}
		\gamma(a):=\min\Big\{m\big(\psi^{\mathrm{int}}(\varphi(a)),l_{\mathrm{int}}(a)\big),\ m\big(\psi^{\mathrm{frac}}(\varphi(a)),l_{\mathrm{frac}}(a)\big),\ \min_{1\le i\le n}m\big(\psi_i(\varphi(a)),y^*_i\big)\Big\}.
	\end{equation}
	Let $\delta_0>0$ and suppose that $\psi^{\mathrm{int}}$, $\psi^{\mathrm{frac}}$ and each $\psi_i$ are $L_\psi$-Lipschitz, with the norm $\norm{\cdot}$ on $\sR^d$ and the maximum norm on the logits, on the closed ball of radius $\delta_0$ around $\varphi(a)$. If $\gamma(a)>0$, then $\rho(a)\ge\min\{\delta_0,\ \gamma(a)/(2L_\psi)\}$.
\end{lemma}

\begin{proposition}[Decoding errors of single operations]\label{prop:decoding_error}
	Let $(a,b)\in\Aadd$ and $z:=\varphi(a)\dot{+}\varphi(b)$.
	\begin{enumerate}
		\item[(a)] If $\epsadd(a,b)<\rho(a+b)$, then $\varphi^\dagger(z)=a+b$.
		\item[(b)] If $\norm{z-\varphi(c)}<\rho(c)$ for some $c\in\Dn$ with $c\neq a+b$, then $\varphi^\dagger(z)=c\neq a+b$.
		\item[(c)] Let $(a,b)$ be drawn from any distribution on $\Aadd$. For every $\tau>0$,
		\begin{equation}\label{eq:decoding_error}
			\Pr\big[\norm{z-\varphi(c)}<\rho(c)\text{ for some }c\neq a+b\big] \le{}\Pr\big[\varphi^\dagger(z)\neq a+b\big] \le \Pr\big[\rho(a+b)\le\tau\big] + \frac{\mathbb{E}\big[\epsadd(a,b)^2\big]}{\tau^2}.
		\end{equation}
	\end{enumerate}
	The same statements hold for $\Amul$, $\dot{\times}$, $\epsmul$ and $ab$.
\end{proposition}

Item (a) is a sufficient condition for a correct result and item (b) one for an error. An operation fails, for example, when its result falls into the decoding ball of a nearby number, such as a number that differs from $a$ in one digit in the multiplicative identity test. In the experiments, the share of test pairs that satisfy (a) is thus a lower bound on the exact match, and one minus the share that satisfies (b) an upper bound (Section~\ref{subsec:error_propagation}).

The first term of the upper bound in Eq.~(\ref{eq:decoding_error}) depends on how well the \emph{results} $a+b$ (or $ab$) are reconstructed, not the operands. The second term is controlled by $\ell_{\mathrm{iso}}$: if $\Delta_E(x,y)=\norm{x-y}^2$ and the pairs follow the training distribution of pairs conditioned on $\Aadd$, then $\mathbb{E}[\epsadd(a,b)^2]$ equals the additive term of $\ell_{\mathrm{iso}}$ on admissible pairs (for the mean squared error, $\Delta_E(x,y)=\norm{x-y}^2/d$, the term is multiplied by $d$). Products of operands in $\Dn$ are much longer than the operands, so for multiplication the first term is large if the autoencoder is trained on operands only. For this reason, the objective also applies the decoder loss to the results of the operations (Section~\ref{sec:papdt}).

Compositions depend on the sensitivity of the learned operators. An operator $\dot{*}$ is \emph{$L$-Lipschitz in its first argument} if $\norm{u\,\dot{*}\,v-u'\,\dot{*}\,v}\le L\norm{u-u'}$ for all $u,u',v\in\sR^d$; by Lemma~\ref{lem:commutativity}, it is then also $L$-Lipschitz in its second argument.

\begin{lemma}[Lipschitz constants of stacked AC operators]\label{lem:ac_lipschitz}
	Suppose that each AC layer satisfies
	\begin{equation}\label{eq:layer_lipschitz}
		\norm{f([x_1,x_2];\theta_i)-f([x'_1,x'_2];\theta_i)}\le\lambda^{(1)}_i\norm{x_1-x'_1}+\lambda^{(2)}_i\norm{x_2-x'_2}
	\end{equation}
	for all $x_1,x_2,x'_1,x'_2\in\sR^d$, and let $c_i:=\lambda^{(1)}_i+\lambda^{(2)}_i$. Then $*_\Theta$ in Eq.~(\ref{eq:neural_op}) is $L$-Lipschitz in each argument with
	\begin{equation}\label{eq:ac_lipschitz}
		L\le\prod_{i=1}^{k}c_i .
	\end{equation}
	If Eq.~(\ref{eq:layer_lipschitz}) holds only on a set that contains the intermediate states $t^{i-1}_1,t^{i-1}_2$ for all inputs in $U\subseteq\sR^d$, then Eq.~(\ref{eq:ac_lipschitz}) holds for $u,u',v\in U$. For a perceptron layer $f(x;\theta_i)=\sigma(W^{(i)}x+b^{(i)})$ with a $1$-Lipschitz coordinate-wise $\sigma$ and $W^{(i)}=[W^{(i)}_1,\ W^{(i)}_2]$, Eq.~(\ref{eq:layer_lipschitz}) holds with $\lambda^{(j)}_i=\norm{W^{(i)}_j}_2$ (spectral norm); with a residual connection, $f([h_1,h_2];\theta_i)=h_1+g([h_1,h_2];\theta_i)$, $\lambda^{(1)}_i$ is at most that of $g$ plus $1$. Without AC layers, vector addition is $1$-Lipschitz, and the Hadamard product satisfies $\norm{u\odot v-u'\odot v}\le\norm{v}_\infty\norm{u-u'}$.
\end{lemma}

Our operators use Transformer layers, and dot-product attention has no global Lipschitz constant~\citep{kim2021lipschitz}. Only the local form of the lemma applies, on a set $U$ that contains the embeddings and the intermediate results of Propositions~\ref{prop:chains} and~\ref{prop:assoc_distrib} (for $\dot{\times}$, whose digit classifier uses an arg max, the propositions compare a perturbed result only with the embedding of the same number, which the classifier reads alike while the perturbation is smaller than its margin), and Section~\ref{sec:expriments} reports empirical estimates of $L$. For $\dot{\times}$, Propositions~\ref{prop:chains} and~\ref{prop:assoc_distrib} thus hold under the additional condition that every perturbation they bound is smaller than the margin of this classifier at the slots of the exact result.

\begin{proposition}[Error accumulation in chained computations]\label{prop:chains}
	Let $M\ge 2$ and $a_1,\dots,a_M\in\Dn$, let $\dot{*}\in\{\dot{+},\dot{\times}\}$ be $L$-Lipschitz in its first argument with corresponding operation $*$ and defect $\varepsilon_*$, and suppose that all partial results $\bar a_j:=a_1*\cdots*a_j$ ($1\le j\le M$) lie in $\Dn$. Define the left fold $z_1:=\varphi(a_1)$ and $z_{j+1}:=z_j\,\dot{*}\,\varphi(a_{j+1})$. Then
	\begin{equation}\label{eq:chains}
		\norm{z_M-\varphi(\bar a_M)} \le \sum_{j=1}^{M-1} L^{\,M-1-j}\,\varepsilon_*(\bar a_j,a_{j+1}) \le \bar\varepsilon\sum_{i=0}^{M-2}L^{\,i},
	\end{equation}
	where $\bar\varepsilon:=\max_{1\le j\le M-1}\varepsilon_*(\bar a_j,a_{j+1})$. Consequently, the chained result is decoded correctly, $\varphi^\dagger(z_M)=\bar a_M$, whenever the middle term of Eq.~(\ref{eq:chains}) is smaller than $\rho(\bar a_M)$.
\end{proposition}

The bound grows at most linearly in $M$ if $L\le 1$ (and stays below $\bar\varepsilon/(1-L)$ if $L<1$), and geometrically if $L>1$. By Lemma~\ref{lem:ac_lipschitz}, $L$ can grow geometrically with the number $k$ of AC layers unless the layers are controlled; this is one reason why deeper operators do not by themselves improve the algebraic tests.

\begin{proposition}[Defects of the associative and distributive laws]\label{prop:assoc_distrib}
	Let $\dot{+}$ and $\dot{\times}$ be $L_+$- and $L_\times$-Lipschitz in each argument, and let $a,b,c\in\Dn$ be such that all pairs below are admissible. Then
	\begin{align}
		\norm{(\varphi(a)\dot{+}\varphi(b))\dot{+}\varphi(c)-\varphi(a)\dot{+}(\varphi(b)\dot{+}\varphi(c))}
		&\le L_+\big[\epsadd(a,b)+\epsadd(b,c)\big]+\epsadd(a+b,c)+\epsadd(a,b+c), \label{eq:assoc_bound}\\
		\norm{\varphi(a)\dot{\times}(\varphi(b)\dot{+}\varphi(c))-(\varphi(a)\dot{\times}\varphi(b))\dot{+}(\varphi(a)\dot{\times}\varphi(c))}
		&\le L_\times\,\epsadd(b,c)+\epsmul(a,b+c) \nonumber\\
		&\quad +L_+\big[\epsmul(a,b)+\epsmul(a,c)\big]+\epsadd(ab,ac). \label{eq:distrib_bound}
	\end{align}
	More precisely, the distances of the two sides of Eq.~(\ref{eq:assoc_bound}) to $\varphi(a+b+c)$ are at most $L_+\epsadd(a,b)+\epsadd(a+b,c)$ and $L_+\epsadd(b,c)+\epsadd(a,b+c)$, and the distances of the two sides of Eq.~(\ref{eq:distrib_bound}) to $\varphi(a(b+c))$ are at most $L_\times\epsadd(b,c)+\epsmul(a,b+c)$ and $L_+[\epsmul(a,b)+\epsmul(a,c)]+\epsadd(ab,ac)$. Hence each side is decoded as the correct result whenever its bound is smaller than $\rho(a+b+c)$, respectively $\rho(a(b+c))$. The analogue of Eq.~(\ref{eq:assoc_bound}) for multiplication holds with $L_\times$ and $\epsmul$.
\end{proposition}

Proposition~\ref{prop:assoc_distrib} shows why the distributive test is the hardest. Its bound contains three multiplicative defects, two of them scaled by $L_+$, and two additive defects, one scaled by $L_\times$ and one taken at the long results $ab$ and $ac$. The associative bound for addition contains only additive defects. Section~\ref{sec:expriments} reports the measured defects, the estimated Lipschitz constants and the resulting bounds.

\paragraph{Capacity of the embedding space} The decoding radii also limit how many numbers a $d$-dimensional embedding can represent robustly. If $\norm{\varphi(a)}\le B$ and $\rho(a)\ge\rho_{\min}$ for all $a\in\Dn$, then $N\le d\log_{10}(1+B/\rho_{\min})$ (Proposition~\ref{prop:capacity}, Appendix~\ref{app:capacity}). For $d=520$ and $N=25$ this only requires $\rho_{\min}\le B/(10^{N/d}-1)\approx 8.5\,B$, so the bound is no real constraint here; Section~\ref{subsubsec:ablation} measures how the dimension must grow with the number length.

\FloatBarrier
\section{Evaluation metrics}\label{sec:evaluation_criteria}

To the best of our knowledge, this is the first work to propose a method for constructing number embeddings that are trained to preserve algebraic properties of both addition and multiplication together with order, and that are evaluated property by property. As a result, it is necessary to establish specific evaluation criteria for assessing the performance of this model. In this section, we present internal evaluation metrics designed to measure the model's accuracy in number reconstruction, addition, and multiplication.

\begin{itemize}
	\item \textbf{Identity test:} The identity test criterion determines the model's ability to perform addition and multiplication operations where one of the operands is the identity element.

	\item \textbf{Invertibility test:} This test determines whether the addition and multiplication of the embedding of a number and the embedding of the inverse of that number results in the identity element. For multiplication, the test uses only the units of $\Dset$ (Lemma~\ref{lem:domain}), since other elements of $\Dset$ have no inverse in $\Dset$; in the field of fractions, the inverse of a pair is the swapped pair (Corollary~\ref{cor:fraction_field}), so the fraction multiplication test also covers division.

	\item \textbf{Closure test:} This test determines how closed the space created by the model is. For this purpose, we can check whether our model can reconstruct the addition and multiplication of two numbers embedded in the vector space or not. The identity test and the invertibility test are two special modes of this test.

	\item \textbf{Associative test:} This criterion specifies whether the model can maintain the associative property or not. For investigation, it is enough to add and multiply the embedding of three numbers with two different bracketing orders.

	\item \textbf{Distributive test:} This criterion is the same as the associative test criterion, with the difference that instead of examining the associative property, the distributive property is investigated.

	\item \textbf{Order tests:} These tests check whether the comparator predicts the correct relation, whether the induced relation is transitive, and whether it agrees with the operations (axioms (O1) and (O2) of Definition~\ref{def:ordered_domain}).

	\item \textbf{Fraction-field tests:} These tests apply the operations of Corollary~\ref{cor:fraction_field} to rational numbers outside $\Dset$, such as $1/3$.
\end{itemize}

\paragraph{Formal definitions}
All tests use test tuples drawn with fixed seeds, independently of the training and validation tuples (Section~\ref{subsec:distribution_and_numerical_dataset}); the pairs $(x,x^{-1})$ of the multiplicative invertibility test are kept out of training. For a test instance with target $a^\star\in\Dn$ and predicted decoder input $\hat z\in\sR^d$, the prediction is $\hat a:=\varphi^\dagger(\hat z)$. \emph{Exact Match} is $1(\hat a=a^\star)$, averaged over the test instances. \emph{Accuracy} is the share of characters of $\str{a^\star}$ that the decoded string reproduces at the same position; for the associative and distributive tests, it is averaged over the two bracketings. Table~\ref{tab:test_definitions} lists the tuples, the predicted decoder input and the target of each test. Each test uses $10{,}000$ tuples, except the multiplicative invertibility test (exhaustive: all $4{,}438$ units $a$ with $a,a^{-1}\in\Dn$), the transitivity test (the triples, out of $60{,}000$ drawn, for which the comparator predicts both premises in the same direction) and the (O2) test (the absolute values $(|a|,|b|)$ of $10{,}000$ admissible multiplication pairs, without those with a zero operand); Table~\ref{tab:main} gives the number of items of each test, and each point of the chain experiment uses $500$ chains. In the order test, 10\% of the pairs have $a=b$, so the relation ``equal'' is tested as well.

\begin{table}[htbp]
	\centering
	
	\caption{Formal definition of the algebraic tests. All tuples are drawn from the test distribution and conditioned on admissibility (Eq.~(\ref{eq:admissible})); $\dot{*}\in\{\dot{+},\dot{\times}\}$ denotes the operator under test and $e_+=0$, $e_\times=1$ its identity element.}
	\label{tab:test_definitions}
	\renewcommand{\arraystretch}{1.15}
	\footnotesize
		\begin{tabular}{@{}l >{\raggedright\arraybackslash}p{0.41\textwidth} >{\raggedright\arraybackslash}p{0.33\textwidth}@{}}
			\toprule
			\textbf{Test} & \textbf{Tested quantity} & \textbf{Target and success criterion} \\
			\midrule
			Identity & $\varphi(a)\,\dot{*}\,\varphi(e_*)$ & $a$ \\
			Invertibility ($+$) & $\varphi(a)\,\dot{+}\,\varphi(-a)$ & $0$ \\
			Invertibility ($\times$) & $\varphi(a)\,\dot{\times}\,\varphi(a^{-1})$, $a=\pm2^i5^j$ with $a,a^{-1}\in\Dn$ & $1$ \\
			Closure & $\varphi(a)\,\dot{*}\,\varphi(b)$ & $a*b$ \\
			Associative & $\hat z_L=(\varphi(a)\dot{*}\varphi(b))\dot{*}\varphi(c)$, $\hat z_R=\varphi(a)\dot{*}(\varphi(b)\dot{*}\varphi(c))$ & $\varphi^\dagger(\hat z_L)=\varphi^\dagger(\hat z_R)=a*b*c$ \\
			Distributive & $\hat z_L=\varphi(a)\dot{\times}(\varphi(b)\dot{+}\varphi(c))$, $\hat z_R=(\varphi(a)\dot{\times}\varphi(b))\dot{+}(\varphi(a)\dot{\times}\varphi(c))$ & $\varphi^\dagger(\hat z_L)=\varphi^\dagger(\hat z_R)=a\times(b+c)$ \\
			Chain of length $M$ & left fold $z_M$ of Proposition~\ref{prop:chains} & $\varphi^\dagger(z_M)=a_1*\cdots*a_M$ \\
			\midrule
			Order & $\bar\omega(\varphi(a),\varphi(b))$ & predicted relation equals the relation of $a$ and $b$ \\
			Antisymmetry consistency & $\omega(\varphi(a),\varphi(b))$, $\omega(\varphi(b),\varphi(a))$ & the two readings of the pair agree (raw comparator, before Eq.~(\ref{eq:symmetrization})) \\
			Transitivity & triples with $\varphi(a)<_{\bar\omega}\varphi(b)$ and $\varphi(b)<_{\bar\omega}\varphi(c)$ & $\varphi(a)<_{\bar\omega}\varphi(c)$ \\
			Compatibility (O1) & triples with $b<c$ & $\varphi(a)\dot{+}\varphi(b)<_{\bar\omega}\varphi(a)\dot{+}\varphi(c)$ \\
			Compatibility (O2) & pairs $(|a|,|b|)$ with $a,b\neq0$ & $\varphi(0)<_{\bar\omega}\varphi(|a|)\dot{\times}\varphi(|b|)$ \\
			\midrule
			Fraction addition & $[\varphi(p_1),\varphi(q_1)]\oplus[\varphi(p_2),\varphi(q_2)]$ & decoded numerator/denominator equal $p_1/q_1+p_2/q_2$ as rational numbers \\
			Fraction multiplication & $[\varphi(p_1),\varphi(q_1)]\otimes[\varphi(p_2),\varphi(q_2)]$ & decoded numerator/denominator equal $(p_1p_2)/(q_1q_2)$ as rational numbers \\
			Fraction order & Eq.~(\ref{eq:frac_ord}) & predicted relation equals the relation of $p_1/q_1$ and $p_2/q_2$ \\
			\bottomrule
		\end{tabular}
\end{table}

All results of Section~\ref{sec:expriments} use these success criteria.

The fraction-field tests use integers $p_1,p_2\in\Dn$ and $q_1,q_2\in\Dn\cap\sZ_{>0}$ with $\gcd(p_i,q_i)=1$ and $q_i$ divisible by a prime other than $2$ and $5$, so that $p_i/q_i\notin\Dset$. They thus test rational numbers that no single digit sequence can represent.
\par

\section{Experiments}\label{sec:expriments}

\subsection{Distribution and numerical dataset} \label{subsec:distribution_and_numerical_dataset}

In order to train the model with numerical data, a distribution over integers has been defined and used for sampling. To generate a number, two samples are drawn from this distribution: the first sample determines the length of the integer part, and the second sample determines the length of the decimal part. Subsequently, integer and decimal parts are randomly generated using uniform distributions based on their respective lengths. The sign is drawn independently, each sign with probability $1/2$, and the resulting string is converted to its canonical form $\str{a}$ (Section~\ref{subsec:number_representation}). Every generated number therefore lies in $\Dset$.

Theoretically, a probability distribution over number lengths should allow for the generation of any decimal fraction. Because number length is a discrete variable that can increase indefinitely, the distribution used should also be discrete and span the interval $[0, +\infty)$.\footnote{Note that a length of zero is required to omit either the integer part or the decimal part, and if both parts have a length of zero, resampling occurs. An integer part of length zero is written as the digit $0$ of the canonical string.} Such a sampler meets the full-support assumption of Section~\ref{sec:papdt}. If every length has positive probability, the digits are uniform and both signs have positive probability, every finite digit string has positive probability, and so has every element of $\Dset$ (of $\Dn$ if the strings are restricted to at most $N$ digits). The negative binomial, Poisson and shifted geometric distributions used below give every length a positive probability.

In practice, short numbers must stay frequent. The additive and multiplicative identities and the results of cancellations are short, and the product of two long numbers falls outside the represented range. A suitable length distribution is therefore discrete, covers $[0,+\infty)$ and favours small lengths.

\begin{figure}[htbp]
	\centering
	\includegraphics[width=\textwidth]{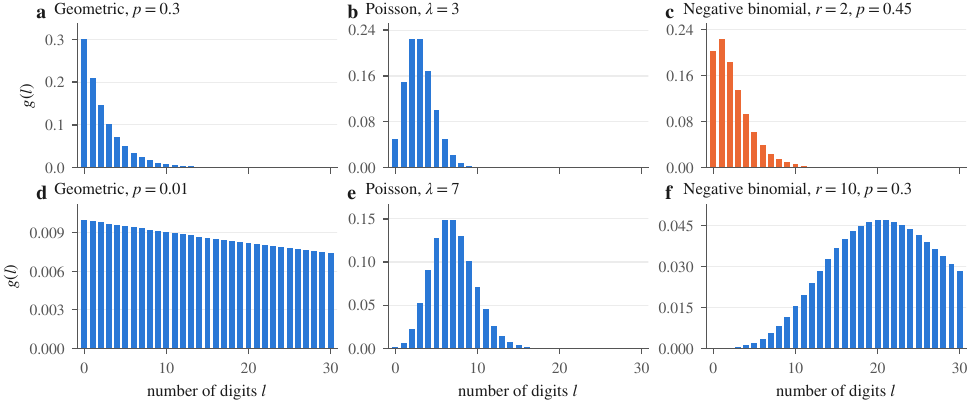}
	\caption{A selection of the tested distributions $g$ of the number of digits of the integer and of the fractional part; the selected distribution (c) is highlighted.}
	\label{fig:different_distribution}
\end{figure}

The candidate distributions for number lengths are: Poisson distribution, Negative Binomial distribution, and Geometric distribution (in the geometric distribution, we subtract one unit from each sampled length to account for a length of zero). By adjusting the parameters of these three types of distributions, over 30 different distributions were examined, a selection of which is illustrated in Figure \ref{fig:different_distribution}. In preliminary experiments, the bias towards smaller lengths yielded significantly better results on the algebraic metrics, and the best result was obtained for the negative binomial distribution with parameters $r_{\mathrm{NB}}=2$ and $p_{\mathrm{NB}}=0.45$ ($r$ and $p$ in Figure~\ref{fig:different_distribution}), that is, $g(l)=(l+1)\,p_{\mathrm{NB}}^{2}(1-p_{\mathrm{NB}})^{l}$ for $l\ge 0$; we keep this distribution fixed in all experiments. Training numbers are generated on the fly, half from this distribution and half with integer and fractional lengths drawn uniformly from $\{0,\dots,12\}$, which covers the long numbers that the negative binomial distribution rarely produces (we call these lengths the \emph{training lengths}). Validation and test tuples come from the negative binomial distribution with fixed seeds, separately from each other and from the training stream, and the validation tuples serve only for model selection.

\subsection{Experimental setup}\label{subsec:setup}

The model has 18.8M parameters and follows Section~\ref{sec:proposedmethod}; Table~\ref{tab:hparams} (Appendix~\ref{app:implementation}) lists all hyperparameters, the size of each component and the hardware. We train it with AdamW \emph{until convergence}: every 500 steps the model is evaluated on validation tuples, training stops when the mean validation exact match has not improved by more than 0.2 points in four consecutive evaluations (after 3000 to 12{,}000 steps), and the checkpoint with the best mean validation exact match is kept. A run takes about 2.2 hours on one NVIDIA L40S GPU. The test sets ($10^4$ tuples per test) are used only for the final results. Unless stated otherwise, the decoder chooses the output length with its length head (\emph{length-free} decoding); for comparison with Table~\ref{tab:performance}, we also report results with the true length given to the decoder (\emph{oracle length}).

\textbf{Runs and reported statistics.} The main configuration is trained with 3 random seeds (0, 1, 2) for initialization and data order; ``mean $\pm$ SD'' is the mean and the sample standard deviation over these runs. The ablations use seed 0. Differences between two models on the same test tuples are tested with McNemar's test \citep{mcnemar1947note} on the per-tuple exact-match outcomes, with Holm--Bonferroni correction within each table \citep{dror2018hitchhiker}. With $10^4$ test tuples, an exact match of 99.9\% has a 95\% Wilson interval of about $\pm0.07$ points. For each run, we keep the configuration, the complete training log (all loss terms, gradient norms and the learning rate every 50 steps, all validation evaluations), the checkpoints and the per-tuple test outcomes, and release them with the code.

\subsection{Results}\label{subsec:results}

\subsubsection{Algebraic metrics}

\begin{figure}[htbp]
	\centering
	
	\includegraphics[width=\textwidth]{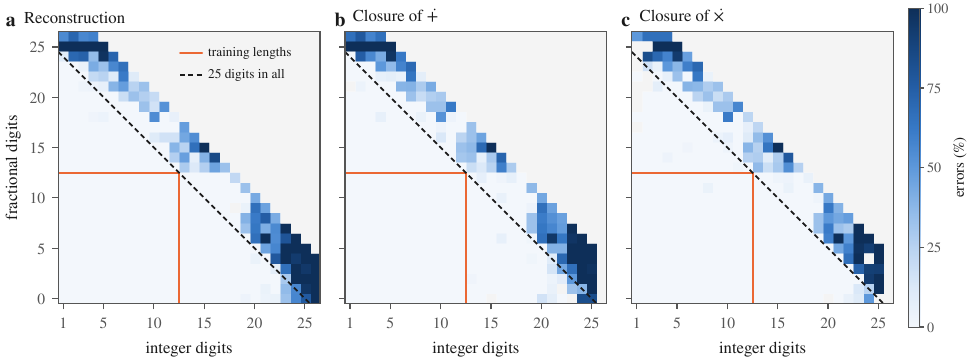}
	\caption{Errors of NIF by the shape of the number that the decoder must produce (the number itself for reconstruction, the result for the closure tests): percentage of test numbers with the given numbers of integer and fractional digits that are not decoded exactly (three seeds pooled, length-free decoding). For each total length from 1 to 30 digits, each seed is tested on 200 numbers or pairs with a random split into the two parts; shapes with fewer than six test numbers are grey. Solid line: training lengths (at most 12 digits per part); dashed line: 25 digits in total, the boundary of $\mathbb{D}_{25}$.}
	\label{fig:lengths}
\end{figure}

\begin{figure}[htbp]
	\centering
	
	\includegraphics[width=\textwidth]{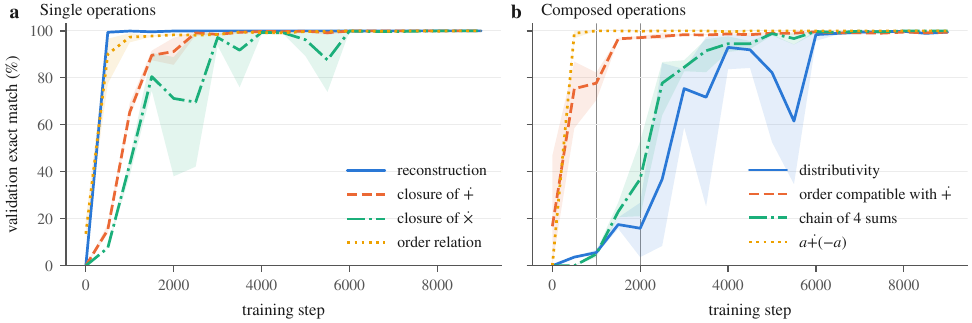}
	\caption{Validation exact match during training (mean of the three seeds; band: range over the seeds). (a) Single operations. (b) Composed operations and exact cancellations; the vertical lines mark the start of the targeted terms (step 1000) and of the compositions (step 2000; Table~\ref{tab:hparams}). All runs stopped after 8000--9000 steps by the convergence criterion of Section~\ref{subsec:setup}.}
	\label{fig:training}
\end{figure}

Table~\ref{tab:main} gives the results of all tests of Table~\ref{tab:test_definitions} for NIF, Figure~\ref{fig:lengths} the errors by number length, and Figure~\ref{fig:training} the validation metrics during training.

\begin{table}[htbp]
	\centering
	
	\caption{Results of NIF on the tests of Table~\ref{tab:test_definitions} (test sets, mean $\pm$ SD over 3 seeds).
	Accuracy is the token-level accuracy and EM the exact match (\%), both with the length chosen by the decoder (length-free);
	the last column gives the exact match when the decoder is given the true length (the protocol of Table~\ref{tab:performance}).
	$n$ is the number of test items. Chains: left folds of $M$ operands (Proposition~\ref{prop:chains}). Inverse (units): all units of $\Dn$; the units whose digit products fit the slot window reach 99.77\% (Section~\ref{subsec:results}). Results by number of digits: Figure~\ref{fig:lengths}.}
	\label{tab:main}
	\renewcommand{\arraystretch}{1.08}
	\footnotesize
	\begin{tabular}{@{}lrccc@{}}
		\toprule
		\textbf{Test} & $n$ & \textbf{Accuracy} & \textbf{EM} & \textbf{EM, oracle length} \\
		\midrule
				\multicolumn{5}{@{}l}{\textit{Reconstruction}} \\
		\quad Reconstruction & 10000 & 100.00\,$\pm$\,0.00 & 100.00\,$\pm$\,0.01 & 100.00\,$\pm$\,0.01 \\
		\multicolumn{5}{@{}l}{\textit{Addition}} \\
		\quad Identity & 10000 & 100.00\,$\pm$\,0.00 & 100.00\,$\pm$\,0.00 & 100.00\,$\pm$\,0.00 \\
		\quad Closure & 10000 & 100.00\,$\pm$\,0.00 & 100.00\,$\pm$\,0.01 & 100.00\,$\pm$\,0.01 \\
		\quad Inverse & 10000 & 100.00\,$\pm$\,0.00 & 100.00\,$\pm$\,0.01 & 100.00\,$\pm$\,0.01 \\
		\quad Associativity & 10000 & 100.00\,$\pm$\,0.00 & 99.98\,$\pm$\,0.02 & 99.98\,$\pm$\,0.02 \\
		\multicolumn{5}{@{}l}{\textit{Multiplication}} \\
		\quad Identity & 10000 & 100.00\,$\pm$\,0.00 & 100.00\,$\pm$\,0.00 & 100.00\,$\pm$\,0.00 \\
		\quad Closure & 10000 & 100.00\,$\pm$\,0.00 & 99.98\,$\pm$\,0.01 & 99.98\,$\pm$\,0.01 \\
		\quad Inverse (units) & 4438 & 92.62\,$\pm$\,0.16 & 85.22\,$\pm$\,0.31 & 85.23\,$\pm$\,0.33 \\
		\quad Associativity & 10000 & 100.00\,$\pm$\,0.00 & 99.94\,$\pm$\,0.02 & 99.94\,$\pm$\,0.02 \\
		\multicolumn{5}{@{}l}{\textit{Both operations}} \\
		\quad Distributivity & 10000 & 99.99\,$\pm$\,0.01 & 99.92\,$\pm$\,0.05 & 99.92\,$\pm$\,0.05 \\
		\multicolumn{5}{@{}l}{\textit{Order}} \\
		\quad Relation $\bar\omega$ & 10000 & -- & 99.79\,$\pm$\,0.10 & 99.79\,$\pm$\,0.10 \\
		\quad Antisymmetry of $\omega$ & 10000 & -- & 99.80\,$\pm$\,0.13 & 99.80\,$\pm$\,0.13 \\
		\quad Transitivity & 19959 & -- & 99.99\,$\pm$\,0.01 & 99.99\,$\pm$\,0.01 \\
		\quad Compatibility with $+$ (O1) & 10000 & -- & 99.44\,$\pm$\,0.24 & 99.44\,$\pm$\,0.24 \\
		\quad Compatibility with $\times$ (O2) & 9759 & -- & 100.00\,$\pm$\,0.00 & 100.00\,$\pm$\,0.00 \\
		\multicolumn{5}{@{}l}{\textit{Fraction field}} \\
		\quad Addition & 10000 & -- & 99.97\,$\pm$\,0.00 & 99.97\,$\pm$\,0.00 \\
		\quad Multiplication & 10000 & -- & 99.99\,$\pm$\,0.01 & 99.99\,$\pm$\,0.01 \\
		\quad Order & 10000 & -- & 96.54\,$\pm$\,0.39 & 96.54\,$\pm$\,0.39 \\
		\multicolumn{5}{@{}l}{\textit{Chains}} \\
		\quad Sums, $M=5$ & 500 & -- & 99.87\,$\pm$\,0.23 & 99.87\,$\pm$\,0.23 \\
		\quad Sums, $M=10$ & 500 & -- & 99.80\,$\pm$\,0.35 & 99.80\,$\pm$\,0.35 \\
		\quad Products, $M=5$ & 500 & -- & 100.00\,$\pm$\,0.00 & 100.00\,$\pm$\,0.00 \\
		\quad Products, $M=10$ & 500 & -- & 100.00\,$\pm$\,0.00 & 100.00\,$\pm$\,0.00 \\
		\bottomrule
	\end{tabular}
\end{table}

\paragraph{Single operations} With length-free decoding, NIF reconstructs 100.00\% of the test numbers exactly. The closure test reaches 100.00\% for $\dot{+}$ and 99.98\% for $\dot{\times}$, and the multiplicative identity and associative tests reach 100.00\% and 99.94\%. When the decoder is given the true length, reconstruction and these four operation tests reach 100.00\%, 100.00\%, 99.98\%, 100.00\% and 99.94\%, the same values. Products are longer than most training numbers. The objective therefore also trains the decoder on the results of the operations, and the staged operator computes products of any length with the same small-valued steps (Lemma~\ref{lem:small_sum}).

\paragraph{Number length} Since the sampler of Table~\ref{tab:main} rarely draws long numbers, Figure~\ref{fig:lengths} tests every length from 1 to 30 digits, split at random into integer and fractional digits, and shows the errors by the shape of the number to be decoded. Within the training lengths (at most 12 digits per part, apart from the tail of the negative binomial distribution), the model is exact (100.00\%, 100.00\% and 99.99\% for reconstruction and the closures of $\dot{+}$ and $\dot{\times}$). For numbers of $\mathbb{D}_{25}$ with a longer part, the three tests still reach 98.07\%, 98.58\% and 99.50\%, with errors only for numbers of close to 25 digits whose two parts differ greatly in length. Beyond 25 digits, the exact match falls slowly, to 53.53\%, 60.47\% and 63.83\% on average for 26--30 digits. Nearly all of these errors come from the length head: 98.6\% of the wrong reconstructions have a wrong number of integer or fractional digits, and with the true length the exact match of reconstruction rises to 92.5\%. The digits of long numbers are thus encoded and decoded correctly, and the range of exact lengths is set by the lengths that the training distribution covers (Section~\ref{subsubsec:ablation} examines how it scales with $N$).

\paragraph{Inverses} The inverse tests, $\varphi(a)\dot{+}\varphi(-a)$ and $\varphi(x)\dot{\times}\varphi(x^{-1})$, reach 100.00\% and 85.22\%. Their results, $0$ and $1$, are the shortest numbers, while the operands can have up to 25 digits, so every digit must cancel exactly. The multiplicative failures have one cause, the slot window: if $x$ and $x^{-1}$ have $f$ and $f'$ fractional digits, their digit products reach down to the place $10^{-(f+f')}$, but the lowest slot is $10^{-26}$. For the 3790 units with $f+f'\le 26$, the test is correct in 99.77\% of the cases. For the other 648 units, the columns below the window (whose sum is an exact carry) are not represented, and every model returns the product cut off at the window, for example $0.99999999999999999999999996$ (0.00\% correct, the same for all three seeds), which is also the training target of products that reach below the window (Table~\ref{tab:hparams}); exact arithmetic on the 52 places would reach only 85.40\%. With a wider window, the truncation disappears: the test reaches 100.00\% for each of the 6 models with $N=50$ and 100 ($K=102$ and 202 slots) of Section~\ref{subsubsec:ablation}, and between 99.19\% and 100.00\% for the 3 models with $N=200$. Exact cancellation must itself be trained: without the targeted pairs of Section~\ref{sec:papdt}, the test drops to 27.17\% (Figure~\ref{fig:profile}).

\paragraph{Laws, order and fractions} Associativity holds for 99.98\% ($\dot{+}$) and 99.94\% ($\dot{\times}$) of the triples and distributivity for 99.92\%. The order relation is correct for 99.79\% of the pairs, transitive in 99.99\% and compatible with addition (O1) and multiplication (O2) in 99.44\% and 100.00\% of the cases. Chains of 10 sums and of 10 products are exact in 99.80\% and 100.00\% of the cases. The fraction-field tests (Corollary~\ref{cor:fraction_field}), which involve rational numbers outside $\Dset$, reach 99.97\%, 99.99\% and 96.54\% for addition, multiplication and order, so arithmetic on rational numbers outside $\Dset$ is practically exact. Composing operations thus costs almost no accuracy; Section~\ref{subsec:error_propagation} explains this through the defects of the single operations. The only order test below 99\% is the order of fractions (96.54\%), which compares two outputs of $\dot{\times}$, $\varphi(p_1)\dot{\times}\varphi(q_2)$ and $\varphi(p_2)\dot{\times}\varphi(q_1)$ (Eq.~(\ref{eq:frac_ord})). Unlike the (O2) test, which compares a product with $\varphi(0)$, it compares two products, and most of its errors occur for pairs whose two products have the same sign and the same number of digits.

\subsubsection{Ablation study}\label{subsubsec:ablation}

\begin{figure}[htbp]
	\centering
	
	\includegraphics[width=\textwidth]{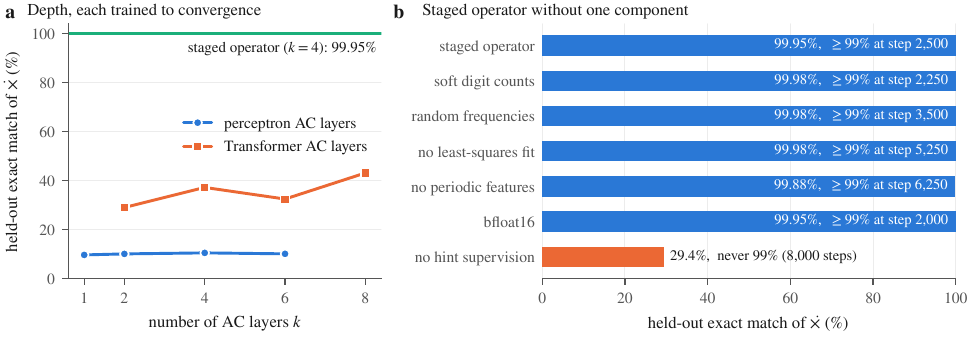}
	\caption{Operators trained alone until convergence on exact place-value inputs (4096 held-out pairs of the sampler). (a) Generic AC operators for $\dot{\times}$ with $k$ layers: perceptron layers $f(x)=\sigma(Wx+b)$ and slot-Transformer layers with a place-coupled bilinear branch, compared with the staged operator of Section~\ref{sec:staged}. (b) The staged operator without one of its components: best held-out exact match and the first evaluation step with at least 99\%. All numbers are listed in Table~\ref{tab:abl_operator} (Appendix~\ref{app:implementation}).}
	\label{fig:operators}
\end{figure}

\paragraph{Operators} To separate the operators from the encoder and the decoder, we train each operator alone until convergence on exact place-value inputs (the one-hot digits of the slots), with a linear output head (Figure~\ref{fig:operators}, Table~\ref{tab:abl_operator}); an operator that fails in this easier setting cannot be expected to work inside NIF. For multiplication, perceptron AC layers $f(x)=\sigma(Wx+b)$ reach 10.52\% exact match ($k=4$), generic Transformer AC layers with a place-coupled bilinear branch 32.49\% ($k=6$), and the staged operator 99.95\%. More depth does not make up for the missing structure: trained to convergence, perceptron AC operators stay at about 10\% for $k=1$ to $6$ (the share of products of one-digit operands), and generic Transformer AC operators improve only from 29.05\% ($k=2$) to 43.14\% ($k=8$). In the staged operator, hint supervision is essential (29.39\% without it), whereas the least-squares fit of the counts, the decimal frequencies and the periodic features only speed up training, by up to a factor of 2.5 (at least 99.88\% without any one of them); soft counts and bfloat16 change neither (Figure~\ref{fig:operators}b). For $\dot{+}$, slot-Transformer AC layers reach 100.00\% and perceptron layers 20.41\%.

\begin{figure}[htbp]
	\centering
	
	\includegraphics[width=\textwidth]{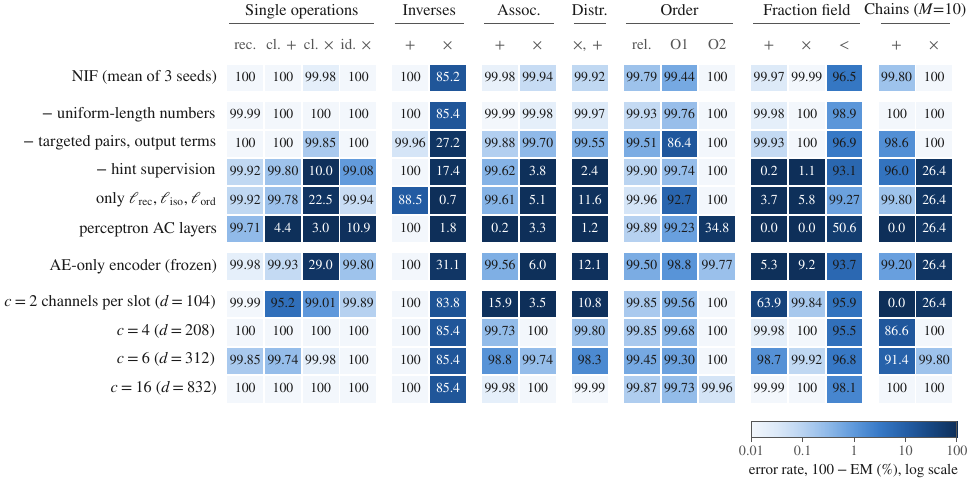}
	\caption{What each component of NIF is needed for. Each cell gives the exact match (\%, length-free decoding) on one test of Table~\ref{tab:main}; top row: NIF, mean of the three seeds; middle: NIF with one component removed or replaced; bottom: NIF with $c$ instead of 10 channels per place-value slot (seed 0, each trained until convergence). The shade encodes the error rate $100-\mathrm{EM}$ on a logarithmic scale, so that 99\% and 99.9\% differ as much as 90\% and 99\%. rec.: reconstruction; cl.: closure; id.: identity; rel.: order relation; O1, O2: compatibility of the order with $\dot{+}$ and $\dot{\times}$; $<$: order of fractions. The values of nine of these tests, of chains of five sums and the McNemar tests of their differences are given in Table~\ref{tab:abl_full} (Appendix~\ref{app:implementation}).}
	\label{fig:profile}
\end{figure}

\paragraph{Full model} Figure~\ref{fig:profile} removes one component of NIF at a time: training with only $\ell_{\mathrm{rec}}$, $\ell_{\mathrm{iso}}$ and $\ell_{\mathrm{ord}}$ isolates the additional terms of Section~\ref{sec:papdt}, the perceptron variant replaces only the Transformer layers $f$ of both operators by perceptron layers, and the variant without uniform lengths draws all training numbers from the sampler of Section~\ref{subsec:distribution_and_numerical_dataset}. The three basic losses alone are not enough for multiplication (closure 22.47\%, distributivity 11.61\%), and neither is the full objective without hint supervision (10.00\% and 2.37\%), with the same operator architecture. With perceptron AC layers, every operation fails while reconstruction and order stay intact, so the embedding alone does not carry the arithmetic. The staged operator in turn needs a clean digit code in the slots: it works on fixed one-hot slots (Table~\ref{tab:abl_operator}) but fails on products with the frozen AE-only encoder. The targeted pairs and the terms on operator outputs give the inverses and the agreement of the order with addition (27.17\% and 86.43\% without them). Training only on sampler numbers leaves the tests of Table~\ref{tab:main} unchanged, since their tuples come from the same sampler; the uniform half matters for numbers with more than 25 digits, where it raises the exact match of reconstruction from 43.30\% to 53.53\% (Table~\ref{tab:lengthdim}).

The last rows of Figure~\ref{fig:profile} vary the number $c$ of channels per slot. With $c=2$, reconstruction and order stay correct (99.99\% and 99.85\%), but the operations degrade (closure of $\dot{+}$ 95.22\%) and their compositions fail (associativity of $\dot{+}$ 15.90\%, distributivity 10.80\%). From $c=4$ on, single operations and laws are correct in at least 98.30\% of the cases (except the multiplicative inverses, limited by the slot window for every $c$, and the order of fractions), and long compositions still gain from more channels: chains of ten sums are exact in 86.60\%, 91.40\%, 99.80\% and 100.00\% of the cases for $c=4$, 6, 10 and 16. The quantities of Section~\ref{sec:approximate} explain both effects (Table~\ref{tab:geometry}). The decoding radius grows with the embedding norm $B=\sqrt{Kc}$ ($\hat\rho$ between $0.10\,B$ and $0.15\,B$, from 1.3 for $c=2$ to 3.4 for $c=16$), while the operator defects do not grow with $c$ (medians between 0.3 and 2.2). For $c=2$, the defect of $\dot{+}$ equals the radius (1.35 against 1.33), so the condition of Proposition~\ref{prop:decoding_error}(a) holds for only 48\% of the sums and a second operation leaves the decoding ball; for $c=4$, a single operation is safe (0.48 against 1.52), but a chain of ten sums comes close to the radius (median 1.07); for $c\ge10$, it stays well inside (1.6 against 2.8).

\paragraph{Number length and embedding dimension} The place-value slots tie the dimension to the number length, $d=Kc$ with $K=2N+2$, so the question is how many channels $c$ a place needs and whether this number grows with $N$. We train NIF for $N\in\{25,50,100,200\}$ and $c\in\{2,4,6,10\}$ ($d$ from 104 to 4020) for at most 12{,}000 steps (Section~\ref{subsec:setup}) and evaluate each run at its best validation checkpoint. Training numbers are drawn as for $N=25$ (Section~\ref{subsec:distribution_and_numerical_dataset}), with training lengths $\{0,\dots,\lfloor N/2\rfloor\}$ per part; for $N>25$, the validation numbers come from the same mixture, so model selection also sees long numbers, and smaller micro-batches with gradient accumulation keep 128 pairs per operation and step. The operator $\dot{\times}$ keeps its four layers: for $N>25$, $L_k\le 9N$ and $H_k\le 8N$ have three digits up to $N=111$ and four for $N=200$ (one more term each in $S_k$), and always $S_k\le 56$ and $T_k\le 14$, so the carries stay $0$ or $1$ by the argument of Lemma~\ref{lem:small_sum}; only the hint heads grow. Every model is tested as in Figure~\ref{fig:lengths}, on numbers with up to $1.2N$ digits (Figure~\ref{fig:lengthdim}, Table~\ref{tab:lengthdim} in Appendix~\ref{app:implementation}).

The results show three regimes. (i) Two channels are not enough for any $N$: the worst of the three tests reaches at most 91.4\% within the training lengths, and for $N\ge 50$ the closure of $\dot{\times}$ stays below 27\%. (ii) With $c\ge 4$, every model is practically exact within its training lengths: each test reaches at least 98.32\% for $N=25$, 50, 100 and 200, that is, for $d$ between 208 and 4020 (lowest value: closure of $\dot{\times}$ for $N=200$ and $c=6$). A place thus needs a fixed number of channels, and the dimension only has to grow linearly with the number length, $d=(2N+2)c$ with $c=4$ ($d=1608$ for $N=200$, close to the hidden size of a small language model); by Proposition~\ref{prop:capacity}, no embedding can do better than linear growth. (iii) Beyond the training lengths, the exact match falls in the same way for $N=25$, 50 and 100 when the longest part is measured relative to the trained part length (Figure~\ref{fig:lengthdim}b,c), and $c=4$ extrapolates best for $N=50$ and 100. For $N=200$, the decline starts earlier: on numbers of $\mathbb{D}_{200}$ with a longer part, reconstruction reaches 5.2\%, 12.9\% and 14.6\% for $c=4$, 6 and 10, and the closure of $\dot{\times}$ between 78.2\% and 83.5\%. As for $N=25$, most of these errors come from the predicted length: with $c=4$, on numbers with a longer part and $0.8N$ to $N$ digits, the true length raises reconstruction from 52.4\% to 98.1\% ($N=100$) and from 4.7\% to 76.4\% ($N=200$). Reconstruction falls faster than the closure of $\dot{\times}$ because the encoder only reads numbers within the training lengths, while the decoder is also trained on the longer results of the operations. The wider slot window also removes the truncation failures of the multiplicative inverse test. Runs take 5--8 hours ($N=50$) and 13--23 hours ($N=100$) on the GPU of Section~\ref{subsec:setup}, and 15 hours ($c=2$) and 54--67 hours ($c\ge4$) for $N=200$ on an NVIDIA A100.

\begin{figure}[htbp]
	\centering
	
	\includegraphics[width=\textwidth]{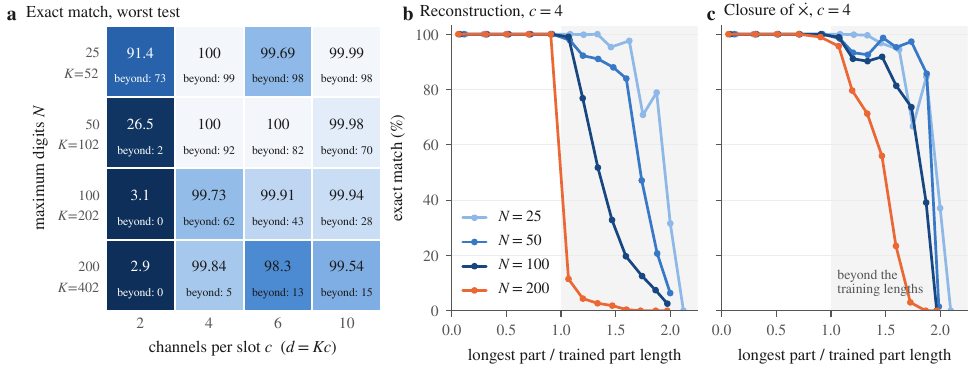}
	\caption{Number length and embedding dimension. (a) Exact match (\%) on the test numbers of $\mathbb{D}_N$ within the training lengths (at most 12 digits per part for $N=25$, $\lfloor N/2\rfloor$ otherwise), for the worst of reconstruction and the closures of $\dot{+}$ and $\dot{\times}$; shade: error rate on a logarithmic scale; small numbers: exact match on the numbers of $\mathbb{D}_N$ with a longer part. Row $N=25$: the models of Figure~\ref{fig:profile} ($c=10$: mean of the seeds), on the items of Figure~\ref{fig:lengths}. (b, c) Exact match against the longest part of the number to decode, relative to the trained part length, for $c=4$. All values are listed in Table~\ref{tab:lengthdim}.}
	\label{fig:lengthdim}
\end{figure}

\subsubsection{Defects, Lipschitz constants and error propagation}\label{subsec:error_propagation}
\begin{figure}[htbp]
	\centering
	
	\includegraphics[width=\textwidth]{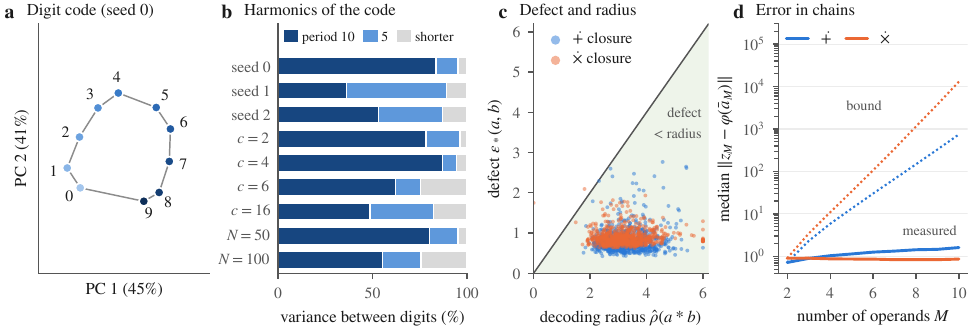}
	\caption{Geometry of the trained embedding. (a) Selected model (seed 0): the $c=10$ channels of one place-value slot for numbers whose digit at that place is $0,\dots,9$ (all other digits random), projected on their first two principal components; the points of the exponents $-3$, $0$, $3$ and $8$ coincide with their digit means (markers). (b) Share of the variance between the ten digit means that each harmonic of the digit index carries (period 10: a ring in the order $0,\dots,9$), for the three seeds, the embedding sizes of Figure~\ref{fig:profile} and $N=50$, 100 ($c=10$). (c) Defect $\varepsilon_*(a,b)$ against the estimated decoding radius of the result for 800 closure pairs of each operation; below the diagonal, the condition of Proposition~\ref{prop:decoding_error}(a) holds (shares over all seeds: Table~\ref{tab:geometry}). (d) Median distance of chained results from the embedding of the exact result (solid) and median of the bound of Eq.~(\ref{eq:chains}) with the empirical constants $\hat L$ (dotted; mean of the seeds, 500 chains per point).}
	\label{fig:geometry}
\end{figure}

To connect the theory of Section~\ref{sec:approximate} with the trained model, we measure the defects $\epsadd$ and $\epsmul$ on 2000 test tuples per test, estimate the decoding radius $\rho$ of the targets with the margin bound of Lemma~\ref{lem:margin}, and estimate the Lipschitz constants of the operators. For the radius, $L_\psi$ is the largest ratio $\norm{\psi_i(z)-\psi_i(\varphi(a))}_\infty/\norm{z-\varphi(a)}$ over 50 random perturbations $z$ of norm $\delta_0=6$ (the embeddings have norm $\sqrt{520}\approx22.8$); the margin includes the two length classifiers, so it covers length-free decoding. The constants $\hat L_+$ and $\hat L_\times$ are the largest ratios $\norm{u\,\dot{*}\,v-u'\,\dot{*}\,v}/\norm{u-u'}$ over 2000 test triples, which are lower bounds of the true constants; since radii and constants are estimated from samples, these shares are estimates, not guarantees; with $L_\psi$ taken over 50 directions, $\hat\rho$ tends to overestimate the radius. Figure~\ref{fig:geometry} shows the digit code of a slot and its harmonics, the defects of single tuples against their decoding radii, and the error in chains of $M=2,\dots,10$ operands (left folds, all partial results in $\Dn$; operands with 1--6 digits for sums and 1--3 digits for products).

\paragraph{What the encoder learned} Nothing in the architecture fixes how a slot represents its digit. In all three seeds, the code is the same at every exponent (cosine similarity of the centred codes of different exponents at least 0.99), does not depend on the other digits of the number (their influence is at most 0.06\% of the variance between the digits), and is low-dimensional (effective dimension 2.6--4.1 of the 10 channels); the same holds, with similar values, for $c=4$, 6 and 16 and for $N=50$ and 100. A discrete Fourier transform over the digit index shows that the two lowest harmonics, with periods 10 and 5, carry 87--95\% of the variance between the digit means of the three seeds (Figure~\ref{fig:geometry}b). In the selected model, the first harmonic alone carries 84\%, so the digits lie on a ring in the order $0,1,\dots,9$, closed between $9$ and $0$ (Figure~\ref{fig:geometry}a); in the other seeds, the second harmonic has a similar weight. The encoder thus learns a low-frequency Fourier code of the digit, which FoNE \citep{zhou2025foneprecisesingletokennumber} sets by hand and which has been found inside language models \citep{zhou2024fourier,kantamneni2025trigonometry}. The code breaks down where the operations fail: with $c=2$, the digit means lie on a line (effective dimension 1.0), and with $c=2$ or perceptron AC layers, the code differs between exponents (cosine similarity 0.53 and 0.57).

\paragraph{Defects and decoding radii} The defects of single operations are small compared with the decoding radii (all quantities in Table~\ref{tab:geometry}, Appendix~\ref{app:implementation}): with the estimated radii, the condition of Proposition~\ref{prop:decoding_error}(a) holds for 99.8\% of the sums and 99.9\% of the products, and the result lies in the decoding ball of a wrong number for at most 0.02\% of the tuples (Figure~\ref{fig:geometry}c). For the laws, Proposition~\ref{prop:assoc_distrib} gives worst-case bounds that multiply the defects by the empirical Lipschitz constants $\hat L_+\approx2.5$ and $\hat L_\times\approx3.9$; the actual errors are not amplified in this way, so both tests are almost always correct (Table~\ref{tab:main}), although the bounds hold for a smaller share of the triples (Table~\ref{tab:geometry}). Chains behave alike (Figure~\ref{fig:geometry}d): the measured error grows from about 0.7 to 1.6 over ten sums and stays near 0.85 for ten products, while the bound of Eq.~(\ref{eq:chains}) grows geometrically. The analysis of Section~\ref{sec:approximate} thus identifies the quantities that control the errors and gives worst-case bounds for composed operations.

\subsubsection{Comparison with existing work}\label{sec:comparison}

To the best of our knowledge, there are no existing works in the literature that propose a method for constructing number embeddings with the explicit aim of preserving algebraic properties such as addition, multiplication, and their associated characteristics (e.g., identity, invertibility, closure, associativity, and distributivity) in the embedding space. As a result, the proposed embedding cannot be directly compared against prior methods using algebraic metrics. Nevertheless, to compare our work, we used the embedding methods LUNA \citep{han2023lunalanguageunderstandingnumber}, FoNE \citep{zhou2025foneprecisesingletokennumber},  xVal~\citep{golkar2023xval} mentioned in Section \ref{subsec:number_embeddings} as baselines, and the comparison results are presented in Table \ref{tab:performance}. We perform the comparison by evaluating the accuracy and exact match of addition and multiplication operations with varying lengths on the embedding methods. To isolate the effect of the algebraic losses, we also report an autoencoder with the NIF architecture trained with $\ell_{\mathrm{rec}}$ only (AE-only). Its encoder is then frozen, like the baseline encoders, and the decoder, operators and comparator of NIF are trained with the full objective.

To ensure a fair and consistent comparison, the data distribution described in Section~\ref{subsec:distribution_and_numerical_dataset} is employed to generate the datasets used for training the embedding vectors. Moreover, each baseline encoder replaces the NIF encoder and is kept fixed, and only the decoder, the addition and multiplication operators (ANOs), and the neural order module are trained for each embedding on the same data. The baseline embeddings have no place-value slots. For them, the decoder receives copies of the embedding with positional encodings, and the operators are perceptron AC layers. The staged operator of Section~\ref{sec:staged} reads the digit of each place from a slot and is therefore not used for them. The comparison thus covers each embedding together with the decoder and operators of this protocol, not the embedding alone. Neural arithmetic units (Section~\ref{subsec:arithmetic_operations_in_neural_networks}) compute on raw scalars and give no embedding, so this protocol does not apply to them. The NIF row reports our model under the same protocol (decoder given the true length; closure and order pairs drawn with maximum length $N$). As shown in Table~\ref{tab:performance}, under this protocol, NIF has the highest accuracy and exact match in every setting. The exact match of both closure tests is at least 99.98\% for all three maximum lengths, while the best baseline, LUNA, reaches at most 91.96\% for addition and 61.45\% for multiplication; the order relation is correct for more than 99.7\% of the pairs. The sampler rarely draws numbers with more than 25 digits, so the column $N=30$ contains few numbers outside the training range; Figure~\ref{fig:lengths} tests longer numbers directly. With the same architecture and trained operators, the AE-only encoder matches NIF for addition (exact match at least 99.95\%) and for the order (at least 99.34\%), but reaches only 29.00--43.66\% for multiplication (Figure~\ref{fig:profile}). Multiplication therefore needs a clean code of the digit in every slot. Fixed one-hot slots give it (Table~\ref{tab:abl_operator}), the AE-only encoder does not, and the encoder of NIF learns it when it is trained jointly with the operators and the algebraic losses. For the order task, both NIF and LUNA demonstrate strong results, with NIF slightly outperforming LUNA across all settings. Notably, xVal shows significant performance degradation, especially on longer numbers, as expected, since xVal represents a number as a scalar multiple of a fixed vector, and the precision of this representation falls as the magnitude grows. Section~\ref{subsubsec:reader} also compares NIF with FoNE and one-hot digit codes under identical conditions.

\begin{table}[htbp]
	\centering
	\caption{Performance of different models on addition, multiplication, and order tasks across varying maximum number lengths $N$ (10, 20, and 30). Results are reported in terms of Accuracy and Exact Match (\%); NIF: mean $\pm$ SD over 3 seeds; $N$ is the maximum length of the sampled operands; for the order, the single value is the share of pairs with the correct relation; baselines: decoder without slots and perceptron AC operators (the staged operator needs place-value slots; Section~\ref{subsubsec:reader} compares the embeddings under identical conditions); AE-only: NIF with an encoder trained with $\ell_{\mathrm{rec}}$ only and then kept fixed, one run.}
	\renewcommand{\arraystretch}{1.2}
	\resizebox{\textwidth}{!}{
		\begin{tabular}{@{}l c *{3}{cc}@{}}
			\toprule
			& \textbf{Model}
			& \multicolumn{2}{c}{\textbf{$N=10$}}
			& \multicolumn{2}{c}{\textbf{$N=20$}}
			& \multicolumn{2}{c}{\textbf{$N=30$}} \\
			\cmidrule(lr){3-4} \cmidrule(lr){5-6} \cmidrule(lr){7-8}
			&
			& \textbf{Accuracy} & \textbf{Exact Match}
			& \textbf{Accuracy} & \textbf{Exact Match}
			& \textbf{Accuracy} & \textbf{Exact Match} \\
			\midrule
			\multirow{5}{*}{\textbf{Addition}}
			& \textbf{LUNA \citep{han2023lunalanguageunderstandingnumber}} & 99.47\,$\pm$\,0.31 & 90.45\,$\pm$\,0.48 & 99.30\,$\pm$\,0.17 & 91.96\,$\pm$\,0.42 & 97.04\,$\pm$\,0.36 & 79.28\,$\pm$\,0.79 \\
			& \textbf{FoNE \citep{zhou2025foneprecisesingletokennumber}} & 98.97\,$\pm$\,0.51 & 86.94\,$\pm$\,0.82 & 98.58\,$\pm$\,0.68 & 82.97\,$\pm$\,0.51 & 97.84\,$\pm$\,0.36 & 73.78\,$\pm$\,0.99 \\
			& \textbf{xVal \citep{golkar2023xval}} & 32.15\,$\pm$\,1.14 & 13.87\,$\pm$\,3.14 & 9.00\,$\pm$\,2.14 & 1.14\,$\pm$\,0.99 & 0 & 0 \\
			& \textbf{AE-only (NIF arch., $\ell_{\mathrm{rec}}$ only)} & 100.00& 99.97& 99.99& 99.95& 100.00& 99.99\\
			& \textbf{NIF (ours)} & 100.00\,$\pm$\,0.00& 100.00\,$\pm$\,0.00& 100.00\,$\pm$\,0.00& 100.00\,$\pm$\,0.00& 100.00\,$\pm$\,0.00& 99.99\,$\pm$\,0.01\\
			\midrule
			\multirow{5}{*}{\textbf{Multiplication}}
			& \textbf{LUNA} & 84.63\,$\pm$\,2.16 & 61.45\,$\pm$\,1.51 & 75.42\,$\pm$\,2.14 & 40.87\,$\pm$\,2.87 & 69.97\,$\pm$\,1.77 & 22.87\,$\pm$\,1.56 \\
			& \textbf{FoNE} & 71.76\,$\pm$\,1.06 & 32.88\,$\pm$\,2.16 & 65.49\,$\pm$\,1.44 & 20.84\,$\pm$\,2.74 & 59.37\,$\pm$\,2.95 & 12.24\,$\pm$\,2.67 \\
			& \textbf{xVal} & 8.44\,$\pm$\,3.12 & 0.14\,$\pm$\,0.09 & 0 & 0 & 0 & 0 \\
			& \textbf{AE-only} & 84.76& 43.66& 75.60& 29.65& 75.18& 29.00\\
			& \textbf{NIF (ours)} & 100.00\,$\pm$\,0.00& 100.00\,$\pm$\,0.00& 100.00\,$\pm$\,0.00& 100.00\,$\pm$\,0.01& 100.00\,$\pm$\,0.00& 99.98\,$\pm$\,0.01\\
			\midrule
			\multirow{5}{*}{\textbf{Order}}
			& \textbf{LUNA} & 98.13\,$\pm$\,0.25 & -- & 97.24\,$\pm$\,0.49 & -- & 96.17\,$\pm$\,0.55 & -- \\
			& \textbf{FoNE} & 96.06\,$\pm$\,0.44 & -- & 94.47\,$\pm$\,0.26 & -- & 93.78\,$\pm$\,0.88 & -- \\
			& \textbf{xVal} & 28.18\,$\pm$\,1.25 & -- & 17.55\,$\pm$\,2.18 & -- & 7.65\,$\pm$\,2.66 & -- \\
			& \textbf{AE-only} & 99.46& -- & 99.34& -- & 99.60& -- \\
			& \textbf{NIF (ours)} & 99.74\,$\pm$\,0.09& -- & 99.78\,$\pm$\,0.07& -- & 99.79\,$\pm$\,0.08& -- \\
			\bottomrule
		\end{tabular}
		\label{tab:performance}
	}
\end{table}

\subsubsection{Place-coupled reading of number vectors}\label{subsubsec:reader}
The place-value slots let a sequence model read a number vector at the place it is currently computing. In a controlled setting, we test whether this kind of reading carries over to numbers longer than those seen in training.

\paragraph{Model and reading} We train a decoder-only Transformer (4 layers, width 192, 4 heads, rotary position encoding \citep{su2024roformer}) from scratch on problems $a+b=$ and $a-b=$. The non-negative integers are written as digit tokens, and the model writes the result digit by digit, units digit first. Each operand is also given as a single vector computed by a frozen encoder. Let $z_{[p]}(a)\in\sR^{c}$ be the slot of $\varphi(a)$ for the exponent $p$. At the position that writes the digit of place $p$, we add to the output of the first layer
\begin{equation}\label{eq:reader}
	r_p=\sum_{j\in\{0,1\}}\left(W^{(1)}_j\, z_{[p-j]}(a)+W^{(2)}_j\, z_{[p-j]}(b)\right)+u_{[p\ge n_a],[p\ge n_b]},
\end{equation}
where $n_a$ and $n_b$ are the numbers of digits of the operands, $u$ is one of four learned vectors, and the terms with $j=1$ are dropped for $p=0$. The matrices $W^{(i)}_j\in\sR^{192\times c}$ are shared by all places, so no trained weight depends on the place. The place $p$ and the two indicators follow from the token positions, and the vectors enter the model only through Eq.~(\ref{eq:reader}). The Transformer computes the digits and carries itself; the operators and decoder of NIF are not used.

\paragraph{Setup} The training set has 30{,}000 problems. The number of digits of the longer operand is uniform in $\{1,\dots,12\}$, that of the other operand is uniform up to this length, the operation and the order of the operands are random, and differences are non-negative. All models see the same problems for 4000 steps (AdamW, batch size 128, learning rate $10^{-3}$ with warm-up and cosine decay). We test them on 200 problems for each of six lengths of the longer operand, three of which exceed the training lengths. We compare five inputs under identical conditions (same model, data, training and test): digit tokens alone, and digit tokens plus vectors read by Eq.~(\ref{eq:reader}) from (i) NIF (seed 0 of Table~\ref{tab:main}, $c=10$); (ii) FoNE \citep{zhou2025foneprecisesingletokennumber} with the features $\cos$ and $\sin$ of $2\pi\,(x \bmod 10^{p+1})/10^{p+1}$ for each place $p$, computed with exact digit arithmetic ($c=2$); (iii) the same features computed from the number as a float64, which holds integers exactly only up to $2^{53}\approx 9\times10^{15}$; and (iv) the one-hot digit of each place ($c=10$).

\begin{table}[htbp]
	\centering
	
	\caption{Exact match (\%) of a Transformer that adds and subtracts integers, by the number of digits of the longer operand; the training operands have at most 12 digits. The number vectors are read place by place (Section~\ref{subsubsec:reader}); 200 test problems per length, one run.}
	\label{tab:reader}
	\renewcommand{\arraystretch}{1.08}
	\footnotesize
	\begin{tabular}{@{}lcccccc@{}}
		\toprule
		& \multicolumn{3}{c}{\textbf{Training lengths}} & \multicolumn{3}{c}{\textbf{Longer than in training}} \\
		\cmidrule(lr){2-4} \cmidrule(lr){5-7}
		\textbf{Input of the model} & 4 & 8 & 12 & 14 & 16 & 20 \\
		\midrule
		Digit tokens only & 98.5 & 99.0 & 92.5 & 0.0 & 0.0 & 0.0 \\
		+ one-hot digits, read place by place & 99.0 & 99.0 & 99.0 & 100.0 & 98.5 & 89.0 \\
		+ FoNE (float64), read place by place & 99.5 & 99.0 & 98.5 & 99.0 & 95.0 & 0.0 \\
		+ FoNE (exact), read place by place & 98.5 & 99.5 & 98.0 & 98.0 & 95.0 & 88.5 \\
		+ NIF (ours), read place by place & 99.0 & 99.5 & 99.5 & 99.5 & 98.0 & 92.5 \\
		\bottomrule
	\end{tabular}
\end{table}

\paragraph{Results} With digit tokens alone (Table~\ref{tab:reader}), the model solves problems of the training lengths (92.5\% at 12 digits) but none of the longer ones. When it reads the NIF vectors place by place, it reaches 99.5\% at 14 digits, 98.0\% at 16 digits and 92.5\% at 20 digits, eight digits beyond the longest training operands. At every length, NIF does as well as exact FoNE and the one-hot digits (88.5\% and 89.0\% at 20 digits). FoNE computed in float64 gives 0.0\% at 20 digits, because the low-order digits of the operands are lost. A sequence model can therefore read the learned NIF slots, which were trained for the operators of Section~\ref{sec:proposedmethod}, as reliably as explicit digits, using reading maps learned on short numbers and applied to longer ones.

\subsection{Downstream evaluation with a language model}\label{subsec:downstream}
We now test whether the operators of NIF are also useful inside a larger model. We insert the frozen NIF model (encoder, $\dot{+}$, $\dot{\times}$ and decoder of seed 0 in Table~\ref{tab:main}) into a language model and evaluate it on word problems, mainly with numbers longer than those in common benchmarks.

\paragraph{Operator layers} The base model is Qwen3.5-0.8B\footnote{\texttt{https://huggingface.co/Qwen/Qwen3.5-0.8B} (accessed November 2026).} (752M parameters, 24 decoder layers); the problem text is unchanged. After the question, a workspace lists each number $a$ of the prompt as a value token between two tag tokens, with input embedding $\varphi(a)$ passed through a trained adapter. The numbers and all results computed so far are kept as NIF vectors, called registers. After decoder layers 9 and 15, an operator layer reads the hidden states of the two tags of a number with vector $y$ and outputs pointers, each selecting a register $u$ or nothing. The frozen operators compute the sum $u\,\dot{+}\,y$ or a difference (with the vector of $-y$ or $-u$, that is, $\varphi(-a)$ for a number $a$ of the prompt and the product with $\varphi(-1)$ for a computed result) and the product $u\,\dot{\times}\,y$; the results become new registers and are added to the hidden states through gated adapters. Each number written in the answer is encoded by $\varphi$ and gets a sum and a product with an earlier register in the same way. The answer is decoded from a mixture of the next-token distribution of the language model and the next characters of the registers, whose digits come from the NIF decoder \citep{see2017get}. The language model thus only chooses among vectors computed by NIF, and no gradient reaches NIF. The operator layers add 10.7M parameters (1.4\% of the base model); Appendix~\ref{app:implementation} gives further details.

\paragraph{Training} All weights of the base model are trained with the adapters for 6{,}000 steps (AdamW, batch size 32, learning rate $10^{-5}$ for the base model and $10^{-3}$ for the adapters, 100 warm-up steps, cosine decay); NIF stays frozen. The loss is the negative log-likelihood of the reference text under the mixture and under the language model alone, plus that of the correct pointers, which need no annotation: a choice of registers is correct if its sum, difference or product is a number of the reference text (for the numbers written in the answer, one of the next three). The training mixture consists of arithmetic problems (40\%: sums, differences and comparisons of numbers with up to 12 digits, products of factors with up to 6 digits, and chains of two to four additions and subtractions), addition and subtraction word problems from 16 templates with step-by-step solutions (15\%), and correct step-by-step solutions written by the base model itself before training: 7{,}508 solutions for 4{,}815 of the 7{,}473 GSM8K training problems \citep{cobbe2021trainingverifierssolvemath} (15\%) and solutions for further word problems from various domains (30\%). These come from one greedy and one sampled pass (temperature 0.7), with at most two different correct solutions per problem, so the targets look like the base model's own answers. Training takes 3.2 hours on a 71\,GB partition of an NVIDIA H200 GPU.

\paragraph{Test sets} GSM-Ranges \citep{shrestha2025gsmranges} replaces the numbers of 100 GSM8K test problems (the original problems, not templates) at six levels, from the original number of digits (level 1) to numbers between $10^6$ and $10^7$ (level 6); we use two variants of each problem per level. We also fill 8 held-out templates of addition and subtraction word problems with numbers of 1 to 20 digits (100 problems for each of the eleven lengths 1, 2, 3, 4, 6, 8, 10, 12, 14, 16 and 20). Both models are asked to solve each problem step by step and to end with the final number; they decode greedily (at most 512 tokens), and an answer is correct if its final number equals the reference value.

We do not report the GSM8K test split, because its numbers are short: 96\% of the numbers in its questions have at most three digits (1.7 digits on average), and 97\% of the intermediate and final results have at most four. At these lengths, the answers of our model do not depend on the number representation; switching off the operator layers changes the exact match on GSM-Ranges level 1, which keeps the GSM8K number lengths, from 55.0\% to 55.5\% (see below). Such a benchmark cannot evaluate a number representation. GSM-Ranges uses the same problems and varies only the size of their numbers, which is the factor studied here.

\begin{table}[htbp]
	\centering
	
	\caption{Exact match (\%) of the final answer on word problems (Section~\ref{subsec:downstream}): the language model as released (base model) and after training with the NIF operator layers, with the same prompts. Operators off: the model trained with NIF, with the operator layers and register outputs switched off (Figure~\ref{fig:llmcausal}). $p$: exact McNemar test of the base model and the model trained with NIF on the same problems, adjusted with the Holm--Bonferroni procedure over the rows; bold: $p<0.05$.}
	\label{tab:llm}
	\renewcommand{\arraystretch}{1.08}
	\footnotesize
	\begin{tabular}{@{}lrcccc@{}}
		\toprule
		\textbf{Test set} & \textbf{Problems} & \textbf{Base model} & \textbf{Trained with NIF} & \textbf{Operators off} & $p$ \\
		\midrule
		GSM-Ranges, level 1 & 200 & 59.5 & 55.0 & 55.5 & 0.92 \\
		GSM-Ranges, level 2 & 200 & 53.5 & 53.5 & 53.0 & 1.00 \\
		GSM-Ranges, level 3 & 200 & 43.0 & 46.5 & 46.0 & 0.92 \\
		GSM-Ranges, level 4 & 200 & 39.0 & 44.5 & 45.0 & 0.77 \\
		GSM-Ranges, level 5 & 200 & 29.5 & 43.0 & 42.5 & \textbf{0.004} \\
		GSM-Ranges, level 6 & 200 & 28.0 & 43.0 & 40.5 & \textbf{$<$0.001} \\
		\midrule
		Templated problems, numbers of 1--12 digits & 800 & 54.2 & 85.1 & 85.2 & \textbf{$<$0.001} \\
		Templated problems, numbers of 14--20 digits & 300 & 1.3 & 39.7 & 12.0 & \textbf{$<$0.001} \\
		\bottomrule
	\end{tabular}
\end{table}

\begin{figure}[htbp]
	\centering
	
	\includegraphics[width=\textwidth]{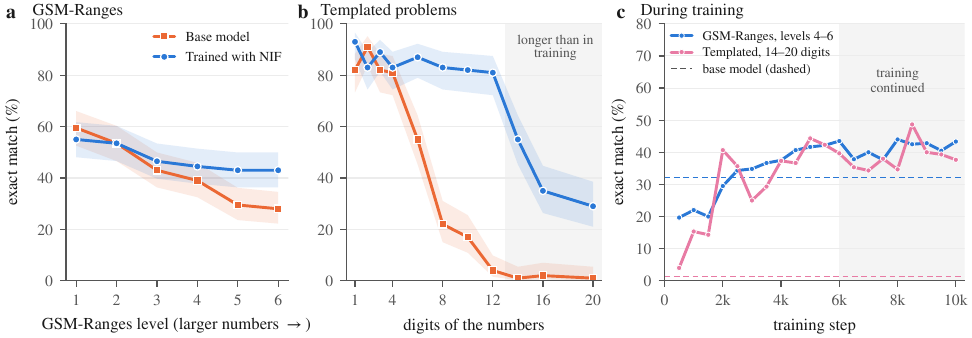}
	\caption{Base model and model trained with the NIF operator layers. (a) GSM-Ranges by level (level 1: the GSM8K number lengths; level 6: numbers between $10^6$ and $10^7$). (b) Templated addition and subtraction problems by number of digits; shaded: longer than any number in training. Bands: 95\% Wilson intervals. (c) Exact match of the model with NIF during training (full test every 500 steps; from step 6{,}000 on, training continues with a second learning-rate cycle) on GSM-Ranges levels 4--6 and on templated problems with 14 to 20 digits; dashed: base model.}
	\label{fig:llm}
\end{figure}

\paragraph{Results} On GSM-Ranges (Table~\ref{tab:llm}, Figure~\ref{fig:llm}a), the base model drops by 32 points from level 1 to level 6 and the model with NIF by 12; at levels 5 and 6, the model with NIF solves 43.0\% and 43.0\% of the problems and the base model 29.5\% and 28.0\% ($p=0.004$ and $p<0.001$). On the templated problems (Figure~\ref{fig:llm}b), the model with NIF solves 85.1\% of the problems with numbers of the training lengths (base model 54.2\%) and 39.7\% of those with 14 to 20 digits (base model 1.3\%, $p<0.001$). During training (Figure~\ref{fig:llm}c), it is ahead of the base model on the 14- to 20-digit templated problems from the first test at 500 steps on, and on GSM-Ranges levels 4--6 from 2{,}500 steps on. Training for 10{,}000 instead of 6{,}000 steps changes no row of Table~\ref{tab:llm} significantly.

\IfFileExists{fig_llm_causal.pdf}{%
\begin{figure}[htbp]
	\centering
	
	\includegraphics[width=\textwidth]{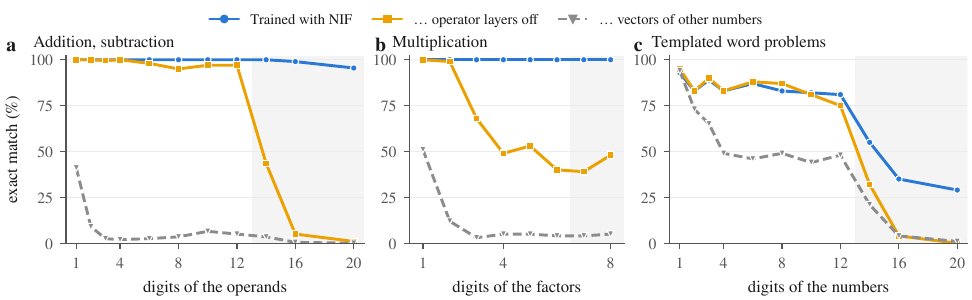}
	\caption{Interventions on the model trained with NIF (step 6{,}000): exact match (\%) as trained, with the operator layers and register outputs switched off (the language model alone writes the answer), and with the vectors of other numbers of the same problem in the workspace. (a) Addition and subtraction, (b) multiplication and (c) templated word problems, by number of digits; shaded: longer than in training (multiplication: factors of up to 6 digits).}
	\label{fig:llmcausal}
\end{figure}}{}

\paragraph{Switching off the operators} Figure~\ref{fig:llmcausal} repeats the tests with two modified versions of the trained model (Holm-adjusted McNemar tests against the model as trained): without the operator layers, NIF vectors and register outputs, so that the trained language model writes the answer alone, and with the vectors of other numbers of the same problem in the workspace. With these wrong vectors, the exact match drops to 2.3\% for additions of 14 to 20 digits and to 13.3\% for products of factors with up to six digits, so the model uses the values stored in the NIF vectors. Without the operator layers, the language model still adds and subtracts numbers of the training lengths (98.0\% and 98.6\%) but fails on longer ones: for 14 to 20 digits, 18.0\% of the sums and 15.0\% of the differences are correct, against 98.7\% and 97.7\% with the operator layers. Its products are correct for 68.2\% of the factors with up to six digits and 43.5\% of those with 7 or 8 digits, against 100.0\% and 100.0\% (all $p<0.001$). Word problems with 14 to 20 digits behave alike (39.7\% with and 12.0\% without the operator layers, $p<0.001$). On GSM-Ranges, whose numbers have at most seven digits, both versions give the same results (43.5\% and 42.7\% at levels 4--6). Switching off the operators thus lowers the accuracy mainly for products and for numbers of 14 to 20 digits.

\FloatBarrier
\section{Conclusion}\label{sec:conclusion}
In this study, we presented a novel approach for constructing number embeddings that preserve algebraic properties, focusing on addition and multiplication operations. Since finite digit sequences represent exactly the decimal fractions, we took their ordered ring as the target structure and proved that, under ideal zero-loss conditions, the embedding space with the learned operators and the antisymmetric neural comparator is an ordered integral domain isomorphic to it, whose field of fractions is an ordered field isomorphic to $\sQ$. For the trained model, whose losses are not zero, we derived bounds that link decoding margins, algebraic defects and the Lipschitz constants of the operators to failures of single operations, of chained computations and of the associative and distributive laws. We also introduced specific evaluation metrics to assess the model's ability to maintain these properties. With perceptron AC layers, multiplication is the weak point of such a model. NIF reads the embedding as place-value slots, also trains the decoder on the results of the operations, and computes products with a staged operator whose steps have values as small as those of an addition and are supervised in training. With length-free decoding, it reaches 99.98\% exact match on the multiplicative closure test, 99.98\% on the associative test of addition and 99.92\% on the distributive test. The ablations identify the components behind this improvement, and the measured defects and Lipschitz constants link the remaining failures of composed computations to the bounds of Section~\ref{sec:approximate}. Inside a language model, the frozen operators compute sums, differences and products for the model: with NIF operator layers, Qwen3.5-0.8B solves 98.7\% of additions and 100.0\% of multiplications beyond the training lengths and 39.7\% of word problems with numbers of 14 to 20 digits; with the operator layers switched off, the same model gives 18.0\%, 43.5\% and 12.0\%.

The results show that a learned embedding space can reproduce most of the algebraic structure of numbers, including multiplication. The operators need a clean code of the digit in every place-value slot. Fixed one-hot slots give it, whereas with an encoder trained for reconstruction only, the same operators fail on products. Without any prescribed digit code, the encoder learns a low-frequency Fourier code of the digit that is the same in every place-value slot; in the selected model, the digits lie on a ring. Four channels per place, fewer than the ten of a one-hot digit code, are enough for reconstruction and both operations on numbers with up to 200 digits whose parts stay within the trained lengths, so the embedding dimension only has to grow linearly with the number length. The slot window decides which products are represented exactly (multiplicative inverse test 85.22\% with $K=52$ slots, at least 99.19\% with $K\ge102$); and since the digits of longer numbers are decoded correctly once their length is known (92.5\% for 26--30 digits when trained up to 25), the length head sets the range of exact lengths. The trained model approximates the ideal structure, and the tests and bounds of this paper measure how close the approximation is.

\FloatBarrier
\section{Future work}\label{sec:future_works}

Future work will focus on addressing the limitations observed in the current approach. Several avenues for improvement and exploration include:

\begin{itemize}
	\item \textbf{Length extrapolation:} Keeping the model exact on numbers longer than those seen in training (Figures~\ref{fig:lengths} and~\ref{fig:lengthdim}). The digits of such numbers are decoded correctly once the length is known, so the length head is the part to improve.

	\item \textbf{Less supervised operators:} Learning the multiplication operator without hint supervision and with fewer structural priors, for example with operators whose Lipschitz constants are controlled (Lemma~\ref{lem:ac_lipschitz}).

	\item \textbf{Extended algebraic operations:} Expanding the scope of the evaluation to include additional algebraic operations, such as division and exponentiation, to assess the model's versatility and robustness in preserving algebraic properties across a broader range of operations.

	\item \textbf{Order of close numbers:} Training the comparator on more pairs of numbers with the same sign and number of digits, the hardest case of the order of fractions.

	\item \textbf{Further downstream evaluation:} Testing the embeddings in other downstream settings beyond the arithmetic benchmarks of Section~\ref{subsec:downstream}, for example numerical prediction on scientific data and tables, and on the output side of language models. Such tests are needed before any claim can be made about the use of the approach in specific application domains.
\end{itemize}

\FloatBarrier

\section*{CRediT authorship contribution statement}
\textbf{Hamidreza Sadeghi:} Conceptualization, Methodology, Software, Validation, Formal analysis, Investigation, Visualization, Writing -- original draft. \textbf{Saeedeh Momtazi:} Conceptualization, Methodology, Supervision, Writing -- review \& editing. \textbf{Reza Safabakhsh:} Supervision, Writing -- review \& editing.

\section*{Data and code availability}
The code of NIF and of all experiments (data generation, training, evaluation), the trained checkpoints, all results, the training logs and the random seeds are available at \url{https://github.com/HRSadeghi/Neural-Isomorphic-Fields}. The data of Sections~\ref{subsec:distribution_and_numerical_dataset}--\ref{subsec:results} are synthetic and can be regenerated exactly with the released scripts.

\section*{Declaration of competing interest}
The authors declare that they have no known competing financial interests or personal relationships that could have appeared to influence the work reported in this paper.

\section*{Funding}
This research did not receive any specific grant from funding agencies in the public, commercial, or not-for-profit sectors.

\section*{Declaration of generative AI and AI-assisted technologies in the manuscript preparation process}
During the preparation of this work, the authors used Gemini and ChatGPT in order to improve the language of the text. After using these tools, the authors reviewed and edited the content as needed and take full responsibility for the content of the published article.

\bibliographystyle{cas-model2-names}
\bibliography{references}

\appendix
\counterwithin{table}{section}\counterwithin{figure}{section}\counterwithin{equation}{section}%
\renewcommand{\thetheorem}{\Alph{section}.\arabic{theorem}}

\section{Implementation details}\label{app:implementation}

Table~\ref{tab:hparams} lists the configuration of NIF; the code, the configuration files and the logs of all runs are released with the paper.

\paragraph{Hint supervision} After each of the four layers of $\dot{\times}$, a linear head reads the slot states and is trained with cross-entropy on the quantities of Lemma~\ref{lem:small_sum}: the operand digits and signs and the three digits of $L_k$ and $H_k$ (first layer), $S_k\bmod 10$ and $\lfloor S_k/10\rfloor$ (second layer), $T_k$ (third layer), and the digits of the product (third and fourth layers). The map from the counts to $(L_k,H_k)$ is fitted separately (next paragraph). The heads are not used at test time.

\paragraph{Least-squares fit of the counts} The map from the pair counts to $(L_k,H_k)$ is linear in its parameters, and its exact solution has integer weights up to $9$. Gradient steps on it are slow and disturbed by the other terms, so the optimizer does not update it. Instead, it is set at fixed intervals to the closed-form least-squares solution on the current batch, which is exact once the digit classifier is correct (the residual is logged).

\paragraph{Further details} In the encoder and the decoder, the cross-attention between the tokens and the slots adds a learned scalar (initialized at 4) to the logits of the tokens whose exponent equals that of the slot. The pair counts $n_k(p,q)$ cover all 100 digit pairs, including the digit 0 of empty slots; they are computed by a linear convolution along the slots (an FFT of length $2K$) followed by the shift that maps $i+j$ to the exponent of the output slot. The first layer of $\dot{\times}$ concatenates the decimal Fourier features of $(L_k,H_k)$, $(L_k,H_k)/100$, the $\tanh$ of the two sign logits and the soft digit probabilities of both slots, projects them linearly and adds them to the slot tokens of $h_1$. The first and the last layer change the number of channels and have no residual connection. The hint heads read the slot tokens of every layer for both orders of the operands, and a further hint head reads the sign of the product from the mean of the slot tokens of the last layer; like the other hint heads, it is used only in training.

\begin{table}[t]
	\centering
	
	\caption{Implementation details and hyperparameters of NIF.}
	\label{tab:hparams}
	\renewcommand{\arraystretch}{1.12}
	\footnotesize
		\begin{tabular}{@{}l >{\raggedright\arraybackslash}p{0.8\textwidth}@{}}
			\toprule
			\textbf{Item} & \textbf{Value} \\
			\midrule
			Vocabulary & the 14 tokens of Section~\ref{subsec:number_representation} (digits 0--9, decimal point, $+$, $-$, start) and a padding token (the length head replaces an end token) \\
			Encoder & pre-norm Transformer (RMSNorm, SwiGLU) with exponent (place-value) embeddings of the tokens, 4 layers, width 256, 8 heads; attention pooling into $K=52$ slots ($e_{\min}=-26$) with exponent embeddings, 2 slot layers, per-slot RMS norm \\
			Embedding & $\varphi(a)\in\sR^{d}$, $d=Kc=520$ ($K=52$ place-value slots, $c=10$ channels per slot) \\
			Decoder & 2 layers over the slots, length head (two classifiers: $1$--$26$ integer and $0$--$26$ fractional digits, as many as the slots hold, so that results of operations fit), 4 Transformer layers (rotary positions) with cross-attention to the slots, width 256, 8 heads; greedy decoding of the predicted skeleton \\
			Operator $\dot{+}$ & $k=4$ AC layers; $f$: slot Transformer (1 block, width 192, 6 heads, periodic units in a quarter of the SwiGLU channels) \\
			Operator $\dot{\times}$ & $k=4$ AC layers aligned with Lemma~\ref{lem:small_sum} (2 blocks each, width 192, 6 heads, $c_{\mathrm{int}}=32$, float32); front end of Section~\ref{sec:staged}; hint heads used only during training \\
			Comparator $\omega$ & Transformer over the slots of both inputs (3 layers, width 192, 6 heads), antisymmetrized by Eq.~(\ref{eq:symmetrization}) \\
			Parameters & encoder 5.01M, decoder 6.08M, $\dot{+}$ 2.12M, $\dot{\times}$ 4.23M, comparator 1.34M, hint heads 0.02M (18.8M in total) \\
			Losses and weights & $\ell_{\mathrm{rec}}$ (tokens and length, 1), $\ell_{\mathrm{iso}}$ (MSE, weight 3 for $\dot{+}$ and 1 for $\dot{\times}$), $\ell^{\mathrm{CE}}_{\mathrm{ord}}$ (weight 0.5), decoder CE on operator outputs (1), compositions and both sides of the distributive law (0.5 each), order of outputs of $\dot{+}$ (0.5), hints (1) \\
			Training numbers & half from the sampler of Section~\ref{subsec:distribution_and_numerical_dataset} (negative binomial, $r_{\mathrm{NB}}=2$, $p_{\mathrm{NB}}=0.45$, $N=25$), half with integer and fractional lengths uniform in $\{0,\dots,12\}$; pairs restricted to $\Aadd$, $\Amul$; generated on the fly \\
			Targeted pairs & $10\%$ exact cancellations $a+(-a)$, $10\%$ near cancellations; 8 products $x\cdot y=10^k$ with $x=\pm2^i5^j$, $i,j,k\in\sZ$, $k\neq0$, per step (for the 14\% of these pairs whose digit products reach below the lowest slot, the target is the product truncated to the slot window) \\
			Optimizer & AdamW ($\beta=(0.9,0.98)$, weight decay 0.01), peak learning rate $5\cdot10^{-4}$, 500 warm-up steps, cosine decay to $5\%$, gradient-norm clipping at 1 \\
			Steps and batch & until convergence (validation every 500 steps; stop after 4 evaluations without an improvement $>0.2$, at least 3000, at most 12{,}000 steps); per step two micro-batches, each with 64 numbers or pairs per term and 32 tuples per composition term \\
			Curriculum & targeted pairs, distributive law and order of outputs of $\dot{+}$ from step 1000, compositions from step 2000 \\
			Least-squares fit & map from the pair counts to $(L_k,H_k)$ refitted in closed form every 100 steps up to step 1000, then every 500 steps \\
			Model selection & checkpoint with the best mean of the validation metrics (512 tuples per metric, every 500 steps; the $x\cdot x^{-1}$ test is excluded) \\
			Test sets & $10^4$ tuples per test, drawn with fixed seeds independently of the training data \\
			Seeds & 3 (0, 1, 2) \\
			Inference cost & per number with 25 digits, batch 512, on a partition of an NVIDIA H200 GPU: encoding 0.10\,ms, $\dot{+}$ 0.08\,ms, $\dot{\times}$ 0.41\,ms, comparator 0.02\,ms, length-free decoding 0.46\,ms; at most 1.7\,GB of GPU memory per batch \\
			Hardware & one GPU per run (NVIDIA L40S with 48\,GB for the full model, NVIDIA A100 with 80\,GB for the models with $N=200$, NVIDIA L4 with 24\,GB for the operators trained alone); about 2.2 hours per run of the main configuration \\
			\bottomrule
		\end{tabular}
\end{table}

\paragraph{Operator layers of the language model (Section~\ref{subsec:downstream})} The workspace holds the first 12 numbers of a prompt, and up to 32 numbers of the answer get registers. In each operator layer, the numbers are processed from left to right: for the $j$-th number, $u$ is an earlier number or a result computed for an earlier number, and $y$ is the number itself or a result that it produced in an earlier operator layer. Each pointer is a softmax over the keys of the allowed registers and a learned entry for ``nothing''; the key of a result is computed from the keys of its operands and its operation, and a three-way head chooses $u\,\dot{+}\,y$, $u\,\dot{+}\,(-y)$ or $(-u)\,\dot{+}\,y$. At test time, every choice is the arg max. In training, all choices whose result is a number of the reference text that no earlier register holds are correct (for the numbers of the answer: one of the next three numbers), the loss is the negative log of their total probability, and the first of them, in the order of the registers and operations, is kept as the teacher choice. A number of the answer is encoded by $\varphi$ once it is written, and its registers can be used from the second token after it. The output mixture can copy the next character of every register whose decoded string continues the number being written, and a learned weight changes the score of registers that were already written. NIF encodes the numbers of the prompt and of the answer, applies its operators to registers and decodes the registers; no gradient reaches it.

Tables~\ref{tab:abl_operator} and~\ref{tab:abl_full} list the numbers behind Figures~\ref{fig:operators} and~\ref{fig:profile}, Table~\ref{tab:lengthdim} those behind Figures~\ref{fig:lengths} and~\ref{fig:lengthdim}, and Table~\ref{tab:geometry} those behind Figure~\ref{fig:geometry}.

\begin{table}[htbp]
	\centering
	
	\caption{Operator ablations. Each operator is trained in isolation until convergence (Section~\ref{subsec:setup}; at most 8000 steps) on fixed one-hot digit slots (no encoder or decoder), and a linear head reads the digits of the result. EM: exact match (\%) on 4096 held-out pairs of the sampler of Section~\ref{subsec:distribution_and_numerical_dataset}, overall and for results with 21--25 digits, at the best evaluation; steps: training steps until convergence; first $\ge99\%$: first evaluation (every 250 steps) with $\text{EM}\ge 99\%$ (--: never). Stage accuracies of $\dot{\times}$ (\%, read by the hint heads): operand digits, $(L_k,H_k)$ within $0.5$, and $T_k$ (--: operator without these stages). The full model uses $k=4$ layers for $\dot{+}$ (Table~\ref{tab:hparams}).}
	\label{tab:abl_operator}
	\renewcommand{\arraystretch}{1.08}
	\footnotesize
	\resizebox{\textwidth}{!}{%
	\begin{tabular}{@{}lccccccc@{}}
		\toprule
		\textbf{Operator} & \textbf{EM} & \textbf{EM, 21--25 digits} & \textbf{Steps} & \textbf{First $\ge99\%$} & \textbf{Digits} & $(L_k,H_k)$ & $T_k$ \\
		\midrule
				$\dot{\times}$: staged operator (Section~\ref{sec:staged}) & 99.95 & 100.00 & 3250 & 2500 & 100.00 & 100.00 & 99.96 \\
		\quad without hint supervision & 29.39 & 0.00 & 8000 & -- & 0.00 & 1.69 & 8.54 \\
		\quad without least-squares fit of the counts & 99.98 & 98.18 & 5750 & 5250 & 100.00 & 100.00 & 100.00 \\
		\quad random instead of decimal frequencies & 99.98 & 100.00 & 5500 & 3500 & 100.00 & 100.00 & 100.00 \\
		\quad without periodic features & 99.88 & 98.18 & 8000 & 6250 & 100.00 & 100.00 & 99.86 \\
		\quad soft instead of one-hot digit counts & 99.98 & 100.00 & 3000 & 2250 & 100.00 & 100.00 & 99.98 \\
		\quad bfloat16 instead of float32 & 99.95 & 98.18 & 4000 & 2000 & 100.00 & 100.00 & 100.00 \\
		\midrule
		$\dot{\times}$: generic AC layers with a place-coupled bilinear branch, $k=6$ & 32.49 & 0.00 & 8000 & -- & -- & -- & -- \\
		$\dot{\times}$: perceptron AC layers, $f(x)=\sigma(Wx+b)$, $k=4$ & 10.52 & 0.00 & 8000 & -- & -- & -- & -- \\
		\midrule
		$\dot{+}$: slot-Transformer AC layers (Section~\ref{sec:ano}), $k=6$ & 100.00 & 100.00 & 3250 & 1750 & -- & -- & -- \\
		$\dot{+}$: perceptron AC layers, $f(x)=\sigma(Wx+b)$, $k=4$ & 20.41 & 0.00 & 8000 & -- & -- & -- & -- \\
		\midrule
		$\dot{\times}$: perceptron AC layers, $k=1$ & 9.69 & 0.00 & 1750 & -- & -- & -- & -- \\
		$\dot{\times}$: perceptron AC layers, $k=2$ & 10.11 & 0.00 & 6500 & -- & -- & -- & -- \\
		$\dot{\times}$: perceptron AC layers, $k=6$ & 10.13 & 0.00 & 4750 & -- & -- & -- & -- \\
		$\dot{\times}$: generic AC layers with the bilinear branch, $k=2$ & 29.05 & 0.00 & 8000 & -- & -- & -- & -- \\
		$\dot{\times}$: generic AC layers with the bilinear branch, $k=4$ & 37.23 & 0.00 & 8000 & -- & -- & -- & -- \\
		$\dot{\times}$: generic AC layers with the bilinear branch, $k=8$ & 43.14 & 0.00 & 8000 & -- & -- & -- & -- \\
		\bottomrule
	\end{tabular}%
	}
\end{table}

\begin{table}[htbp]
	\centering
	
	\caption{Ablations of the full model: exact match (\%, length-free decoding) on the test sets. Each variant changes one component of NIF and is trained with the same budget and seed 0; the first row is the mean over the seeds of Table~\ref{tab:main}, the second the seed-0 run. $^{*}$ (all columns but the last): different from the seed-0 run (McNemar's test, Holm--Bonferroni-corrected $p<0.05$ within the table). Rec.: reconstruction; Cl.: closure; Inv.: inverse; Ass.: associativity; Distr.: distributivity; O1: compatibility of the order with $\dot{+}$; last column: chains of 5 sums (500 chains). The other tests of Figure~\ref{fig:profile} are released with the code.}
	\label{tab:abl_full}
	\renewcommand{\arraystretch}{1.08}
	\footnotesize
	\resizebox{\textwidth}{!}{%
	\begin{tabular}{@{}lcccccccccc@{}}
		\toprule
		\textbf{Variant} & Rec. & Cl.\,$\dot{+}$ & Cl.\,$\dot{\times}$ & Inv.\,$\dot{+}$ & Inv.\,$\dot{\times}$ & Ass.\,$\dot{+}$ & Ass.\,$\dot{\times}$ & Distr. & O1 & $\dot{+}$, $M{=}5$ \\
		\midrule
				NIF (all terms) & 100.00 & 100.00 & 99.98 & 100.00 & 85.22 & 99.98 & 99.94 & 99.92 & 99.44 & 99.87 \\
		NIF, seed 0 (reference of the tests) & 99.99 & 99.99 & 99.97 & 99.99 & 84.86 & 99.99 & 99.95 & 99.88 & 99.57 & 100.00 \\
		\midrule
		only $\ell_{\mathrm{rec}}$, $\ell_{\mathrm{iso}}$, $\ell_{\mathrm{ord}}$ & 99.92 & 99.78$^{*}$ & 22.47$^{*}$ & 88.47$^{*}$ & 0.72$^{*}$ & 99.61$^{*}$ & 5.12$^{*}$ & 11.61$^{*}$ & 92.71$^{*}$ & 100.00 \\
		without hint supervision & 99.92 & 99.80$^{*}$ & 10.00$^{*}$ & 100.00 & 17.44$^{*}$ & 99.62$^{*}$ & 3.78$^{*}$ & 2.37$^{*}$ & 99.74 & 99.80 \\
		without targeted pairs and terms on operator outputs & 100.00 & 100.00 & 99.85 & 99.96 & 27.17$^{*}$ & 99.88 & 99.70$^{*}$ & 99.55$^{*}$ & 86.43$^{*}$ & 100.00 \\
		training numbers only from the sampler (no uniform lengths) & 99.99 & 100.00 & 100.00 & 100.00 & 85.40$^{*}$ & 99.99 & 99.98 & 99.97 & 99.76 & 100.00 \\
		perceptron AC layers in both operators & 99.71$^{*}$ & 4.40$^{*}$ & 3.04$^{*}$ & 100.00 & 1.80$^{*}$ & 0.20$^{*}$ & 3.34$^{*}$ & 1.22$^{*}$ & 99.23$^{*}$ & 0.00 \\
		AE-only encoder (trained with $\ell_{\mathrm{rec}}$ only, then frozen) & 99.98 & 99.93 & 29.02$^{*}$ & 100.00 & 31.14$^{*}$ & 99.56$^{*}$ & 5.96$^{*}$ & 12.08$^{*}$ & 98.84$^{*}$ & 99.60 \\
		embedding size $c=2$ ($d=104$) & 99.99 & 95.22$^{*}$ & 99.01$^{*}$ & 100.00 & 83.82$^{*}$ & 15.90$^{*}$ & 3.45$^{*}$ & 10.80$^{*}$ & 99.56 & 1.60 \\
		embedding size $c=4$ ($d=208$) & 100.00 & 100.00 & 100.00 & 100.00 & 85.40$^{*}$ & 99.73$^{*}$ & 100.00 & 99.80 & 99.68 & 96.60 \\
		embedding size $c=6$ ($d=312$) & 99.85$^{*}$ & 99.74$^{*}$ & 99.98 & 100.00 & 85.40$^{*}$ & 98.76$^{*}$ & 99.74$^{*}$ & 98.30$^{*}$ & 99.30 & 97.40 \\
		embedding size $c=16$ ($d=832$) & 100.00 & 100.00 & 100.00 & 100.00 & 85.40$^{*}$ & 99.98 & 100.00 & 99.99 & 99.73 & 100.00 \\
		\bottomrule
	\end{tabular}%
	}
\end{table}

\begin{table}[htbp]
	\centering
	
	\caption{Number length and embedding dimension: exact match (\%, length-free decoding) by the shape of the number that the decoder has to produce (the number itself for reconstruction, the result for the closure tests; items of Figures~\ref{fig:lengths} and~\ref{fig:lengthdim}). ``Within'': integer and fractional part within the training lengths (at most 12 digits each for $N=25$ and $\lfloor N/2\rfloor$ otherwise) and at most $N$ digits in total; ``longer part'': a longer part, at most $N$ digits; ``more than $N$ digits'': outside $\mathbb{D}_N$. $N=25$, $c=10$: mean of the three seeds; all other rows: one run.}
	\label{tab:lengthdim}
	\renewcommand{\arraystretch}{1.08}
	\footnotesize
	\begin{tabular}{@{}rrr@{\hskip 10pt}ccc@{\hskip 10pt}ccc@{\hskip 10pt}ccc@{}}
		\toprule
		 & & & \multicolumn{3}{c}{\textbf{Within the training lengths}} & \multicolumn{3}{c}{\textbf{Longer part}} & \multicolumn{3}{c}{\textbf{More than $N$ digits}} \\
		\cmidrule(lr){4-6} \cmidrule(lr){7-9} \cmidrule(l){10-12}
		$N$ & $c$ & $d$ & Rec. & $\dot{+}$ & $\dot{\times}$ & Rec. & $\dot{+}$ & $\dot{\times}$ & Rec. & $\dot{+}$ & $\dot{\times}$ \\
		\midrule
		$25$ & $2$ & 104 & 100 & 91.4 & 98.7 & 93.2 & 73.0 & 78.9 & 34.0 & 25.9 & 23.7 \\
		 & $4$ & 208 & 100 & 100 & 100 & 98.5 & 98.5 & 99.93 & 64.5 & 72.4 & 78.6 \\
		 & $6$ & 312 & 99.94 & 99.69 & 99.94 & 98.3 & 98.3 & 98.1 & 51.9 & 59.5 & 64.0 \\
		 & $10$ & 520 & 100 & 100 & 99.99 & 98.1 & 98.6 & 99.50 & 53.5 & 60.5 & 63.8 \\
		 & $16$ & 832 & 100 & 100 & 100 & 99.26 & 99.30 & 99.86 & 72.0 & 75.5 & 79.8 \\
		\multicolumn{3}{@{}l}{\quad $c=10$, no uniform lengths} & 100 & 100 & 100 & 97.1 & 98.9 & 99.36 & 43.3 & 50.8 & 55.9 \\
		\multicolumn{3}{@{}l}{\quad $c=10$, no targeted pairs} & 100 & 100 & 99.94 & 92.1 & 94.6 & 98.1 & 49.3 & 54.1 & 65.3 \\
		\midrule
		$50$ & $2$ & 204 & 100 & 85.9 & 26.5 & 22.7 & 14.5 & 1.8 & 3.1 & 1.0 & 0.1 \\
		 & $4$ & 408 & 100 & 100 & 100 & 91.8 & 93.3 & 99.12 & 52.3 & 57.5 & 83.2 \\
		 & $6$ & 612 & 100 & 100 & 100 & 81.9 & 87.5 & 84.4 & 41.1 & 47.3 & 66.1 \\
		 & $10$ & 1020 & 100 & 100 & 99.98 & 69.8 & 77.3 & 77.9 & 27.0 & 32.2 & 54.1 \\
		\midrule
		$100$ & $2$ & 404 & 38.8 & 4.5 & 3.1 & 0.2 & 0.0 & 0.0 & 0.0 & 0.0 & 0.0 \\
		 & $4$ & 808 & 100 & 99.73 & 100 & 62.4 & 73.3 & 98.4 & 24.4 & 25.9 & 66.6 \\
		 & $6$ & 1212 & 100 & 99.94 & 99.91 & 43.4 & 58.2 & 98.2 & 8.0 & 13.4 & 36.8 \\
		 & $10$ & 2020 & 100 & 99.94 & 100 & 28.1 & 42.5 & 91.7 & 3.0 & 6.2 & 69.1 \\
		\midrule
		$200$ & $2$ & 804 & 65.1 & 8.5 & 2.9 & 0.0 & 0.0 & 0.0 & 0.0 & 0.0 & 0.0 \\
		 & $4$ & 1608 & 100 & 99.94 & 99.84 & 5.2 & 11.5 & 83.5 & 0.8 & 1.0 & 25.9 \\
		 & $6$ & 2412 & 100 & 99.64 & 98.3 & 12.9 & 44.6 & 78.2 & 3.1 & 14.2 & 34.4 \\
		 & $10$ & 4020 & 100 & 99.86 & 99.54 & 14.6 & 22.2 & 82.0 & 2.2 & 2.2 & 39.9 \\
		\bottomrule
	\end{tabular}
\end{table}

\begin{table}[htbp]
	\centering
	
	\caption{Quantities of Section~\ref{sec:approximate} on 2000 test tuples per test (length-free decoding, mean over seeds; EM: exact match on these tuples, a subset of the test sets of Table~\ref{tab:main}). Defect: $\epsadd$, $\epsmul$ (for the laws: unweighted sum of the defects in the bound of Proposition~\ref{prop:assoc_distrib}), median / 90th percentile; $\hat\rho$: median estimated decoding radius (Lemma~\ref{lem:margin}, 50 random perturbations of norm $\delta_0=6$, with the length head included in the margin); in bound: tuples for which Proposition~\ref{prop:decoding_error}(a), respectively the bounds of Proposition~\ref{prop:assoc_distrib} weighted with $\hat L$, give the correct result for the estimated radius; wrong ball: the result lies in the decoding ball of a wrong number, Proposition~\ref{prop:decoding_error}(b); $\hat L$: empirical Lipschitz constant (the rows of the laws use $\hat L_+$ and $\hat L_\times$ of the first two rows). The radii and these shares are estimates, not guarantees. Over the results of the closure tests, $B=\max\norm{\varphi(a)}=22.81$, the smallest estimated radius is $\hat\rho_{\min}=1.26$ ($\dot{\times}$) and $1.50$ ($\dot{+}$), and Eq.~(\ref{eq:capacity}) gives $N\le670$ for $d=520$. Rows $c=2,\dots,16$: the models with $c$ channels per slot of Figure~\ref{fig:profile} (single runs); each cell gives the values for $\dot{+}$ / $\dot{\times}$, the defect column the medians.}
	\label{tab:geometry}
	\renewcommand{\arraystretch}{1.08}
	\footnotesize
	\begin{tabular}{@{}lcccccc@{}}
		\toprule
		\textbf{Test} & \textbf{EM (\%)} & \textbf{Defect} & $\hat\rho$ & \textbf{In bound (\%)} & \textbf{Wrong ball (\%)} & $\hat L$ \\
		\midrule
				$\dot{+}$, closure & 100.00 & 0.73 / 1.10 & 2.88 & 99.8 & 0.0 & 2.52 \\
		$\dot{\times}$, closure & 99.98 & 0.83 / 1.08 & 2.80 & 99.9 & 0.0 & 3.94 \\
		Associativity of $\dot{+}$ & 100.00 & 1.49 / 2.07 & 2.80 & 42.9 & -- & -- \\
		Distributivity & 99.92 & 1.66 / 2.07 & 2.67 & 1.1 & -- & -- \\
		\midrule
		$c=2$: closure of $\dot{+}$ / $\dot{\times}$ & 94.75 / 99.20 & 1.35 / 2.18 & 1.33 / 1.34 & 47.8 / 12.4 & 0.7 / 0.0 & 2.71 / 10.49 \\
		$c=4$: closure of $\dot{+}$ / $\dot{\times}$ & 100.00 / 100.00 & 0.48 / 0.34 & 1.52 / 1.51 & 99.2 / 100.0 & 0.0 / 0.0 & 2.22 / 4.12 \\
		$c=6$: closure of $\dot{+}$ / $\dot{\times}$ & 99.65 / 100.00 & 1.52 / 1.78 & 2.16 / 2.05 & 87.5 / 72.0 & 0.2 / 0.0 & 2.24 / 3.80 \\
		$c=16$: closure of $\dot{+}$ / $\dot{\times}$ & 100.00 / 100.00 & 0.88 / 1.29 & 3.35 / 3.18 & 99.2 / 99.5 & 0.0 / 0.0 & 3.96 / 4.03 \\
		\bottomrule
	\end{tabular}
\end{table}

\section{Proofs}\label{appendix:theorems}

Section~\ref{app:auxiliary} collects auxiliary results. Results of the main text are not restated; their proofs follow in Sections~\ref{app:proofs_ideal}--\ref{app:proof_small_sum}, in the order in which the results appear. 

\subsection{Auxiliary results}\label{app:auxiliary}

\begin{lemma}[{Homomorphic image of a group; see, e.g., \citealp{herstein1996abstract}}]\label{con_hom_group}
	Let $(G_1, *_1, e)$ be a group (respectively, an abelian group), let $G_2$ be a set, let $\varphi\colon G_1 \rightarrow G_2$ be a map and let $*_2\colon G_2 \times G_2 \rightarrow G_2$ be a binary operation such that, for all $a, b \in G_1$,
	\begin{align}
		\varphi(a *_1 b) = \varphi(a) *_2 \varphi(b).
	\end{align}
	Then $\varphi(G_1)$ is closed under $*_2$, and $(\varphi(G_1), *_2, \varphi(e))$, with $*_2$ restricted to $\varphi(G_1)$, is a group (respectively, an abelian group).
\end{lemma}

\begin{proof}
	Every element of $\varphi(G_1)$ is of the form $\varphi(a)$ with $a\in G_1$. Closure: $\varphi(a)*_2\varphi(b)=\varphi(a*_1b)\in\varphi(G_1)$. Associativity: $\varphi(a)*_2(\varphi(b)*_2\varphi(c))=\varphi(a*_1(b*_1c))=\varphi((a*_1b)*_1c)=(\varphi(a)*_2\varphi(b))*_2\varphi(c)$. Identity and inverses: $\varphi(a)*_2\varphi(e)=\varphi(a*_1e)=\varphi(a)$ and $\varphi(a)*_2\varphi(a^{-1})=\varphi(a*_1a^{-1})=\varphi(e)$, and likewise with the factors exchanged. If $G_1$ is abelian, then $\varphi(a)*_2\varphi(b)=\varphi(a*_1b)=\varphi(b*_1a)=\varphi(b)*_2\varphi(a)$.
\end{proof}

\begin{proposition}[Homomorphic images of a field]\label{lem:field_image}
	Let $(F,+_1,\times_1)$ be a field, let $X$ be a set, let $\varphi:F\rightarrow X$ be a map and let $+_2,\times_2:X\times X\rightarrow X$ be binary operations such that $\varphi(a+_1b)=\varphi(a)+_2\varphi(b)$ and $\varphi(a\times_1b)=\varphi(a)\times_2\varphi(b)$ for all $a,b\in F$. Then $\varphi(F)$ is closed under $+_2$ and $\times_2$, and the following statements are equivalent: (i) $(\varphi(F),+_2,\times_2)$ is a field; (ii) $\varphi(0)\neq\varphi(1)$; (iii) $\varphi$ is injective.
\end{proposition}
\begin{proof}
	Closure follows as in the proof of Lemma~\ref{con_hom_group}. Let $J:=\{a\in F:\varphi(a)=\varphi(0)\}$. For $a,b\in J$ and $x\in F$, we have $\varphi(a+_1b)=\varphi(0)+_2\varphi(0)=\varphi(0+_10)=\varphi(0)$ and $\varphi(x\times_1a)=\varphi(x)\times_2\varphi(0)=\varphi(x\times_10)=\varphi(0)$; in particular $-a=(-1)\times_1a\in J$. Hence $J$ is an ideal of $F$. A field has only the ideals $\{0\}$ and $F$.

	(i)$\Rightarrow$(ii): if $\varphi(0)=\varphi(1)$, then $1\in J$, so $J=F$ and $\varphi(F)=\{\varphi(0)\}$ has a single element; since a field has at least the two distinct elements $0$ and $1$, $\varphi(F)$ is not a field.

	(ii)$\Rightarrow$(iii): if $\varphi(0)\neq\varphi(1)$, then $1\notin J$, so $J=\{0\}$. If $\varphi(a)=\varphi(b)$, then $\varphi(a-b)=\varphi(a)+_2\varphi(-b)=\varphi(b)+_2\varphi(-b)=\varphi(0)$, i.e., $a-b\in J$, hence $a=b$.

	(iii)$\Rightarrow$(i): if $\varphi$ is injective, it is a bijection from $F$ onto $\varphi(F)$ that preserves both operations, and every field axiom transfers to $\varphi(F)$ exactly as in Steps~2--3 of the proof of Proposition~\ref{prop:transport} below.
\end{proof}
\noindent Proposition~\ref{lem:field_image} shows why injectivity, which Theorem~\ref{theorem:total_conds} obtains from $\ell_{\mathrm{rec}}$ and $\ell_{\mathrm{ord}}$, cannot be dropped: a map that preserves both operations of a field may collapse it to a single point (for example, a constant map), and its image is then not a field.

\begin{lemma}\label{hom_order_set}
	Let $<_1$ be an order on a set $A_1$, let $\varphi:A_1 \rightarrow A_2$ be a map, and let $<_2\ \subseteq A_2\times A_2$ be a relation such that, for all $a,b\in A_1$,
	\begin{align}
		a <_1 b \Longleftrightarrow \varphi(a) <_2 \varphi(b).
	\end{align}
	Then $\varphi$ is injective, and the restriction of $<_2$ to $\varphi(A_1)$ is an order on $\varphi(A_1)$.
\end{lemma}

\begin{proof}

	\emph{Injectivity.} Let $a,b\in A_1$ with $a\neq b$, and suppose that $\varphi(a)=\varphi(b)$. By trichotomy in $A_1$, $a<_1b$ or $b<_1a$; without loss of generality $a<_1b$. Then $\varphi(a)<_2\varphi(b)=\varphi(a)$, i.e., $\varphi(a)<_2\varphi(a)$, and the assumed equivalence with $b$ replaced by $a$ gives $a<_1a$. This contradicts trichotomy in $A_1$, since $a=a$. Hence $\varphi$ is injective.

	\emph{Trichotomy.} Let $h_1,h_2\in\varphi(A_1)$ and choose $a,b\in A_1$ with $h_1=\varphi(a)$ and $h_2=\varphi(b)$. Exactly one of $a<_1b$, $a=b$, $b<_1a$ holds. By the assumed equivalence and by injectivity, these three statements are equivalent to $h_1<_2h_2$, $h_1=h_2$ and $h_2<_2h_1$, respectively; hence exactly one of the latter holds.

	\emph{Transitivity.} If $\varphi(a)<_2\varphi(b)$ and $\varphi(b)<_2\varphi(c)$, then $a<_1b$ and $b<_1c$, hence $a<_1c$ and therefore $\varphi(a)<_2\varphi(c)$. \qedhere
	
\end{proof}

\subsection{Proofs of the results of Sections~\ref{subsec:number_representation}--\ref{sec:papdt}}\label{app:proofs_ideal}

\begin{proof}[Proof of Lemma~\ref{lem:domain}]
	(a) We have $0,1\in\Dset$ and $\sZ\subseteq\Dset$ (take $\nu=0$ in Eq.~(\ref{eq:decimal_fractions})). For $a=m10^{-\nu}$ and $b=m'10^{-\nu'}$ in $\Dset$, $a-b=(m10^{\nu'}-m'10^{\nu})10^{-(\nu+\nu')}\in\Dset$ and $ab=mm'10^{-(\nu+\nu')}\in\Dset$. Hence $\Dset$ is a subring of $\sQ$. As a subring of a field containing $1$, it is commutative, satisfies $1\neq0$ and has no zero divisors, i.e., it is an integral domain. The restriction of the order of $\sQ$ to $\Dset$ is an order on $\Dset$, and (O1) and (O2) hold in $\Dset$ because they hold in the ordered field $\sQ$.

	(b) If $p/q=m10^{-\nu}$, then $p\,10^{\nu}=mq$. Since $\gcd(p,q)=1$, $q$ divides $10^{\nu}=2^{\nu}5^{\nu}$, hence $q=2^i5^j$ with $0\le i,j\le\nu$. Conversely, if $q=2^i5^j$ and $M:=\max\{i,j\}$, then $p/q=p\,2^{M-i}5^{M-j}10^{-M}\in\Dset$.

	(c) Let $a=\pm2^i5^j$ with $i,j\in\sZ$. In lowest terms, the numerators and the denominators of $a$ and of $1/a=\pm2^{-i}5^{-j}$ are products of powers of $2$ and $5$, so $a,1/a\in\Dset$ by (b), i.e., $a\in\Dset^{\times}$. Conversely, let $a\in\Dset^{\times}$, i.e., $a\neq0$ and $1/a\in\Dset$, and write $a=p/q$ in lowest terms with $q>0$. By (b), applied to $a$ and to $1/a=\operatorname{sgn}(p)\,q/|p|$, which is also in lowest terms, both $q$ and $|p|$ are products of powers of $2$ and $5$; hence $a=\pm2^i5^j$ for some $i,j\in\sZ$. Finally, $1/3\notin\Dset$ by (b), so $3\notin\Dset^{\times}$ and $\Dset$ is not a field.

	(d) The inclusion $\iota:\Dset\rightarrow\sQ$ is an injective ring homomorphism into a field. By the universal property of the field of fractions (field of quotients) of an integral domain \citep{herstein1996abstract}, it extends uniquely to a field homomorphism $\bar\iota:\mathrm{Frac}(\Dset)\rightarrow\sQ$, $[a,b]\mapsto a/b$. Field homomorphisms are injective, and $\bar\iota$ is surjective because every rational number is a quotient $p/q$ of integers $p,q\in\sZ\subseteq\Dset$ with $q\neq0$. Hence $\mathrm{Frac}(\Dset)\cong\sQ$.
\end{proof}

\begin{proof}[Proof of Lemma~\ref{lem:commutativity}]
	Let $(t^i_1,t^i_2)$ be the sequence of Eqs.~(\ref{eq:ac_rec1})--(\ref{eq:ac_rec2}) for the input $(h_1,h_2)$, and let $(\tilde t^i_1,\tilde t^i_2)$ be the sequence for the input $(h_2,h_1)$. We show by induction that $\tilde t^i_1=t^i_2$ and $\tilde t^i_2=t^i_1$ for $0\le i\le k$. For $i=0$, $\tilde t^0_1=h_2=t^0_2$ and $\tilde t^0_2=h_1=t^0_1$. If the claim holds for $i-1$, then $\tilde t^i_1=f([\tilde t^{i-1}_1,\tilde t^{i-1}_2];\theta_i)=f([t^{i-1}_2,t^{i-1}_1];\theta_i)=t^i_2$, and similarly $\tilde t^i_2=t^i_1$. Hence $h_2*_\Theta h_1=\tilde t^k_1+\tilde t^k_2=t^k_2+t^k_1=h_1*_\Theta h_2$ by Eq.~(\ref{eq:neural_op}).
\end{proof}

\begin{proof}[Proof of Proposition~\ref{prop:antisymmetric}]
	Antisymmetry means $\omega_1(h_2,h_1)=\omega_2(h_1,h_2)$, $\omega_2(h_2,h_1)=\omega_1(h_1,h_2)$ and $\omega_3(h_2,h_1)=\omega_3(h_1,h_2)$.

	(a) For $h_1=h_2=h$, antisymmetry gives $\omega_1(h,h)=\omega_2(h,h)$, so the strict inequality $\omega_1(h,h)>\omega_2(h,h)$ in Eq.~(\ref{eq:induced_relation}) fails.

	(c) By Definition~\ref{def:induced_relation}, $h_2<_\omega h_1$ means $\omega_1(h_2,h_1)>\max\{\omega_2(h_2,h_1),\omega_3(h_2,h_1)\}$, which by antisymmetry is $\omega_2(h_1,h_2)>\max\{\omega_1(h_1,h_2),\omega_3(h_1,h_2)\}$.

	(b) If $h_1<_\omega h_2$, then $\omega_1(h_1,h_2)>\omega_2(h_1,h_2)$, so the condition in (c) fails and $h_2<_\omega h_1$ does not hold.

	For the final statement, let $h_1,h_2\in S$. If $h_1=h_2$, then by (a) neither $h_1<_\omega h_2$ nor $h_2<_\omega h_1$, so exactly one of the three statements in (N1) holds. If $h_1\neq h_2$, then by (b) at most one of $h_1<_\omega h_2$ and $h_2<_\omega h_1$ holds, and by connectedness at least one holds. Hence (N1) is equivalent to connectedness, and (N2) is transitivity.
\end{proof}

\begin{proof}[Proof of Proposition~\ref{prop:transport}]
	\emph{Step 1 (injectivity and order).} The order of an ordered integral domain is an order on the set $R$. By Eq.~(\ref{eq:order_field}) and Lemma~\ref{hom_order_set}, $\varphi$ is injective and $<_2$ is an order on $\varphi(R)$.

	\emph{Step 2 (closure).} Every $h\in\varphi(R)$ is of the form $h=\varphi(a)$ with a unique $a\in R$. For $h_1=\varphi(a_1)$ and $h_2=\varphi(a_2)$, Eqs.~(\ref{eq:addition_iso})--(\ref{eq:multiplication_iso}) give $h_1+_2h_2=\varphi(a_1+_1a_2)\in\varphi(R)$ and $h_1\times_2h_2=\varphi(a_1\times_1a_2)\in\varphi(R)$.

	\emph{Step 3 (ring and field axioms).} Let $h_i=\varphi(a_i)$, $i=1,2,3$. Commutativity and associativity of $+_2$ and $\times_2$ follow as in the proof of Lemma~\ref{con_hom_group}. For distributivity,
	\begin{align*}
		h_1\times_2(h_2+_2h_3)&=\varphi(a_1)\times_2\varphi(a_2+_1a_3)=\varphi(a_1\times_1(a_2+_1a_3))\\
		&=\varphi(a_1\times_1a_2+_1a_1\times_1a_3)=\varphi(a_1\times_1a_2)+_2\varphi(a_1\times_1a_3)=(h_1\times_2h_2)+_2(h_1\times_2h_3).
	\end{align*}
	The element $\varphi(0)$ is a zero, since $h_1+_2\varphi(0)=\varphi(a_1+_10)=h_1$, and $\varphi(-a_1)$ is an additive inverse of $h_1$. Similarly, $\varphi(1)$ is an identity, and $\varphi(1)\neq\varphi(0)$ because $1\neq0$ and $\varphi$ is injective. If $h_1\times_2h_2=\varphi(0)$, then $\varphi(a_1\times_1a_2)=\varphi(0)$, hence $a_1\times_1a_2=0$ by injectivity, so $a_1=0$ or $a_2=0$, i.e., $h_1=\varphi(0)$ or $h_2=\varphi(0)$; thus there are no zero divisors. If $R$ is a field and $h_1\neq\varphi(0)$, then $a_1\neq0$ and $h_1\times_2\varphi(a_1^{-1})=\varphi(1)$.

	\emph{Step 4 (compatibility of the order).} (O1): if $h_2<_2h_3$, then $a_2<_1a_3$ by Eq.~(\ref{eq:order_field}), hence $a_1+_1a_2<_1a_1+_1a_3$ and $h_1+_2h_2=\varphi(a_1+_1a_2)<_2\varphi(a_1+_1a_3)=h_1+_2h_3$. (O2): if $\varphi(0)<_2h_1$ and $\varphi(0)<_2h_2$, then $0<_1a_1$ and $0<_1a_2$, hence $0<_1a_1\times_1a_2$ and $\varphi(0)<_2h_1\times_2h_2$.

	\emph{Step 5 (isomorphism).} By Steps 1--4, $\varphi:R\rightarrow\varphi(R)$ is a bijection that preserves both operations and satisfies $a<_1b\Leftrightarrow\varphi(a)<_2\varphi(b)$, i.e., an isomorphism of ordered rings (of ordered fields if $R$ is a field).
\end{proof}

\begin{proof}[Proof of Lemma~\ref{lem:rec}]
	
	Since $\Dset$ is countable, $\ell_{\mathrm{rec}}=\sum_{a\in\Dset}\distP(a)\,\Delta_{\Dset}(\varphi^\dagger(\varphi(a)),a)$, a series of non-negative terms. If it vanishes, every term vanishes; since $\distP(a)>0$, $\Delta_{\Dset}(\varphi^\dagger(\varphi(a)),a)=0$ and hence $\varphi^\dagger(\varphi(a))=a$ for all $a\in\Dset$. If $\varphi(a_1)=\varphi(a_2)$, then $a_1=\varphi^\dagger(\varphi(a_1))=\varphi^\dagger(\varphi(a_2))=a_2$, so $\varphi$ is injective.
\end{proof}

\begin{proof}[Proof of Lemma~\ref{lem:iso}]
	As above, $\sum_{a,b\in\Dset}\distP(a)\distP(b)\,\Delta_E(\varphi(a*b),\varphi(a)\,\dot{*}\,\varphi(b))=0$ with $\distP(a)\distP(b)>0$ implies $\Delta_E(\varphi(a*b),\varphi(a)\,\dot{*}\,\varphi(b))=0$, hence $\varphi(a*b)=\varphi(a)\,\dot{*}\,\varphi(b)$, for all $a,b\in\Dset$.
\end{proof}

\begin{remark}\label{rem:groups_app}
	For $*=+$, Lemma~\ref{lem:iso} and Lemma~\ref{con_hom_group} show that $(\varphi(\Dset),\dot{+})$ is an abelian group with identity $\varphi(0)$. For $*=\times$, the same argument applies only to the group of units $\Dset^{\times}$ (Lemma~\ref{lem:domain}(c)); the set $\Dset\setminus\{0\}$ is a commutative monoid, but not a group, under multiplication. 
\end{remark}

\begin{proof}[Proof of Lemma~\ref{lem:ord}]
	For two statements $A$ and $B$, $[1(A)-1(B)]^2$ equals $1$ if exactly one of them holds and $0$ otherwise. As above, $\ell_{\mathrm{ord}}=0$ implies $1(a<b)=1(\varphi(a)<_\omega\varphi(b))$ for all $a,b\in\Dset$, i.e., $a<b\Leftrightarrow\varphi(a)<_\omega\varphi(b)$. By Lemma~\ref{hom_order_set}, $\varphi$ is injective and $<_\omega$ is an order on $\varphi(\Dset)$, i.e., a neural order on $\varphi(\Dset)$ in the sense of Definition~\ref{def:neural_order}.
\end{proof}

\begin{proof}[Proof of Theorem~\ref{theorem:total_conds}]
	The three losses are non-negative, so $\ell_{\mathrm{total}}=0$ (or a weighted sum with positive weights equal to $0$) implies $\ell_{\mathrm{rec}}=\ell_{\mathrm{iso}}=\ell_{\mathrm{ord}}=0$; moreover, both terms of $\ell_{\mathrm{iso}}$ vanish. By Lemmas~\ref{lem:rec}, \ref{lem:iso} and~\ref{lem:ord}, $\varphi^\dagger\circ\varphi=\mathrm{id}_{\Dset}$, Eqs.~(\ref{eq:addition_iso})--(\ref{eq:order_field}) hold with $R=\Dset$, $X=\sR^d$, $+_2=\dot{+}$, $\times_2=\dot{\times}$ and ${<_2}={<_\omega}$, and $<_\omega$ is a neural order on $H$. Since $(\Dset,+,\times,<)$ is an ordered integral domain (Lemma~\ref{lem:domain}(a)), Proposition~\ref{prop:transport} shows that $H$ is closed under $\dot{+}$ and $\dot{\times}$, that $(H,\dot{+},\dot{\times},<_\omega)$ is an ordered integral domain, and that $\varphi:\Dset\rightarrow H$ is an isomorphism of ordered rings. Hence $(H,\dot{+},\dot{\times},<_\omega)$ is a neural ordered integral domain (Definition~\ref{def:neural_ordered_domain}). Finally, for $h=\varphi(a)\in H$, $\varphi^\dagger(h)=a=\varphi^{-1}(h)$.
\end{proof}

\begin{proof}[Proof of Corollary~\ref{cor:fraction_field}]
	By Theorem~\ref{theorem:total_conds}, $\varphi:\Dset\rightarrow H$ is an isomorphism of ordered rings; hence so is its inverse $\varphi^{-1}:H\rightarrow\Dset$, and $\varphi^{-1}(h')\neq0$ for $h'\in H^*$. Define $\kappa:H\times H^*\rightarrow\sQ$ by $\kappa(h,h'):=\varphi^{-1}(h)/\varphi^{-1}(h')$.

	\emph{The relation $\sim$.} Since $\varphi^{-1}$ is injective and multiplicative, $(h_1,h_1')\sim(h_2,h_2')$ holds if and only if $\varphi^{-1}(h_1)\varphi^{-1}(h_2')=\varphi^{-1}(h_2)\varphi^{-1}(h_1')$, i.e., if and only if $\kappa(h_1,h_1')=\kappa(h_2,h_2')$. Hence $\sim$ is the equivalence relation ``same value of $\kappa$'', and $\kappa$ induces an injective map $\Psi:F_H\rightarrow\sQ$, $\Psi([h,h']):=\kappa(h,h')$. It is surjective: every $r\in\sQ$ can be written as $r=p/q$ with $p\in\sZ$ and $q\in\sZ_{>0}$, and $\Psi([\varphi(p),\varphi(q)])=r$.

	\emph{Positive denominators.} Let $h'\in H^*$. Since $<_\omega$ is an order on $H$, either $\varphi(0)<_\omega h'$ or $h'<_\omega\varphi(0)$. In the second case, (O1) with $x=-h'$ gives $\varphi(0)=-h'\dot{+}h'<_\omega-h'\dot{+}\varphi(0)=-h'$, and $(h,h')\sim(-h,-h')$ because $h\dot{\times}(-h')=-(h\dot{\times}h')=(-h)\dot{\times}h'$ in the ring $H$.

	\emph{Operations and order.} For $\xi_1=[h_1,h_1']$ and $\xi_2=[h_2,h_2']$, the multiplicativity and additivity of $\varphi^{-1}$ give
	\begin{align*}
		\kappa\big(h_1\dot{\times}h_2'\dot{+}h_2\dot{\times}h_1',\ h_1'\dot{\times}h_2'\big)&=\frac{\varphi^{-1}(h_1)\varphi^{-1}(h_2')+\varphi^{-1}(h_2)\varphi^{-1}(h_1')}{\varphi^{-1}(h_1')\varphi^{-1}(h_2')}=\Psi(\xi_1)+\Psi(\xi_2),\\
		\kappa\big(h_1\dot{\times}h_2,\ h_1'\dot{\times}h_2'\big)&=\Psi(\xi_1)\Psi(\xi_2),
	\end{align*}
	where $h_1'\dot{\times}h_2'\in H^*$ because $\varphi^{-1}(h_1')\varphi^{-1}(h_2')\neq0$. Since $\Psi$ is injective and the right-hand sides depend only on $\xi_1$ and $\xi_2$, the classes defined in Eqs.~(\ref{eq:frac_add})--(\ref{eq:frac_mul}) do not depend on the chosen representatives, and $\Psi(\xi_1\oplus\xi_2)=\Psi(\xi_1)+\Psi(\xi_2)$, $\Psi(\xi_1\otimes\xi_2)=\Psi(\xi_1)\Psi(\xi_2)$. If $\varphi(0)<_\omega h_1'$ and $\varphi(0)<_\omega h_2'$, then $\varphi^{-1}(h_1'),\varphi^{-1}(h_2')>0$, and
	\begin{equation*}
		h_1\dot{\times}h_2'<_\omega h_2\dot{\times}h_1' \iff \varphi^{-1}(h_1)\varphi^{-1}(h_2')<\varphi^{-1}(h_2)\varphi^{-1}(h_1') \iff \Psi(\xi_1)<\Psi(\xi_2),
	\end{equation*}
	so $\prec$ is well defined and $\xi_1\prec \xi_2\Leftrightarrow\Psi(\xi_1)<\Psi(\xi_2)$. This proves (a).

	(b) and (c): Let $\Phi:=\Psi^{-1}:\sQ\rightarrow F_H$. By the above, $\Phi(r_1+r_2)=\Phi(r_1)\oplus\Phi(r_2)$, $\Phi(r_1r_2)=\Phi(r_1)\otimes\Phi(r_2)$ and $r_1<r_2\Leftrightarrow\Phi(r_1)\prec\Phi(r_2)$ for all $r_1,r_2\in\sQ$. Since $(\sQ,+,\times,<)$ is an ordered field and $\Phi(\sQ)=F_H$, Proposition~\ref{prop:transport} shows that $(F_H,\oplus,\otimes,\prec)$ is an ordered field and that $\Phi$ is an isomorphism of ordered fields. For $a,b\in\Dset$ with $b>0$, $\Psi([\varphi(a),\varphi(b)])=a/b$, hence $\Phi(a/b)=[\varphi(a),\varphi(b)]$; in particular, this class does not depend on the representation of $a/b$.

	(d) The representation $r=p/q$ in lowest terms with $q>0$ is unique, and $\varphi$ is injective; hence $r\mapsto(\varphi(p),\varphi(q))$ is injective. By (c), $(\varphi(p),\varphi(q))$ is a representative of $\Phi(r)$.
\end{proof}

\begin{proof}[Proof of Lemma~\ref{lem:surrogates}]
	\emph{Reconstruction.} Fix $a$, suppose that greedy decoding of $\varphi(a)$ does not return $a$, and let $\lambda(a)$ denote the sum inside the expectation that defines $\ell^{\mathrm{CE}}_{\mathrm{rec}}$. If a length classifier predicts a wrong length, the predicted class has at least the probability of the correct one, so the correct length has probability at most $1/2$ and its term in $\lambda(a)$ is at least $\ln2$. Otherwise, both lengths are correct, and the output positions have the exponents of $\str{a}$. Let $i$ be the first position at which the decoded token differs from $y^*_i$. The decoded prefix up to position $i-1$ equals $y^*_{<i}$, so at position $i$ the decoder uses the distribution $\pi_i(\cdot\mid y^*_{<i},\varphi(a))$ and selects a token $y\neq y^*_i$ with $\pi_i(y\mid\cdot)\ge \pi_i(y^*_i\mid\cdot)$ (the type mask allows $y^*_i$, so it cannot remove the better token). Since the two probabilities sum to at most one, $\pi_i(y^*_i\mid\cdot)\le1/2$ and $-\ln \pi_i(y^*_i\mid\cdot)\ge\ln2$. All other terms of $\lambda(a)$ are non-negative. Hence $1(\varphi^\dagger(\varphi(a))\neq a)\le\lambda(a)/\ln2$ for every $a$, and taking expectations gives the first inequality.

	\emph{Order.} Fix $a,b$, write $\bar\omega=\bar\omega(\varphi(a),\varphi(b))$, and suppose that $1(a<b)\neq1(\varphi(a)<_{\bar\omega}\varphi(b))$. If $a<b$, i.e., $\rel(a,b)=1$, then $\bar\omega_1\le\max\{\bar\omega_2,\bar\omega_3\}$, so $\bar\omega_1\le1/2$. Otherwise, $\rel(a,b)\in\{2,3\}$ and $\varphi(a)<_{\bar\omega}\varphi(b)$, so $\bar\omega_1>\bar\omega_{\rel(a,b)}$ and again $\bar\omega_{\rel(a,b)}<1/2$. In both cases $-\ln\bar\omega_{\rel(a,b)}\ge\ln2$. Hence $[1(a<b)-1(\varphi(a)<_{\bar\omega}\varphi(b))]^2\le-\ln\bar\omega_{\rel(a,b)}/\ln2$ for all $a,b$, and taking expectations gives the second inequality. The final statement follows because the right-hand sides vanish if the cross-entropy losses vanish, and $\ell_{\mathrm{rec}}=\Pr[\varphi^\dagger(\varphi(a))\neq a]$ for the discrete metric.
\end{proof}

\subsection{Proofs of the results of Section~\ref{sec:approximate}}\label{app:proofs_approx}

\begin{proof}[Proof of Lemma~\ref{lem:margin}]
	Let $\delta_1:=\min\{\delta_0,\gamma(a)/(2L_\psi)\}>0$ and $z\in\sR^d$ with $\norm{z-\varphi(a)}<\delta_1$. The inequality below, applied to $\psi^{\mathrm{int}}$ and $\psi^{\mathrm{frac}}$ instead of $\psi_i$, shows that both length classifiers select the correct lengths, so the output positions have the exponents of $\str{a}$. We show by induction over $i$ that greedy decoding of $z$ selects $y^*_i$ at position $i$. Suppose that the prefix $y^*_{<i}$ has been selected. For every token $y\neq y^*_i$,
	\begin{align*}
		\psi_i(z)_{y^*_i}-\psi_i(z)_y
		&\ge\psi_i(\varphi(a))_{y^*_i}-\psi_i(\varphi(a))_y-2\norm{\psi_i(z)-\psi_i(\varphi(a))}_\infty\\
		&\ge \gamma(a)-2L_\psi\norm{z-\varphi(a)}>0,
	\end{align*}
	where the Lipschitz condition applies because $z$ lies in the ball of radius $\delta_0$ around $\varphi(a)$. Hence $y^*_i$ is the unique arg-max and is selected. Therefore $\varphi^\dagger(z)=a$ for all such $z$, i.e., $\delta_1\in\mathcal{R}(a)$, and $\rho(a)\ge \delta_1$.
\end{proof}

\begin{proof}[Proof of Proposition~\ref{prop:decoding_error}]
	(a) This is the defining property of $\rho$, since $\norm{z-\varphi(a+b)}=\epsadd(a,b)$.

	(b) By Definition~\ref{def:radius}, every point at distance less than $\rho(c)$ from $\varphi(c)$ is decoded as $c$.

	(c) By (b), the event on the left of Eq.~(\ref{eq:decoding_error}) is contained in the error event, which gives the lower bound. By (a), a decoding error implies $\epsadd(a,b)\ge\rho(a+b)$. Hence the error event is contained in $\{\rho(a+b)\le\tau\}\cup\{\rho(a+b)>\tau,\ \epsadd(a,b)\ge\rho(a+b)\}\subseteq\{\rho(a+b)\le\tau\}\cup\{\epsadd(a,b)\ge\tau\}$. The union bound and Markov's inequality, $\Pr[\epsadd(a,b)\ge\tau]=\Pr[\epsadd(a,b)^2\ge\tau^2]\le\mathbb{E}[\epsadd(a,b)^2]/\tau^2$, give the upper bound. The proof for multiplication is identical.
\end{proof}

\begin{proof}[Proof of Lemma~\ref{lem:ac_lipschitz}]
	Let $u,u',v\in\sR^d$ (in $U$ for the local statement), let $(t^i_1,t^i_2)$ and $(t'^i_1,t'^i_2)$ be the sequences of Eqs.~(\ref{eq:ac_rec1})--(\ref{eq:ac_rec2}) for the inputs $(u,v)$ and $(u',v)$, and let $\eta_i:=\norm{t^i_1-t'^i_1}+\norm{t^i_2-t'^i_2}$; thus $\eta_0=\norm{u-u'}$. By Eq.~(\ref{eq:layer_lipschitz}),
	\begin{align*}
		\norm{t^i_1-t'^i_1}&\le\lambda^{(1)}_i\norm{t^{i-1}_1-t'^{i-1}_1}+\lambda^{(2)}_i\norm{t^{i-1}_2-t'^{i-1}_2},\\
		\norm{t^i_2-t'^i_2}&\le\lambda^{(1)}_i\norm{t^{i-1}_2-t'^{i-1}_2}+\lambda^{(2)}_i\norm{t^{i-1}_1-t'^{i-1}_1}.
	\end{align*}
	Adding both inequalities gives $\eta_i\le c_i\eta_{i-1}$, hence $\eta_k\le\big(\prod_{i=1}^kc_i\big)\eta_0$. Finally,
	\begin{equation*}
		\norm{u*_\Theta v-u'*_\Theta v}=\norm{(t^k_1+t^k_2)-(t'^k_1+t'^k_2)}\le\eta_k\le\Big(\prod_{i=1}^kc_i\Big)\norm{u-u'}.
	\end{equation*}
	The Lipschitz constant in the second argument is the same by Lemma~\ref{lem:commutativity}. For a perceptron layer, $\sigma$ is $1$-Lipschitz for the Euclidean norm because it is $1$-Lipschitz coordinate-wise, so $\norm{\sigma(W_1x_1+W_2x_2+b)-\sigma(W_1x'_1+W_2x'_2+b)}\le\norm{W_1}_2\norm{x_1-x'_1}+\norm{W_2}_2\norm{x_2-x'_2}$. Without AC layers, $\norm{(u+v)-(u'+v)}=\norm{u-u'}$ and $\norm{u\odot v-u'\odot v}=\norm{(u-u')\odot v}\le\norm{v}_\infty\norm{u-u'}$.
\end{proof}

\begin{proof}[Proof of Proposition~\ref{prop:chains}]
	Let $E_j:=\norm{z_j-\varphi(\bar a_j)}$, so $E_1=0$. Since $(\bar a_j,a_{j+1})$ is admissible and $\bar a_{j+1}=\bar a_j*a_{j+1}$,
	\begin{align*}
		E_{j+1}&\le\norm{z_j\,\dot{*}\,\varphi(a_{j+1})-\varphi(\bar a_j)\,\dot{*}\,\varphi(a_{j+1})}+\norm{\varphi(\bar a_j)\,\dot{*}\,\varphi(a_{j+1})-\varphi(\bar a_{j+1})}\\
		&\le L\,E_j+\varepsilon_*(\bar a_j,a_{j+1}).
	\end{align*}
	Unrolling this recursion from $E_1=0$ gives $E_M\le\sum_{j=1}^{M-1}L^{M-1-j}\varepsilon_*(\bar a_j,a_{j+1})$, and bounding each defect by $\bar\varepsilon$ gives the second inequality of Eq.~(\ref{eq:chains}). The final statement follows from the definition of $\rho(\bar a_M)$.
\end{proof}

\begin{proof}[Proof of Proposition~\ref{prop:assoc_distrib}]
	Write $u_a=\varphi(a)$ etc. For the left side of Eq.~(\ref{eq:assoc_bound}),
	\begin{align*}
		\norm{(u_a\dot{+}u_b)\dot{+}u_c-\varphi(a+b+c)}
		&\le\norm{(u_a\dot{+}u_b)\dot{+}u_c-\varphi(a+b)\dot{+}u_c}+\norm{\varphi(a+b)\dot{+}u_c-\varphi(a+b+c)}\\
		&\le L_+\epsadd(a,b)+\epsadd(a+b,c),
	\end{align*}
	and, using the Lipschitz property in the second argument, $\norm{u_a\dot{+}(u_b\dot{+}u_c)-\varphi(a+b+c)}\le L_+\epsadd(b,c)+\epsadd(a,b+c)$. The triangle inequality gives Eq.~(\ref{eq:assoc_bound}). For Eq.~(\ref{eq:distrib_bound}),
	\begin{align*}
		\norm{u_a\dot{\times}(u_b\dot{+}u_c)-\varphi(a(b+c))}
		&\le\norm{u_a\dot{\times}(u_b\dot{+}u_c)-u_a\dot{\times}\varphi(b+c)}+\norm{u_a\dot{\times}\varphi(b+c)-\varphi(a(b+c))}\\
		&\le L_\times\epsadd(b,c)+\epsmul(a,b+c).
	\end{align*}
	Since $\dot{+}$ is $L_+$-Lipschitz in each argument, $\norm{(v\dot{+}w)-(v'\dot{+}w')}\le L_+(\norm{v-v'}+\norm{w-w'})$, and with $ab+ac=a(b+c)$,
	\begin{equation*}
		\norm{(u_a\dot{\times}u_b)\dot{+}(u_a\dot{\times}u_c)-\varphi(a(b+c))}\le L_+\big[\epsmul(a,b)+\epsmul(a,c)\big]+\epsadd(ab,ac).
	\end{equation*}
	The triangle inequality gives Eq.~(\ref{eq:distrib_bound}); the decoding statements follow from the definition of the decoding radius. The multiplicative analogue of Eq.~(\ref{eq:assoc_bound}) is proved in the same way.
\end{proof}

\paragraph{Capacity of the embedding space}\label{app:capacity} The decoding radii relate the embedding dimension to the number of digits and to the defects that can be tolerated.

\begin{proposition}[Embedding dimension and number of digits]\label{prop:capacity}
	Suppose that $\norm{\varphi(a)}\le B$ for all $a\in\Dn$.
	\begin{enumerate}
		\item[(a)] For every $\tau>0$, at most $(1+B/\tau)^d$ elements $a\in\Dn$ satisfy $\rho(a)\ge\tau$.
		\item[(b)] If $\rho(a)\ge\rho_{\min}>0$ for all $a\in\Dn$, then
		\begin{equation}\label{eq:capacity}
			N\le d\,\log_{10}\!\Big(1+\frac{B}{\rho_{\min}}\Big),
			\qquad\text{equivalently}\qquad
			\rho_{\min}\le\frac{B}{10^{N/d}-1}\le\frac{B\,d}{N\ln 10}.
		\end{equation}
	\end{enumerate}
\end{proposition}

By Proposition~\ref{prop:capacity}, the number of robustly decodable digits grows at most linearly with $d$, and the largest defect that Proposition~\ref{prop:decoding_error}(a) can certify for all results is at most $B\,d/(N\ln 10)$, so it shrinks in proportion to $N/d$. Finite precision adds a limit: rounding to a floating-point format with unit roundoff $u$ moves an embedding by up to $uB$, so if every embedding must stay decodable after rounding, then $\rho_{\min}\ge uB$ and $N\le d\log_{10}(1+1/u)$, that is, $N\le 2.41\,d$ in bfloat16 ($u=2^{-8}$). The bound is a necessary condition, not a guarantee that a trained model reaches it.

\begin{proof}[Proof of Proposition~\ref{prop:capacity}]
	For distinct $a,c\in\Dn$, the open balls of radii $\rho(a)$ and $\rho(c)$ around $\varphi(a)$ and $\varphi(c)$ are disjoint, because a point in both would be decoded both as $a$ and as $c$ (Definition~\ref{def:radius}).

	(a) Let $S:=\{a\in\Dn:\rho(a)\ge\tau\}$. The balls of radius $\tau$ around $\varphi(a)$, $a\in S$, are pairwise disjoint and contained in the ball of radius $B+\tau$ around the origin. Comparing volumes in $\sR^d$ gives $|S|\,\tau^d\le(B+\tau)^d$, i.e., $|S|\le(1+B/\tau)^d$.

	(b) Apply (a) with $\tau=\rho_{\min}$ and $S=\Dn$. Since $\Dn$ contains the $10^N$ integers $0,1,\dots,10^N-1$ (only the digits count towards $N$, Eq.~(\ref{eq:Dn})), we obtain $10^N\le(1+B/\rho_{\min})^d$, which is the first inequality in Eq.~(\ref{eq:capacity}). Solving for $\rho_{\min}$ gives the second, and $e^y-1\ge y$ with $y=(N/d)\ln 10$ gives the third.
\end{proof}

\subsection{Proof of Lemma~\ref{lem:small_sum}}\label{app:proof_small_sum}

\begin{proof}
	For a fixed $k$, every pair $(i,j)$ with $i+j=k$ contains one digit of the shorter operand, so at most $m$ pairs contribute to $L_k$ and $H_k$. Since $a_ib_j\le 81$, $a_ib_j\bmod 10\le 9$ and $\lfloor a_ib_j/10\rfloor\le 8$; hence $L_k\le 9m\le 225$ and $H_k\le 8m\le 200$ have at most three digits, and $L^{(2)}_k,H^{(2)}_k\le 2$. Writing $a_ib_j=(a_ib_j\bmod 10)+10\lfloor a_ib_j/10\rfloor$,
	\begin{equation*}
		|ab|=\sum_k\sum_{i+j=k}a_ib_j10^k=\sum_kL_k10^k+\sum_kH_k10^{k+1}=\sum_k\sum_{r=0}^{2}\big(L^{(r)}_{k-r}+H^{(r)}_{k-1-r}\big)10^k=\sum_kS_k10^k .
	\end{equation*}
	Each of the six terms of $S_k$ is a decimal digit, and two of them are at most $2$, so $0\le S_k\le 4\cdot 9+2\cdot 2=40$. With $S_k=(S_k\bmod 10)+10\lfloor S_k/10\rfloor$, the same shift gives $|ab|=\sum_kT_k10^k$ and $0\le T_k\le 9+4=13$. Finally, $\kappa_{k_0}=0$ at the lowest exponent $k_0$, and if $\kappa_k\le 1$, then $T_k+\kappa_k\le 14$ and $\kappa_{k+1}=\lfloor(T_k+\kappa_k)/10\rfloor\le 1$. With the sum running up to one exponent above the highest non-zero $T_k$, so that the last carry is included, the carry pass preserves $\sum_k(T_k+\kappa_k-10\kappa_{k+1})10^k=\sum_kT_k10^k$, and it produces the digits $d_k\in\{0,\dots,9\}$ of $|ab|$.

	For $25<m\le 200$ (the models of Section~\ref{subsubsec:ablation}), $L_k\le 9m\le1800$ and $H_k\le 8m\le1600$ have at most four digits, with $L^{(3)}_k,H^{(3)}_k\le1$. The sum over $r$ then runs up to $r=3$, so $S_k$ has the two further terms $L^{(3)}_{k-3}$ and $H^{(3)}_{k-4}$, and the same steps give $S_k\le 6\cdot9+2\cdot1=56$, $T_k\le 9+5=14$, $T_k+\kappa_k\le15$ and again $\kappa_{k+1}\le1$.
\end{proof}

\FloatBarrier

\end{document}